\documentclass[journal]{IEEEtran}

\usepackage{bobbystyle_IEEEjournal}
\usepackage{ulem}

\usepackage{tikz}
\usetikzlibrary{spy}

\usepackage{subfigure}
\usepackage{chngcntr}
\counterwithout{equation}{section}

\makeatletter
\newcommand{\vast}{\bBigg@{3}}
\newcommand{\Vast}{\bBigg@{5}}
\makeatother

\renewcommand\thetheorem{\arabic{theorem}}
\renewcommand{\theequation}{\arabic{equation}}

\newcommand{\dquotes}[1]{``#1''}

\newcolumntype{C}[1]{>{\centering\let\newline\\\arraybackslash\hspace{0pt}}m{#1}}

\newenvironment{myquote}[1]%
  {\list{}{\leftmargin=#1\rightmargin=#1}\item[]}%
  {\endlist}

\usepackage{xr}
\begin{document}

\title{Sparse-View X-Ray CT Reconstruction \\ Using $\ell_1$ Prior with Learned Transform \vspace{0.05in}}

\author{Xuehang Zheng$^\dagger$, Il Yong Chun$^\dagger$, \textit{Member}, \textit{IEEE}, Zhipeng Li, \textit{Student Member}, \\ Yong Long, \textit{Member}, \textit{IEEE}, and Jeffrey A. Fessler, \textit{Fellow}, \textit{IEEE}

\thanks{
$\dagger$\textit{These two authors contributed equally to this work.}
}


\thanks{This work is supported in part by the National Natural Science Foundation of China under Grant 61501292, and in part by the NIH under Grant U01 EB018753. \textit{ (Corresponding author: Yong Long.)}}

\thanks{
Xuehang Zheng, Zhipeng Li and Yong Long are with the University of Michigan - Shanghai Jiao Tong University Joint Institute, Shanghai Jiao Tong University, Shanghai 200240, China (email: \{zhxhang, zhipengli, yong.long\}@sjtu.edu.cn). 
	
Il Yong Chun and Jeffrey A. Fessler are with the Department of Electrical Engineering and Computer Science, 
The University of Michigan, Ann Arbor, MI 48019 USA (email: \{iychun, fessler\}@umich.edu).  
}
}

\maketitle

\begin{abstract}
A major challenge in X-ray computed tomography (CT) is reducing radiation dose while maintaining high quality of reconstructed images.
To reduce the radiation dose, one can reduce the number of projection views (sparse-view CT); however, it becomes difficult to achieve high-quality image reconstruction as the number of projection views decreases. 
Researchers have applied the concept of learning sparse representations from (high-quality) CT image dataset to the sparse-view CT reconstruction.
We propose a new statistical CT reconstruction model that combines penalized weighted-least squares (PWLS) and $\ell_1$ prior with learned sparsifying transform (PWLS-ST-$\ell_1$), and a corresponding efficient algorithm based on Alternating Direction Method of Multipliers (ADMM).
To moderate the difficulty of tuning ADMM parameters, we propose a new ADMM parameter selection scheme based on approximated condition numbers.
We interpret the proposed model by analyzing the minimum mean square error of its ($\ell_2$-norm relaxed) image update estimator.
Our results with the extended cardiac-torso (XCAT) phantom data and clinical chest data show that, for sparse-view 2D fan-beam CT and 3D axial cone-beam CT, PWLS-ST-$\ell_1$ improves the quality of reconstructed images compared to the CT reconstruction methods using edge-preserving regularizer and $\ell_2$ prior with learned ST. 
These results also show that, for sparse-view 2D fan-beam CT, PWLS-ST-$\ell_1$ achieves comparable or better image quality and requires much shorter runtime than PWLS-DL using a learned overcomplete dictionary.
Our results with clinical chest data show that, methods using the unsupervised learned prior generalize better than a state-of-the-art deep \dquotes{denoising} neural network that does not use a physical imaging model. 

\end{abstract}



\section{Introduction}
\label{sec:intro}
Radiation dose reduction is a major challenge in X-ray computed tomography (CT).
Sparse-view CT reduces dose by acquiring fewer projection views \cite{Chen&Tang&Leng:08MP, Chun&Talavage:13Fully3D}. 
However, as the number of projection views decreases, it becomes harder to achieve high quality (high resolution, contrast, and signal-to-noise ratio) image reconstruction.
Inspired by compressed sensing theories exploiting sparsity of signals \cite{Foucart&Rauhut:book, Adcock&etal:05bookCh, Chun&Adcock:17TIT}, there have been many studies of sparse-view CT reconstruction with total variation \cite{Sidky&Kao&Pan:06, Yu&Wang:09PMB, Bian&etal:10PMB, Ramani&Fessler:12MI, Nie&etal:14PMB} or other sparsity promoting regularizers \cite{Chen&Tang&Leng:08MP, Chun&Talavage:13Fully3D}. 

Researchers have applied (deep) neural networks (NNs) to sparse-view and low-dose CT reconstruction problems. 
Early works focused on image denoising \cite{Chen&etal:17TMI, Kang&Min&Ye:17MP, Wolterink&etal:17TMI, Jin&etal:17TIP, Ye&Han&Cha:18SJIS} using the good mapping capabilities of deep NNs.
However, the greater mapping capability can increase the chance of causing some artificial features when test images are not similar to training images. (See Fig.~\ref{fig:fbpconvnet:ref}.)
More recent works combined image mapping NNs with model-based image reconstruction (MBIR) frameworks that consider CT physics \cite{Chen&etal:18TMI, wu:17:ild, Chun&Fessler:18IVMSP, Chun&etal:18Allerton, Chun&etal:18arXiv:momnet}. 
However, for \textit{general} image mapping NNs, it is difficult to explicitly write the corresponding optimization problems within an MBIR framework. Without explicit cost functions, it is challenging to guarantee the non-expansiveness (or $1$-Lipschitz continuity) of the image mapping NNs and obtain \dquotes{optimal} and convergent image reconstruction, especially when the mapping NNs are identical across iterations \cite{Chun&etal:18Allerton}.
In addition, considering that the methods are trained with supervised learning, one would expect optimal results by using pairs of \dquotes{noiseless} and \dquotes{noisy} images in the training processes. 
In practice, however, it is challenging to obtain noiseless images to construct such paired training dataset in CT imaging.
Based on a recent \dquotes{Noise2Noise} training method \cite{lehtinen:18:nli}, some recent works \cite{pelt:18:itr, Yuan&etal:19Fully3D} show that training image mapping NNs with pairs of noisy images could provide satisfactory image quality in certain applications. However, training with noisy images has certain limitations in sparse-view CT reconstruction (see Section~\ref{sec:noisy_targets} in the supplement).

Alternatively, researchers have learned prior information in an unsupervised way by using (unpaired) datasets that consist of high-quality images, and exploited it for solving inverse problems \cite{Chun&Fessler:18cao, Chun&etal:19SPL, chun:18:cdl, Chun&Fessler:17SAMPTA, Aharon&Elad&Bruckstein:06TSP, Cai&etal:14ACHA, Ravishankar&Bresler:15TSP, Xu&etal:12TMI, pfister:14:mbi, Zhang&etal:16BEO, Zheng&etal:16IVMSP, Zheng&etal:17Fully3D, Zheng&etal:17arXiv}.
This unsupervised framework can resolve the aforementioned issues of the supervised framework.
The corresponding learned operators sparsify a specific set of training images, but have the potential to represent a broader range of test images compared to the supervised image mapping NNs. (See Fig.~\ref{fig:fbpconvnet:ref}.)
In addition, one can explicitly formulate an optimization problem for image recovery using the learned sparsifying operators.
Particularly, in MBIR algorithms, the authors in \cite{Chun&etal:18Allerton} show that the learned convolutional analysis operator using tight-frame (TF) filters \cite{Chun&Fessler:18cao} becomes a non-expansive image mapping autoencoder (of encoding convolution, nonlinear thresholding, and decoding convolution \cite{Chun&Fessler:18cao}).
The unsupervised framework has been widely applied in image denoising problems and provided promising results \cite{Aharon&Elad&Bruckstein:06TSP, Cai&etal:14ACHA, Ravishankar&Bresler:15TSP, chun:18:cdl, Chun&Fessler:17SAMPTA}.
Recently, patch-based sparsifying operator learning frameworks \cite{Aharon&Elad&Bruckstein:06TSP, Cai&etal:14ACHA, Ravishankar&Bresler:15TSP} have been successfully applied to improve low-dose CT reconstruction \cite{Xu&etal:12TMI, pfister:14:mbi, Zhang&etal:16BEO, Zheng&etal:16IVMSP, Zheng&etal:17Fully3D, Zheng&etal:17arXiv}.  
The authors in \cite{Zheng&etal:17arXiv} reported that a union of transforms learned via clustering different features can further improve image quality of reconstructions over the low-dose CT reconstruction method using a (square) sparsifying transform (ST) \cite{Zheng&etal:16IVMSP}.
In some computer vision applications, the studies \cite{Lu&etal:13CVPR, Jiang&etal:15IJCAI} show that robust dictionary learning incorporating $\ell_1$ prior outperforms that using $\ell_2$ prior when outliers exist.

\begin{figure*}[!t]
	
	
	\centering
	\small\addtolength{\tabcolsep}{-10.5pt}
	
	\begin{tabular}{ccccc}
	
	{} & \small{@$1\mathrm{st}$ outer iteration} & \small{@$2\mathrm{nd}$ outer iteration} & \small{@$500\rth$ outer iteration} & \small{@$1000\rth$ outer iteration} \\
	
	\raisebox{-.5\height}{\begin{turn}{+90} \scriptsize{Probability density function} \end{turn}}~ &
	\raisebox{-.5\height}{
		\begin{tikzpicture}
		\begin{scope}[spy using outlines={rectangle,red,magnification=1.6,size=16mm, connect spies}]
		\node {\includegraphics[width=0.25\textwidth]{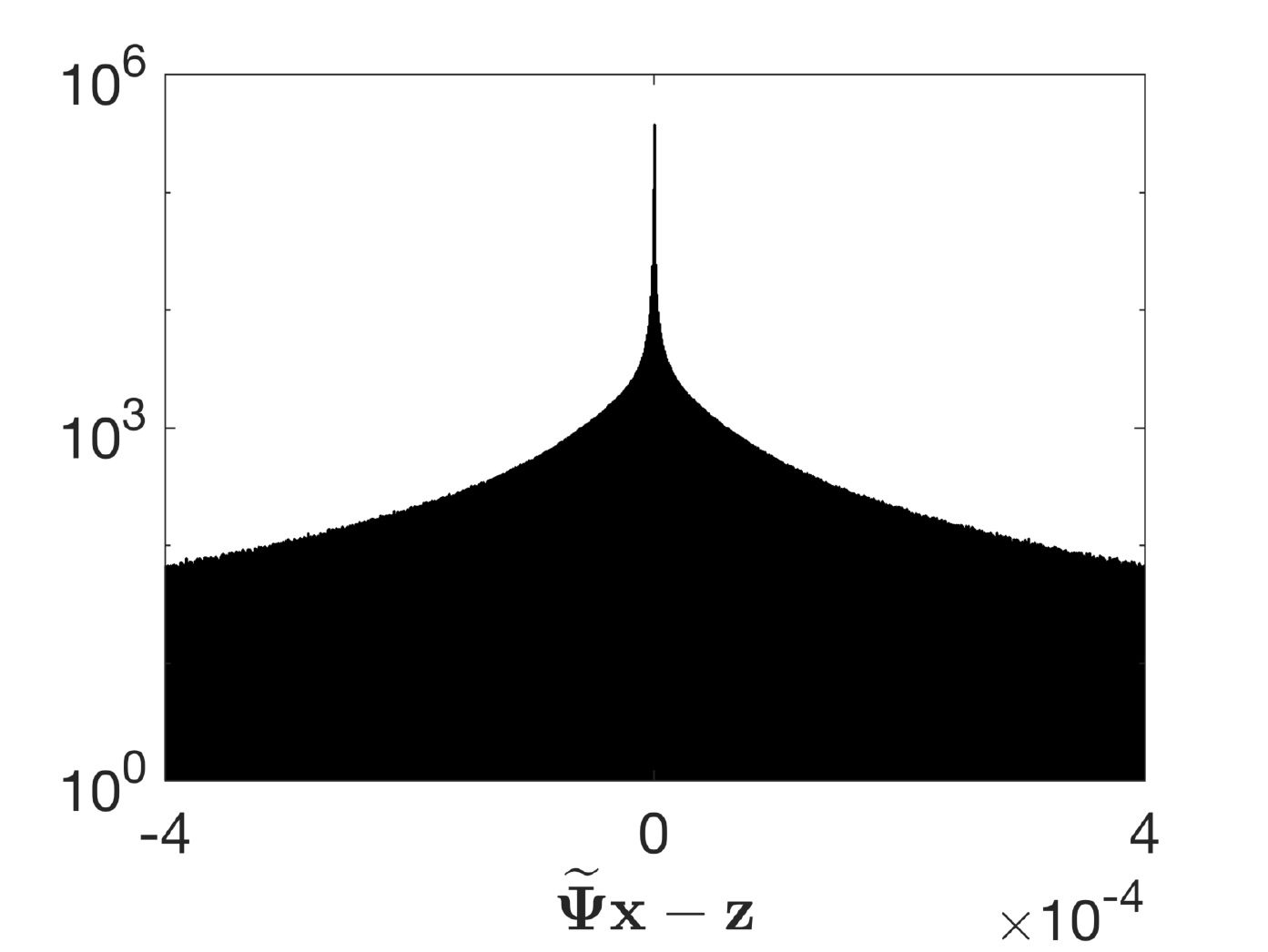}};
		\end{scope}
		\end{tikzpicture}} &
	
	\raisebox{-.5\height}{
		\begin{tikzpicture}
		\begin{scope}[spy using outlines={rectangle,red,magnification=1.6,size=16mm, connect spies}]
		\node {\includegraphics[width=0.25\textwidth]{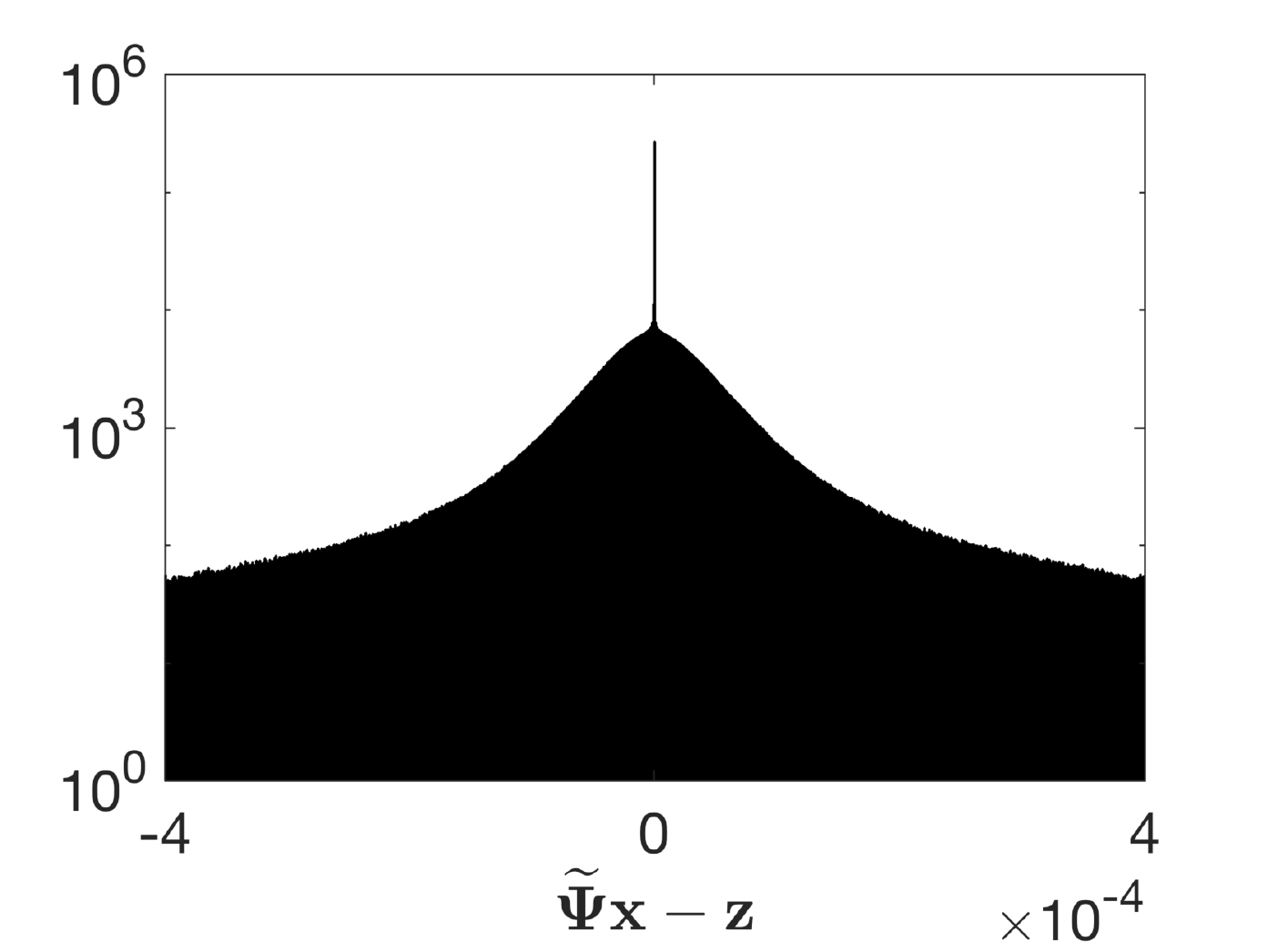}};
		\end{scope}
		\end{tikzpicture}} &
	
	\raisebox{-.5\height}{
		\begin{tikzpicture}
		\begin{scope}[spy using outlines={rectangle,red,magnification=1.6,size=16mm, connect spies}]
		\node {\includegraphics[width=0.25\textwidth]{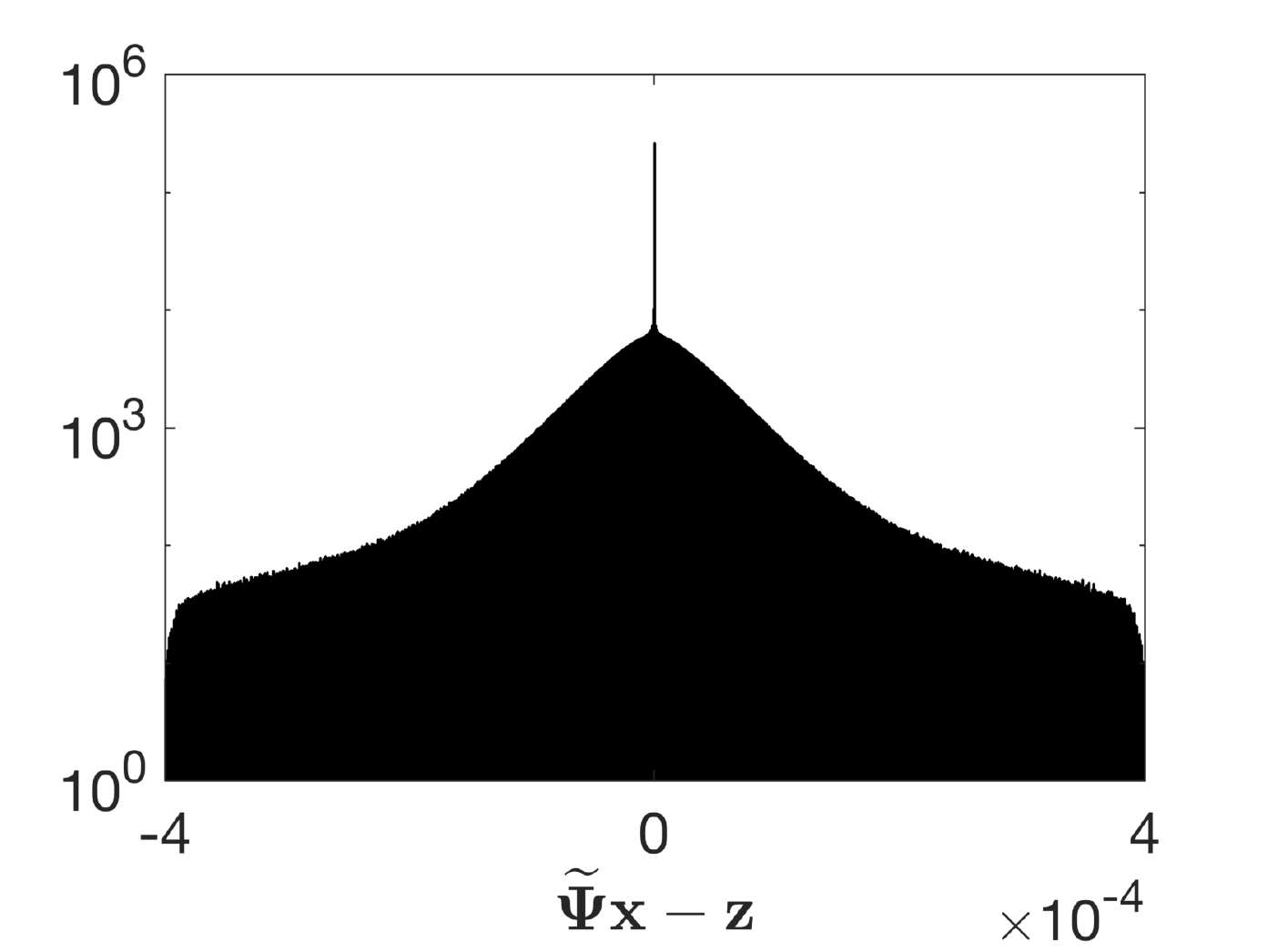}};
		\end{scope}
		\end{tikzpicture}} &
	
	\raisebox{-.5\height}{
		\begin{tikzpicture}
		\begin{scope}[spy using outlines={rectangle,red,magnification=1.6,size=16mm, connect spies}]
		\node {\includegraphics[width=0.25\textwidth]{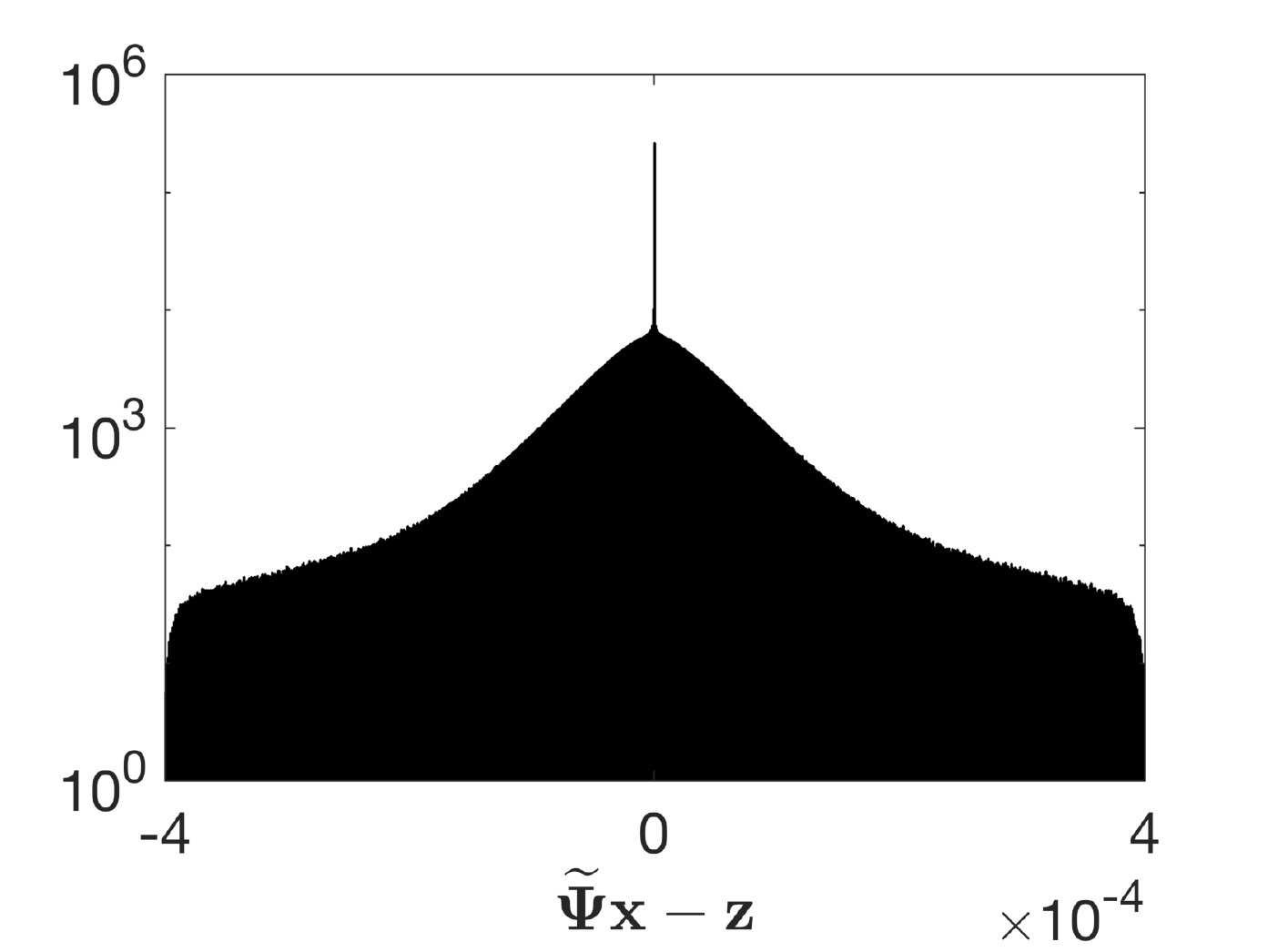}};
		\end{scope}
		\end{tikzpicture}} 
	
\end{tabular}
	
	\caption{Histograms of sparsification error $\widetilde{\mb{\Psi}} \mb{x} - \mb{z}$ at different outer iterations of the PWLS-ST-$\ell_2$ method (in the XCAT phantom experiment with 2D fan-beam geometry and $12.5\%$ ($123$) projection views). Y-axis uses a logarithmic scale, and the range of $\widetilde{\mb{\Psi}} \mb{x} - \mb{z}$ is within $[-4, 4]\times10^{-4}$. Over the iterations, the sparsification error histograms appear more like a Laplace distribution than a Gaussian distribution.}
	\label{fig:histogram}
\end{figure*}

This paper was inspired by a simple observation related to our recent study \cite{Zheng&etal:16IVMSP}: 
for the penalized weighted-least squares (PWLS) reconstruction method using $\ell_2$ prior with a learned ST (PWLS-ST-$\ell_2$) \cite{Zheng&etal:16IVMSP}, the sparsification error histograms match a Laplace distribution over the iterations; see Fig.~\ref{fig:histogram}.
The question then arises, \dquotes{Does the learned prior experience model mismatch in testing stage?}
To answer this question, we aim to investigate learned STs for regularization.
This paper 
	\begin{itemize}
		\item[\textit{1)}] proposes a new MBIR model that combines PWLS and $\ell_1$ prior with learned ST (PWLS-ST-$\ell_1$), 
		\item[\textit{2)}] develops a corresponding efficient algorithm based on Alternating Direction Method of Multipliers (ADMM) \cite{Boyd&Parikh&Chu&Peleato&Eckstein:11FTML} with a new ADMM parameter selection scheme based on approximated condition numbers, 
		\item[\textit{3)}] and interprets the proposed model by analyzing the empirical mean square error (MSE) of its image update estimator, and 
		the minimum mean square error (MMSE) of its $\ell_2$-norm relaxed image update estimator.
	\end{itemize}

Our results with the extended cardiac-torso (XCAT) phantom data \cite{Segars&etal:08MP} and clinical chest data show that, for sparse-view 2D fan-beam CT and 3D axial cone-beam CT, PWLS-ST-$\ell_1$ improves the reconstruction quality compared to a PWLS reconstruction method with an edge-preserving regularizer (PWLS-EP), and PWLS-ST-$\ell_2$. 
These results also show that, for sparse-view 2D fan-beam CT, PWLS-ST-$\ell_1$ achieves comparable or better image quality and requires much shorter runtime than PWLS-DL using a learned overcomplete dictionary.
Our results with clinical chest data show that, MBIR methods using the learned prior in an unsupervisedly fashion generalize better than FBPConvNet \cite{Jin&etal:17TIP}, a state-of-the-art deep \dquotes{denoising} neural network.

For sparse-view CT application, a similar approach that uses $\ell_1$ prior with a dictionary was introduced in \cite{Zhang&etal:16BEO}; however, there are three major differences. First, we focus on an \textit{analysis} approach (e.g., transform and convolutional analysis operator \cite{Chun&Fessler:18cao}), whereas \cite{Zhang&etal:16BEO} is based on a \textit{synthesis} perspective (e.g., dictionary and convolutional dictionary \cite{chun:18:cdl}). 
Second, we pre-learn our signal model in an unsupervised way and exploit it in CT MBIR as a prior, whereas \cite{Zhang&etal:16BEO} adaptively estimates the dictionary in reconstruction. Because their dictionary changes during reconstruction, their main concern is not related to the model mismatch. Third, we directly solve the $\ell_1$ minimization via ADMM, whereas \cite{Zhang&etal:16BEO} uses a reweighted-$\ell_2$ minimization framework.
Our previous conference paper \cite{Chun&etal:17Fully3D} presented a brief study of the proposed PWLS-ST-$\ell_1$ model for sparse-view 2D fan-beam CT scans with XCAT phantom. 
	This paper extends our previous work to 3D axial cone-beam CT, 
	describes and investigates our parameter selection strategy for PWLS-ST-$\ell_1$,
	analyzes the proposed model by the empirical MSE and MMSE of its image update estimator,
	and performs more comprehensive comparisons with recent methods using both simulated and real clinical data.

The remainder of this paper is organized as follows. 
Section~\ref{sec:methods} describes the formulation for pre-learning STs, and proposes the MBIR model and algorithm for PWLS-ST-$\ell_1$.
For the proposed algorithm, Section~\ref{sec:methods} introduces our preconditioner designs for sparse-view 2D fan-beam CT and 3D axial cone-beam CT, and proposes a new ADMM parameter selection scheme based on approximated condition numbers.
Section~\ref{sec:methods} provides interpretations of the proposed model via analyzing the empirical MSE of its image update estimator, and the MMSE of its $\ell_2$-norm relaxed image update estimator.
Section~\ref{sec:results} reports detailed experimental results and comparisons to several recent methods.
Section~\ref{sec:conclusion} presents our conclusions and mentions future directions.

\section{Proposed Models and Algorithm}\label{sec:methods}


The proposed approach has two stages: training and testing. First, we learn a square ST from a dataset of high-quality CT images. 
Then, we apply the learned ST with $\ell_1$ prior to reconstruct images from lower dose (or sparse-view) CT data.
This section describes the formulation for pre-learning a square ST, proposes the PWLS reconstruction model using $\ell_1$ prior with learned ST and its corresponding algorithm, and interprets the proposed model.

\subsection{Offline Learning Sparsifying Transform}
\label{subsec:training_model}
We pre-learn a ST by solving the following problem \cite{Ravishankar&Bresler:15TSP} (mathematical notations are detailed in Appendix \ref{sec:notation}):
\ea{
\label{eq:l1LearnSpTrasf}
\argmin_{\substack{\mb{\Psi} \in \bbR^{n \times n}}} \min_{\{ \mb{z}_j \in \bbR^{n} \} }  &~ \sum_{j=1}^{J'} \left\| \mb{\Psi} \mb{x}_j  - \mb{z}_j \right\|_2^2 + \gamma' \left\| \mb{z}_j \right\|_0 
\nn \\
&~ + \tau \left( \xi \| \mb{\Psi} \|_F^2 - \log \left| \det \mb{\Psi} \right| \right) 
}
where $\mb{\Psi} \in \bbR^{n \times n}$ is a square ST, $\{ \mb{x}_j \in \bbR^n : j = 1,\ldots,J' \}$ is a set of $J'$ patches extracted from training data, $\mb{z}_j \in \bbR^n$ is the sparse code corresponding to the $j\text{th}$ patch $\mb{x}_j$, $n$ is the number of pixels (voxels) in each (vectorized) patch, $J'$ is the total number of the image patches, and $\gamma', \tau,\xi > 0$ are regularization parameters.

\subsection{CT Reconstruction Model Using $\ell_1$ Prior with Learned Sparsifying Transform: PWLS-ST-$\ell_1$}

To reconstruct a linear attenuation coefficient image $\mb{x} \in \bbR^N$ from post-log measurement $\mb{y} \in \bbR^m$, we solve the following non-convex MBIR problem using PWLS and the ST $\mb{\Psi}$ learned via \R{eq:l1LearnSpTrasf} \!\!\cite{Chun&etal:17Fully3D}:
\be{
\label{sys:L1trsf}
\argmin_{\mb{x} \in \bbR^{N} } \min_{\mb{z} \in \bbR^{n J}} \frac{1}{2} \left\| \mb{y} - \mb{A} \mb{x} \right\|_\mb{W}^2 + \lambda \left\| \widetilde{\mb{\Psi}} \mb{x} - \mb{z} \right\|_1 + \gamma \left\| \mb{z} \right\|_0,
\tag{P}
}
where
\bes{ 
\widetilde{\mb{\Psi}} = \left[ \begin{array}{c} \mb{\Psi} \mb{P}_1 \\ \vdots \\ \mb{\Psi} \mb{P}_J \end{array} \right] \quad \textmd{and} \quad \mb{z} = \left[ \begin{array}{c} \mb{z}_1 \\ \vdots \\ \mb{z}_J \end{array} \right]. \nn
}
Here, $\mb{A} \in \bbR^{m \times N}$ is a CT scan system matrix, $\mb{W} \in \bbR^{m \times m}$ is a diagonal weighting matrix with elements $\{ W_{l,l} = \rho_l^2 / ( \rho_l + \sigma^2 ) : l = 1,\ldots,m \}$ based on a Poisson-Gaussian model for the pre-log measurements $\rhobf \in \bbR^m$ with electronic readout noise variance $\sigma^2$ \cite{Thibault&Bouman&Sauer&Hsieh:06SPIE}, $\mb{P}_j \in \bbR^{n \times N}$ is a patch-extraction operator for the $j\text{th}$ patch, $\mb{z}_j \in \bbR^n$ is unknown sparse code for the $j\text{th}$ patch, $J$ is the number of extracted patches, and $\lambda, \gamma > 0$ are regularization parameters. 

The term $\| \widetilde{\mb{\Psi}} \mb{x}  - \mb{z} \|_1$ denotes a $\ell_1$-based sparsification error \cite{Foucart&Rauhut:book, Adcock&etal:05bookCh, Chun&Adcock:17TIT}. 
We expect $\ell_1$ to be more robust to sparsity model mismatch than the $\ell_2$-based sparsification error $\| \widetilde{\mb{\Psi}} \mb{x}  - \mb{z} \|_2^2$ used in \cite{Zheng&etal:16IVMSP, Zheng&etal:17arXiv}. 
Fig.~\ref{fig:histogram} shows histograms of sparsification error $\widetilde{\mb{\Psi}} \mb{x} - \mb{z}$ at different outer iterations of the PWLS-ST-$\ell_2$ method. Over the iterations, the sparsification error histograms appear more like a Laplace distribution than a Gaussian distribution. This observation suggests that the proposed $\ell_1$ prior model is more suitable than the $\ell_2$ prior model for PWLS-ST-based reconstruction.
Section~\ref{subsec: MBIR} shows that the proposed $\ell_1$-based sparsification error term, $\| \widetilde{\mb{\Psi}} \mb{x}  - \mb{z} \|_1$, improves the accuracy of reconstruction compared to the $\ell_2$ prior model in \cite{Zheng&etal:16IVMSP, Zheng&etal:17arXiv}.


\subsection{Proposed Algorithm for PWLS-ST-$\ell_1$}  \label{sec:Alg}

\begin{algorithm}[!b]
\caption{PWLS-ST-$\ell_1$ CT Reconstruction}
\label{alg:PWLS-ST-l1}

\begin{algorithmic}
\REQUIRE $\mb{y}$, $\mb{x}^{(1)}$, $\mb{z}^{(1)}$, $\mb{\Psi}$ learned from \R{eq:l1LearnSpTrasf}, $\mb{M}$, $\mb{W}$, $\lambda, \gamma, \mu, \nu \geq 0$, $i=1$

\WHILE{a stopping criterion is not satisfied}

\STATE Set $\mb{b}_{a} = \mb{b}_{\psi} = 0$

\FOR{$i' = 1,\ldots, \mathrm{Iter}_\mathrm{ADMM}$}

\STATE Obtain $\displaystyle \mb{x}^{(i'+1)}$ by solving \R{eq:algoAR} with PCG($\mb{M}$)
\STATE $\displaystyle \mb{d}_a^{(i'+1)} = \big( \mb{W} + \mu \mb{I}_m \big)^{-1} \Big( \mb{W} \mb{y} + \mu \Big( \mb{A} \mb{x}^{(i'+1)} + \mb{b}_a^{(i')} \Big) \Big)$
\STATE $\displaystyle d_{\psi,j}^{(i'+1)} = \cS \bigg( \! \Big( \widetilde{\mb{\Psi}} \mb{x}^{(i'+1)} - \mb{z}^{(i)} + \mb{b}_{\psi}^{(i')} \Big)_{j}, \frac{\lambda}{\mu \nu} \bigg)$, $\forall j$
\STATE $\displaystyle \mb{b}_{a}^{(i'+1)} = \mb{b}_{a}^{(i')} - \Big( \mb{d}_{a}^{(i'+1)} - \mb{A} \mb{x}^{(i'+1)} \Big)$
\STATE $\displaystyle \mb{b}_{\psi}^{(i'+1)} = \mb{b}_{\psi}^{(i')} - \Big(  \mb{d}_{\psi}^{(i'+1)} - \Big( \widetilde{\mb{\Psi}} \mb{x}^{(i'+1)} - \mb{z}^{(i)} \Big) \Big)$

\ENDFOR


\STATE $\displaystyle z_j^{(i+1)} = \cH \bigg( \! \Big( \widetilde{\mb{\Psi}} \mb{x}^{(\mathrm{Iter}_\mathrm{ADMM}+1)} \Big)_{j} , \frac{\gamma}{\lambda} \bigg)$, $\forall j$

	
\STATE $i = i + 1$
	
\ENDWHILE
	
\end{algorithmic}

\end{algorithm}

To solve \R{sys:L1trsf}, our proposed algorithm alternates between updating the image $\mb{x}$ (\textit{image update step}) and the sparse codes $\mb{z}$ (\textit{sparse coding step}).
For the image update, we apply ADMM \cite{Boyd&Parikh&Chu&Peleato&Eckstein:11FTML, Ramani&Fessler:12MI, Chun&Talavage:13Fully3D} by introducing an auxiliary variable to separate the effects of a certain variable \cite{Ramani&Fessler:12MI, Chun&Adcock&Talavage:15TMI}. 
For efficient sparse coding, we apply an analytical solution for $\mb{z}$.
The following subsections provide details for solving \R{sys:L1trsf}, summarized in Algorithm~\ref{alg:PWLS-ST-l1}, introduce our preconditioner designs, and decribe a new ADMM parameter selection scheme based on approximated condition numbers.

\subsubsection{Image Update - ADMM} \label{sec:image}

Using the current sparse code estimates $\mb{z}$, we update the image $\mb{x}$ by augmenting \R{sys:L1trsf} with auxiliary variables \!\!\cite{Chun&etal:17Fully3D}:
\eas{
\argmin_{\substack{\mb{x} \in \bbR^N, \mb{d}_a \in \bbR^m, \\ \mb{d}_{\psi} \in \bbR^{nJ}}} &~ \frac{1}{2} \left\| \mb{y} - \mb{d}_a \right\|_\mb{W}^2 + \lambda \left\| \mb{d}_{\psi} \right\|_1 
\\
\mbox{subject to} ~~&~ \left[ \begin{array}{c} \mb{d}_a \\  \mb{d}_{\psi} \end{array} \right] = \left[ \begin{array}{c} \mb{A} \\  \widetilde{\mb{\Psi}} \end{array} \right] \mb{x} - \left[ \begin{array}{c} \mb{0} \\  \mb{z} \end{array} \right].
}
The corresponding augmented Lagrangian has the form
\begingroup
\allowdisplaybreaks
\eas{
&~ \frac{1}{2} \left\| \mb{y} - \mb{d}_a \right\|_\mb{W}^2 + \lambda \left\| \mb{d}_{\psi} \right\|_1 + \frac{\mu}{2} \left\| \mb{d}_a - \mb{A}\mb{x} - \mb{b}_a \right\|_2^2
\\
&~ + \frac{\mu \nu}{2} \left\| \mb{d}_{\psi} - \left( \widetilde{\mb{\Psi}} \mb{x} - \mb{z} \right) - \mb{b}_{\psi} \right\|_2^2.
}
\endgroup
We descend/ascend this augmented Lagrangian using the following iterative updates of the primal, auxiliary, dual variables -- $\mb{x}$, $\{ \mb{d}_a$, $\mb{d}_{\psi} \}$, and $\{ \mb{b}_a$, $\mb{b}_{\psi} \}$, respectively:
\begingroup
\allowdisplaybreaks
\ea{
\label{eq:algoAR}
 \mb{G} \mb{x}^{(i+1)} & = \mb{A}^T \Big( \mb{d}_a^{(i)} - \mb{b}_a^{(i)} \Big) + \nu \widetilde{\mb{\Psi}}^T \Big( \mb{d}_\psi^{(i)} - \mb{b}_\psi^{(i)} + \mb{z} \Big);
\\
\begin{split}
\label{eq:algoWI}
\mb{d}_a^{(i+1)} &= \Big( \mb{W} + \mu \mb{I}_m \Big)^{-1} \left( \mb{W} \mb{y} + \mu \left( \mb{A} \mb{x}^{(i+1)} + \mb{b}_a^{(i)} \right)  \right);
\end{split}
\\
d_{\psi,j}^{(i+1)} & = \cS \! \left( \left( \widetilde{\mb{\Psi}} \mb{x}^{(i+1)} - \mb{z} + \mb{b}_{\psi}^{(i)} \right)_{j}, \frac{\lambda}{\mu \nu} \right),
j = 1,\ldots, nJ;
\nn \\
\mb{b}_{a}^{(i+1)} & = \mb{b}_{a}^{(i)} - \left( \mb{d}_{a}^{(i+1)} - \mb{A} \mb{x}^{(i+1)} \right);
\nn \\
\mb{b}_{\psi}^{(i+1)} & = \mb{b}_{\psi}^{(i)} - \left(  \mb{d}_{\psi}^{(i+1)} - \left( \widetilde{\mb{\Psi}} \mb{x}^{(i+1)} - \mb{z} \right) \right),
\nn
}
\endgroup
where the Hessian matrix $\mb{G} \in \bbR^{N \times N}$ is defined by
\be{
	\label{eq:G}
	\mb{G} :=  \mb{A}^T \mb{A} + \nu  \widetilde{\mb{\Psi}}^T  \widetilde{\mb{\Psi}} ,
}
and the soft-shrinkage operator is defined by $\cS(\alpha,\beta) \!:=\! \sgn(\alpha) \max(|\alpha| - \beta, 0)$.
Similar to \cite[Fig.~1]{Ramani&Fessler:12MI}, we reset $\mb{b}_{a}$ and $\mb{b}_{\psi}$ as a zero-vector before running the ADMM image updates.
To approximately solve \R{eq:algoAR}, we use the preconditioned conjugate gradient (PCG) method with a preconditioner $\mb{M}$ for the matrix $\mb{G}$ in \R{eq:G}. 
PCG($\mb{M}$) in Algorithm~\ref{alg:PWLS-ST-l1} denotes PCG method using a preconditioner $\mb{M}$. 
Section~\ref{subsec:preconditioner} describes details of the preconditioner designs.

\subsubsection{Sparse Coding} \label{sec:Recon_spCd}

Given the current estimates of the image $\mb{x}$, we update the sparse codes $\mb{z}$ by solving the following optimization problem:
\be{
\label{eq:sparseCode}
\min_{\mb{z} \in \bbR^{n J}} ~ \lambda \left\| \widetilde{\mb{\Psi}} \mb{x} - \mb{z} \right\|_1 + \gamma \left\| \mb{z} \right\|_0.
}
The optimal solution of \R{eq:sparseCode} is given by an element-wise operator:
\be{
\label{eq:soln:sparsCode}
z_j^\star = \cH \! \left( \left( \widetilde{\mb{\Psi}} \mb{x} \right)_j , \frac{\gamma}{\lambda} \right), \quad j = 1,\ldots,nJ,
}
where the hard-shrinkage operator $\cH(\alpha,\beta)$ is defined equal to $\alpha$ for $| \alpha | \geq \beta$, and $0$ otherwise.


\subsubsection{Preconditioner Designs for Solving \R{eq:algoAR} via PCG} \label{subsec:preconditioner}

For a 2D fan-beam CT problem, a circulant preconditioner for the Hessian matrix $\mb{G}$ defined in \R{eq:G} is well suited because \textit{1)} it is effective for the \dquotes{nearly} shift-invariant matrix $\mb{A}^T \mb{A}$ \cite{Ramani&Fessler:12MI, Chun&Talavage:13Fully3D} and \textit{2)} $ \widetilde{\mb{\Psi}}^T  \widetilde{\mb{\Psi}}  = \sum_{j=1}^J \mb{P}_j^T \mb{\Psi}^T \mb{\Psi} \mb{P}_j$ is a block circulant circulant block (BCCB) matrix when we use the overlapping \dquotes{stride} $1$ and the \dquotes{wrap around} image patch assumption. For an orthogonal transform $\mb{\Psi}$, $\widetilde{\mb{\Psi}}^T  \widetilde{\mb{\Psi}}$ is approximately $(n/\iota) \mb{I}_N$, where $\iota$ denotes the stride parameter. Therefore, a circulant preconditioner is a reasonable choice to approximate $\mb{G}$ in \R{eq:G} in 2D fan-beam CT.

\begin{figure}[!t]
	\centering
	\small\addtolength{\tabcolsep}{-5pt}
	\begin{tabular}{c}
		\includegraphics[scale=0.45,clip, trim=0.1em 0.3em 2em 1.5em]{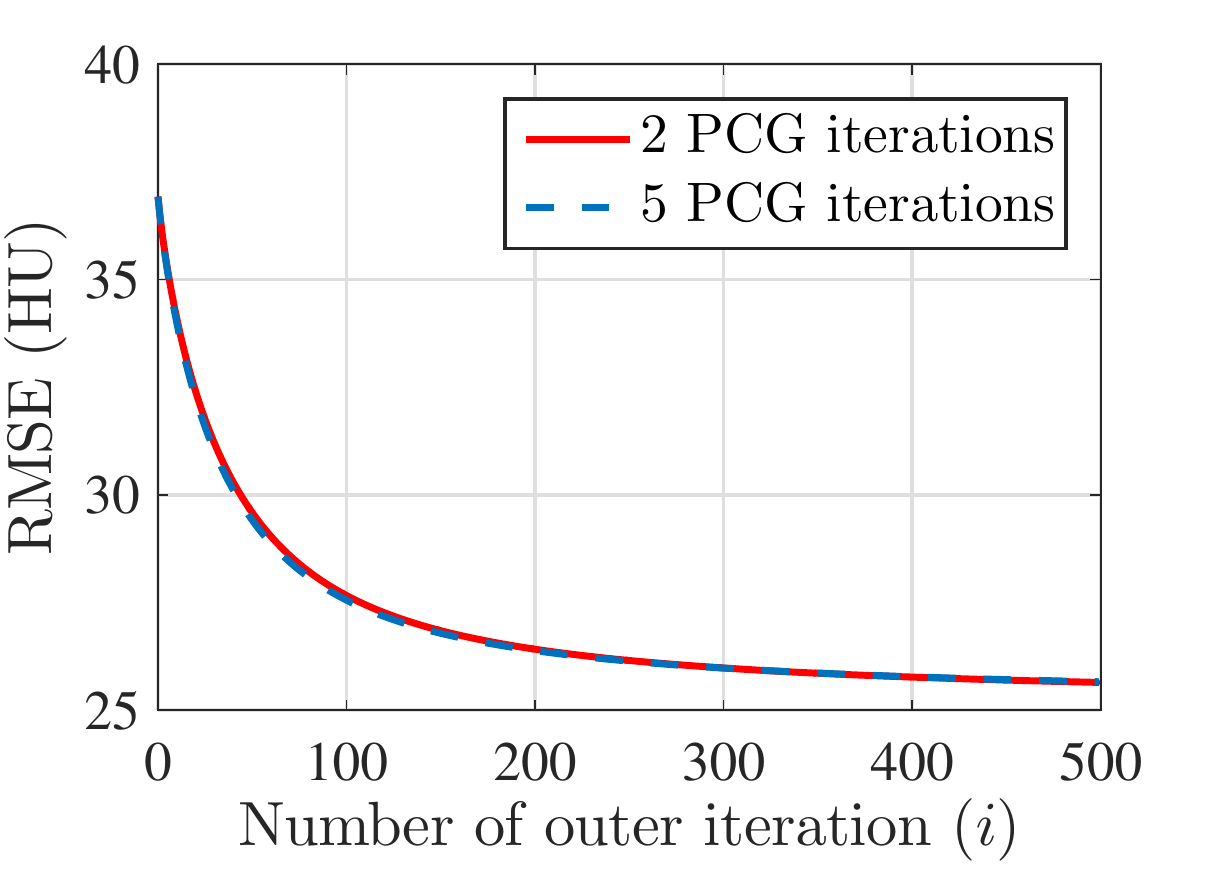} 
	\end{tabular}
	\caption{Comparison of the RMSE convergence behavior for PWLS-ST-$\ell_1$ with $2$ and $5$ PCG iterations in 3D cone-beam CT reconstruction ($12.5\%$ ($123$) views; $\lambda =  8 \!\times\! 10^6, \gamma/\lambda = 15, \kappa_{\mathrm{des},\nu} = 10, \kappa_{\mathrm{des},\mu} =40 $; RMSE stands for the root mean square error). Although a circulant preconditioner is known to be less accurate to approximate $\mb{A}^T \mb{A}$ in \R{eq:G} for the 3D CT problem, Algorithm~\ref{alg:PWLS-ST-l1} using it with $2$ PCG iterations gives a similar convergence rate compared to that with $5$ PCG iterations.}
	\label{fig:pcg}
\end{figure}

For a 3D cone-beam CT problem, circulant preconditioning is less accurate because the matrix $\mb{A}^T \mb{A}$ is inherently shift-variant due to the system geometry and/or spatial variations in detector response \cite{Fessler&Booth:99IP}. 
Despite this fact, we select a circulant preconditioner to approximate $\mb{G}$ in \R{eq:G}, and solve \R{eq:algoAR} in 3D CT reconstruction using more PCG inner iterations.
The reason is three fold. First, a circulant preconditioner is still one of the classical options to approximate a shift-variant matrix (e.g., $\mb{A}^T \mb{A}$) and accelerate CG (see, for example, \cite{Fessler&Booth:99IP, Booth&Fessler:95ICIP}).
Second, effective learned transforms are generally close to orthogonal (the same applies to the 2D case), and a scaled identity preconditioner can approximate the term $ \widetilde{\mb{\Psi}}^T  \widetilde{\mb{\Psi}} $. 
Third, a few PCG iterations in Algorithm~\ref{alg:PWLS-ST-l1} can provide fast convergence: \textit{1)} Fig.~\ref{fig:pcg} shows that $2$ and $5$ PCG iterations give very similar convergence rates; \textit{2)} in the 3D CT reconstruction, the convergence rates of Algorithm~\ref{alg:PWLS-ST-l1} are comparable to those provided in 2D CT reconstruction -- see Fig.~\ref{fig:convg}.
More sophisticated preconditioners might provide faster convergence \cite{Fu&etal:13Fully3D, Fu&etal:17Fully3D}.

Considering the reasons above, we construct a circulant preconditioner $\mb{M}$ \cite{Fessler&Booth:99IP} for the Hessian matrix $\mb{G}$ defined in \R{eq:G} as follows. 
We first approximate $\mb{G}$ by
\be{
\label{eq:Gapprox}
\mb{G} \approx \mb{Q}^H (\mb{\Lambda}_{A}  + \nu \mb{\Lambda}_{\tilde{\Psi}} ) \mb{Q},
}
where $\mb{Q}$ is the orthogonal (2D or 3D) DFT matrix, and $\mb{\Lambda}_{A}$ and $\mb{\Lambda}_{\tilde{\Psi}}$ are approximated eigenvalue matrices of $\mb{A}^T \mb{A}$ and $\widetilde{\mb{\Psi}}^T \widetilde{\mb{\Psi}}$ in $\mb{G}$.
Next, we obtain $\mb{\Lambda}_{A}$ and $\mb{\Lambda}_{\tilde{\Psi}}$ as follows:
\be{
\mb{\Lambda}_{A} =  \diag (\mathsf{fft} (\mb{A}^T \mb{A} \mb{e}_c)   ) 
\quad \mbox{and} \quad
\mb{\Lambda}_{\tilde{\Psi}} =  \diag (\mathsf{fft} (\widetilde{\mb{\Psi}}^T \widetilde{\mb{\Psi}} \mb{e}_c)   ),
\label{eq:Lambda}
}
where $\diag(\cdot)$ denotes the conversion of a vector into a diagonal matrix, $\mathsf{fft} (\cdot)$ denotes the (2D or 3D) fast Fourier transforms (FFT), and $\mb{e}_c$ is a standard basis vector corresponds to the center pixel of the image. 
Finally, we construct $\mb{M} = \mb{Q}^H (\mb{\Lambda}_{A}  + \nu\mb{\Lambda}_{\tilde{\Psi}} )^{-1} \mb{Q}$. 
For PCG($\mb{M}$) in Algorithm~\ref{alg:PWLS-ST-l1}, we use (inverse) FFTs to compute the circulant preconditioner $\mb{M}$ designed above.




\subsubsection{Parameter Selection based on Condition Numbers} \label{sec:param_select}

In practice, ADMM can require difficult parameter tuning processes for fast and stable convergence.
We moderate this problem by selecting ADMM parameters (e.g., $\nu, \mu$) based on (approximated) condition numbers \cite{Ramani&Fessler:12MI}.
Observe that, for two square Hermitian matrices $\mb{A} \succeq 0$ and $\mb{B} \succ 0$,
\be{
\label{eq:cond}
\kappa( \mb{A} + \mb{B} ) := \frac{\sigma_{\max} (\mb{A}+\mb{B})}{\sigma_{\min} (\mb{A}+\mb{B})} \leq \frac{\sigma_{\max}(\mb{A}) + \sigma_{\max}(\mb{B})}{\sigma_{\min}(\mb{A}) + \sigma_{\min}(\mb{B})},
}
by Weyl's inequality, where the notations $\kappa(\cdot)$, $\sigma_{\max}(\cdot)$, and $\sigma_{\min}(\cdot)$ denote the condition number, the largest eigenvalue, and the smallest eigenvalue of a matrix, respectively.

Applying the bound \R{eq:cond} to the Hessian matrix $\mb{G}$ defined in \R{eq:G}, we select $\nu$ by
\be{
\label{eq:kappa_nu}
\nu = \frac{\sigma_{\max} (\mb{\Lambda}_{A}) - \kappa_{\mathrm{des},\nu} \cdot \sigma_{\min} (\mb{\Lambda}_{A}) }{ \kappa_{\mathrm{des},\nu} \cdot \sigma_{\min} (\mb{\Lambda}_{\tilde{\Psi}}) - \sigma_{\max} (\mb{\Lambda}_{\tilde{\Psi}})  },
}
where $\kappa_{\mathrm{des},\nu}$ denotes the desired \dquotes{upper bounded} condition number of the matrix $\mb{G}$, and $\mb{\Lambda}_{A}$ and $\mb{\Lambda}_{\tilde{\Psi}}$ are given as in \R{eq:Lambda}.
(The eigenvalue approximation for  $\mb{A}^T \mb{A}$ and $\widetilde{\mb{\Psi}}^T \widetilde{\mb{\Psi}}$ can be improved by the power iteration used in \cite{Ramani&Fessler:12MI}, with a cost of higher computational complexity.)
Note that equality holds in \R{eq:cond} when either $\mb{A}$ or $\mb{B}$ is a scaled identity matrix. 
In other words, when the learned ST $\mb{\Psi}$ is close to orthogonal, i.e., $\widetilde{\mb{\Psi}}^T  \widetilde{\mb{\Psi}} \approx (n/\iota) \mb{I}_N $, $\kappa_{\mathrm{des},\nu}$ becomes close to the condition number of the approximated $\mb{G}$ in \R{eq:Gapprox}.
Applying the bound \R{eq:cond} to the Hessian matrix $\mb{W} + \mu \mb{I}_m$ of \R{eq:algoWI}, we select $\mu$ by
\be{
\label{eq:kappa_mu}
\mu = \frac{\sigma_{\max} (\mb{W}) - \kappa_{\mathrm{des},\mu} \cdot \sigma_{\min} (\mb{W})}{\kappa_{\mathrm{des},\mu} - 1},
}
where $\kappa_{\mathrm{des},\mu}$ denotes the desired condition number of $\mb{W} + \mu \mb{I}_m$.

The proposed ADMM parameter selection scheme using (approximated) condition numbers 
has several benefits over direct ADMM parameter tuning: 
\begin{itemize}
\item Suppose that a CT geometry, i.e., the system matrix $\mb{A}$ and a square ST $\mb{\Psi}$ are fixed.
For different measurements,
one would not need to tune the ADMM parameter $\nu$ in \R{eq:algoAR},
because the system matrix $\mb{A}$ in \R{eq:algoAR} is fixed;
however, the ADMM parameter $\mu$ in \R{eq:algoWI} requires tuning processes,
because the weighting matrix $\mb{W}$ in \R{eq:algoWI} depends on the pre-log measurements $\rhobf$. 
The proposed ADMM parameter selection scheme above can moderate the issue of tuning the ADMM parameter $\mu$
by using a desired condition number $\kappa_{\mathrm{des},\mu}$.
\item We found that 
\textit{1)} the well-tuned desired condition numbers $\kappa_{\mathrm{des},\nu}, \kappa_{\mathrm{des},\mu}$ 
in one representative CT image reconstruction case, work well in different CT datasets, including real clinical data (see details in Section~\ref{sec:param});
\textit{2)} $\kappa_{\mathrm{des},\nu}$ is robust to mild variations in CT system matrix $\mb{A}$, for example, 2D fan-beam CT and 3D axial CT scans (see details in Sections~\ref{subsubsec:2d_recon}~\&~\ref{subsubsec:3d_recon}).
\item It is more intuitive to tune the desired condition numbers, $\kappa_{\mathrm{des},\nu}$ and $\kappa_{\mathrm{des},\mu}$,
compared to directly tuning their corresponding ADMM parameters, $\nu$ and $\mu$.
We empirically found that $\kappa_{\mathrm{des},\nu}, \kappa_{\mathrm{des},\mu} \in [10,50]$ are reasonable values for fast and stable convergence of Algorithm~\ref{alg:PWLS-ST-l1}.
\end{itemize}



\subsection{Interpreting the Proposed Model~\R{sys:L1trsf}}
\label{subsec:interpret}
This section interprets the proposed PWLS-ST-$\ell_1$ reconstruction model \R{sys:L1trsf}.
Signal $\mb{x}^{(i)}$ should be sparse in the learned transform ($\widetilde{\mb{\Psi}}$)-domain. Particularly, $\widetilde{\mb{\Psi}} \mb{x}^{(i)}$ should have a few large and some small coefficients that usually correspond to local high-frequency features (e.g., edges) and noisy features, respectively.  Thresholding in the sparse coding step removes the small signal coefficients (hopefully noise) while preserving the large ones. Using the \dquotes{denoised} sparse codes $\mb{z}^{(i)}$ for the next image update, the method balances the data fidelity (i.e., $\frac{1}{2} \| \mb{y} - \mb{A} \mb{x} \|_\mb{W}^2$) and the learned $\ell_1$ prior (i.e., $\lambda \| \widetilde{\mb{\Psi}} \mb{x} - \mb{z}^{(i)} \|_1$) that is robust to the model mismatch between $\widetilde{\mb{\Psi}} \mb{x}$ and $\mb{z}^{(i)}$, $\forall i$. Repeating these processes refines the reconstructed image. 


 Given $\mb{z}^{(i)}$, the update of $\mb{x}$ in \R{sys:L1trsf} would be
\be{
\label{sys:L1trsf:x}
\mb{x}^{(i+1)} = \argmin_{\mb{x} \in \bbR^{N} } \frac{1}{2} \left\| \mb{y} - \mb{A} \mb{x} \right\|_\mb{W}^2 + \lambda \left\| \widetilde{\mb{\Psi}} \mb{x} - \mb{z}^{(i)} \right\|_1.
}
One can expect the update $\mb{x}^{(i+1)}$ to improve, as the denoised sparse codes $\mb{z}^{(i)}$ become closer to those of the true signal $\mb{x}^{\mathrm{true}}$. 
To support this argument,
we empirically calculated MSE of the following estimator:  
$
\hat{\mb{x}} = \argmin_{\mb{x} \in \bbR^{N} } \frac{1}{2} \| \mb{y} - \mb{A} \mb{x} \|_\mb{W}^2 + \lambda \| \widetilde{\mb{\Psi}} \mb{x} - \mb{z} \|_1.
$
In particular, we solved the above optimization problem by the image updating iterations in Algorithm~\ref{alg:PWLS-ST-l1} with a hundred random realizations of $\mb{z}$,
where we randomly generated $\mb{z}$ by corrupting
$\widetilde{\mb{\Psi}} \mb{x}^{\mathrm{true}}$ with three different levels 
-- $10$, $20$, and $30$dB signal-to-noise ratio (SNR) -- 
of random additive white Gaussian noise.
For $\mb{z}$ with $10$, $20$, and, $30$~dB SNR levels,
the empirical MSE values of the estimator $\hat{\mb{x}}$ were (approximately) $9.4 \times 10^7$, $1.2\times 10^7$, and $1.3\times 10^6$, respectively -- in Hounsfield units, HU\footnote{Modified Hounsfield units, where air is $0$ HU and water is $1000$ HU.}.
These empirical results support that the better the quality of $\mb{z}^{(i)}$, the more accurate $\mb{x}^{(i+1)}$ in \R{sys:L1trsf:x}.


We formally state the above intuition by relaxing the $\ell_1$-norm with a $\ell_2$-norm in \R{sys:L1trsf:x} in the following.

\prop{\label{p:mse}
Consider the following model:
\be{
\label{p:mse:sys}
\mb{y} = \mb{A} \mb{x}^{\mathrm{true}} + \bm{\varepsilon} \quad\mbox{and}\quad \mb{z}^{(i)} = \widetilde{\mb{\Psi}} \mb{x}^{\mathrm{true}} + \mb{e}^{(i)},
}
where $\mb{y} \in \bbC^m$ is a measurement vector, $\mb{A} \in \bbC^{m \times N}$ is a system matrix, $\widetilde{\mb{\Psi}} \in \bbC^{N' \times N}$ is a sparsifying transform with $N' \geq N$, $\mb{z}^{(i)} \in \bbC^{N'}$ is the denoised signal at the $(i-1)\rth$ iteration, and the noise vector $ \bm{\varepsilon} \in \bbC^m$ and error vector $ \mb{e}^{(i)} \in \bbC^{N'}$ are assumed to follow zero-mean Gaussian distribution, i.e., $\bm{\varepsilon} \sim \cN(\mb{0}, \mb{C}_\varepsilon)$ and $\mb{e}^{(i)} \sim \cN(\mb{0}, \mb{C}_{e,i})$, where $\mb{C}_\varepsilon \in \bbC^{m \times m}$ and $\mb{C}_{e,i} \in \bbC^{N' \times N'}$ are covariance matrices.
Assuming that $\bm{\varepsilon}$ and $ \mb{e}^{(i)}$ are uncorrelated, the minimum-variance unbiased estimator (MVUE) is given by 
\ea{
\mb{x}^{(i+1)} =& \left( \mb{A}^H \mb{C}_\varepsilon^{-1} \mb{A} + \widetilde{\mb{\Psi}}^H \mb{C}_{e,i}^{-1} \widetilde{\mb{\Psi}} \right)^{-1} 
\nn \\
& \cdot \left( \mb{A}^H \mb{C}_\varepsilon^{-1} \mb{y}  + \widetilde{\mb{\Psi}}^H \mb{C}_{e,i}^{-1}  \mb{z}^{(i)} \right).
\label{p:mse:soln}
}
Assuming that $\mb{A}^H \mb{A}$ and $\widetilde{\mb{\Psi}}^H \widetilde{\mb{\Psi}}$ are decomposed by some identical orthogonal matrices, and setting $\mb{C}_\varepsilon = \sigma_\varepsilon^2 \mb{I}$ and $\mb{C}_{e,i} = \sigma_{e,i}^2 \mb{I}$, the minimum variance (i.e., the MMSE for unbiased estimator)\footnote{Rigorously speaking, so called variance or MSE in our paper is the sum of pixel-wise variances or MSEs (i.e., trace of variance matrix or MSE matrices, e.g., $\tr( \mathrm{Var} (\cdot) )$). For brevity, we refer the $\tr( \mathrm{Var} (\cdot) )$ as variance.} of the solution \R{p:mse:soln} is given by 
\be{
\label{p:mse:anal}
\mathrm{var} \Big( \mb{x}^{(i+1)}  \Big) = \sum_{j=1}^N \frac{1}{ \frac{1}{ \sigma_\varepsilon^2}  (\lambda_{A})_j +  \frac{1}{ \sigma_{e,i}^2}  (\lambda_{\tilde{\Psi}})_j    },
}
where $\{ (\lambda_{A})_j \geq 0 : \forall j \}$ and $\{ (\lambda_{\tilde{\Psi}})_j \geq 0 : \forall j \}$ are the spectrum of $\mb{A}^H \mb{A}$ and $\widetilde{\mb{\Psi}}^H \widetilde{\mb{\Psi}}$, respectively.
}

Proposition~\ref{p:mse} is the first analytical result that quantifies the performance of learned analysis regularizers (e.g., learned convolutional analysis operator \cite{Chun&Fessler:18cao} and learned transform \cite{Ravishankar&Bresler:15TSP}) in signal recovery.
When the $\ell_1$-norm is relaxed with a $\ell_2$-norm (and setting $\mb{W} = \mb{C}_\varepsilon^{-1}$ and $\lambda = 1/\sigma_{e,i}^2$), the MVUE solution in \R{p:mse:soln} with $\mb{C}_{e,i} = \sigma_{e,i}^2 \mb{I}$ corresponds to that of the image update problem \R{sys:L1trsf:x}.
The assumption of uncorrelated $ \bm{\varepsilon}$ and $ \mb{e}^{(i)}$ is satisfied, if the noise in the measurement domain and the error in the $\widetilde{\mb{\Psi}}$-domain are uncorrelated.

For any $\mb{A}$ and $\widetilde{\mb{\Psi}}$, the minimum variance $\mathrm{var} ( \mb{x}^{(i+1)} )$ in \R{p:mse:anal} can be further reduced as the error variance $\sigma_{e,i}^2$ becomes smaller, for some fixed $\sigma_\varepsilon^2 \in (0, \infty)$. For example, if $\sigma_{e,i}^2 \rightarrow \epsilon$, where $0< \epsilon \ll 1$, then $\mathrm{var} (\mb{x}^{(i+1)})$ becomes very small. 
If the \dquotes{denoised} sparse codes $\mb{z}^{(i)}$ become close to $\widetilde{\mb{\Psi}} \mb{x}^{\mathrm{true}}$ as $i \rightarrow \infty$, one obtains accurate image reconstruction after sufficiently iterating the updates for \R{sys:L1trsf} in Algorithm~\ref{alg:PWLS-ST-l1}.
To better \dquotes{denoise} the update $\mb{x}^{(i+1)}$ in particular, we pre-learn a ST $\mb{\Psi}$ via \R{eq:l1LearnSpTrasf} from high-quality training datasets.

\section{Results and Discussions}
\label{sec:results}

\subsection{Experimental Setup} \label{sec:exp}

We evaluated the proposed PWLS-ST-$\ell_1$ method for sparse-view CT reconstruction with 2D fan-beam and 3D axial cone-beam scans of a XCAT phantom that has overall $500$ slices \cite{Segars&etal:08MP}.
We also evaluated PWLS-ST-$\ell_1$ for sparse-view CT reconstruction with 2D fan-beam real GE clinical data. 
We compared the quality of images reconstructed by PWLS-ST-$\ell_1$ with those of: 
	\begin{itemize}
	\item \emph{FBP}: Conventional filtered back-projection method using a Hanning window (for 3D experiments, the Feldkamp-Davis-Kress method \cite{Feldkamp&David&Kress:84JOSAA} was used).
	\item \emph{PWLS-EP}: Conventional MBIR method using PWLS and an edge-preserving regularizer $\sum_{j  =1}^{N} \sum_{k\in N_{j}}\iota_{j} \iota_{k} \varphi(x_j - x_k)$, where $N_j$ is the set of neighbors of $x_j$, $\iota_j$ and $\iota_k$ are regularization parameters that encourage uniform noise \cite{Cho&Fessler:15TMI}, and ${\varphi}{(t)}:= \delta^2\left(  \sqrt{1+| t/\delta |^2 }  -1  \right)$ for 2D, ${\varphi}{(t)}:= \delta^2 (  | t/\delta | - \log(1+| t/\delta |) )$ for 3D ($\delta \!=\! 10$ HU). 
	We adopted the relaxed linearized augmented Lagrangian method with ordered-subsets (relaxed OS-LALM) proposed in \cite{Nien&Fessler:16:TMI} to accelerate the reconstruction.
	\item \emph{PWLS-ST-$\ell_2$} (Zheng et al., 2018): MBIR method using PWLS and $\ell_2$ prior with a learned ST \cite{Zheng&etal:16IVMSP, Zheng&etal:17arXiv}. For fair comparison, we performed the image update of the algorithm proposed for PWLS-ST-$\ell_2$ without the non-nonnegativity constraint.
	\item \emph{PWLS-DL} (Xu et al., 2012): MBIR method using PWLS and $\ell_2$ prior with a learned overcomplete synthesis dictionary \cite{Xu&etal:12TMI}.
	We replaced the separable quadratic surrogate method with ordered-subsets based acceleration (SQS-OS) in \cite{Xu&etal:12TMI} with relaxed OS-LALM to accelerate image updates.
	For fair comparison, we ran this method without the non-nonnegativity constraint.
 PWLS-DL is far slower for 3D reconstruction with large 3D patches, compared to 2D reconstruction \cite{Zheng&etal:17arXiv}; thus, we focus our comparisons between PWLS-ST-$\ell_1$ and PWLS-DL for 2D reconstruction.
   	\item \emph{FBPConvNet} (Jin et al., 2017): A non-MBIR \dquotes{denoising} method whose network structure is modified from U-Net \cite{Jin&etal:17TIP}.  As suggested in \cite{Jin&etal:17TIP}, we trained the network by minimizing the $\ell_2$-based training loss function that used paired training images -- specifically, pairs of ground truth images and their noisy versions reconstructed by applying FBP to (simulated) undersampled sinograms.
	\end{itemize}

We quantitatively evaluated the reconstruction quality in phantom experiments by RMSE (in HU) in a region of interest (ROI). The $\textmd{RMSE}$ is defined by $\mathrm{RMSE} \!:=\! ( \sum_{j=1}^{N_{\mathrm{ROI}}}(\hat{x}_j-x_j^{\text{true}})^2/{N_{\mathrm{ROI}}} )^{1/2}$, where $\hat{\mb{x}}$ is the reconstructed image (after clipping negative values), $\mb{x}^{\text{true}}$ is the ground truth image, and $N_{\mathrm{ROI}}$ is the number of pixels in a ROI.

\subsubsection{2D Fan-Beam - Imaging} \label{sec:exp:fan:imaging}

To avoid an inverse crime, our 2D imaging simulation used a $840 \!\times\! 840 $ slice (air cropped, $\Delta_x \!=\! \Delta_y \!=\! 0.4883$ mm) of the XCAT phantom, which was different from the training slices.
We simulated sinograms of size $888$ (detector channels) $\times \{$$246$, $123$\} (regularly spaced projection views or angles; $984$ is the number of full views) with GE LightSpeed fan-beam geometry corresponding to a monoenergetic source with $\rho_0 = 10^5$ incident photons per ray and no background events, and electronic noise variance $\sigma^2 = 5^2$. We reconstructed a $420  \!\times\! 420$ image with a coarser grid, where $\Delta_x \!=\! \Delta_y \!=\! 0.9766$ mm. The ROI here was a circular (around center) region containing all the phantom tissues.

The clinical chest data was collected by the GE scanner using the same CT geometry described above.
We reconstructed a $716 \!\times\! 716$ image with $\Delta_x \!=\! \Delta_y \!=\! 0.9777$ mm. The tube voltage and tube current were $120$ kVp and $160$ mA, respectively.

\subsubsection{2D Fan-Beam - Training}
\label{subsubsec:2d_training}
Before executing reconstructions with the PWLS-ST-$\ell_1$, PWLS-ST-$\ell_2$, PWLS-DL, and FBPConvNet methods, we pre-learned or trained their priors or networks from training data.
For the PWLS-ST-$\ell_1$ and PWLS-ST-$\ell_2$ methods, we learned square ($64 \!\times\! 64$) STs from $8 \!\times\! 8$ image patches extracted from five different slices of the XCAT phantom (with $1 \!\times\! 1$ overlapping stride). To learn well-conditioned transforms, we chose a large enough $\tau$, e.g., $\tau \!=\! 5.85 \!\times\! 10^{14}$. We chose $\gamma' = 110$ and $\xi = 1$. Initialized with the 2D discrete cosine transform (DCT), we ran $1000$ iterations of the alternating minimization algorithm proposed in \cite{Ravishankar&Bresler:15TSP} to ensure learned transforms were well converged. 
For PWLS-DL, we learned a $64 \!\times\! 256$-sized overcomplete dictionary from the same set of $8 \!\times\! 8$-sized patches used in learning square STs (see above). We used a maximum patch-wise sparsity level of $20$ and a sparse coding error threshold of $10^{-1}$.
In FBPConvNet training, we used $390$ paired images for training (each image corresponded to a slice of the XCAT phantom). Note that the testing phantom image is sufficiently different from training phantom images (specifically, they are at least $3.3$cm away from training images).
We used FBP-reconstructed images from the sparse-view sinograms simulated in Section~\ref{sec:exp:fan:imaging} and the ground truth images of the XCAT phantom (with no noise), as training pairs.
We trained networks using the data augmentation stratergy and optimization method (i.e., stochastic gradient descent method) suggested in \cite{Jin&etal:17TIP}.
We set training hyperparameters (similar to those used in \cite{Jin&etal:17TIP}) as follows: $151$ epochs; learning rate decreased logarithmically from $10^{-2}$ to $10^{-3}$; batch size of $1$; \dquotes{momentum} parameter $0.99$; and the clipping value for gradient of $10^{-2}$.

 \begin{figure*}[!t]
 	\centering
 	\small\addtolength{\tabcolsep}{-7.5pt}

 	\begin{tabular}{ccccc}
 		
 		{} & \small{PWLS-EP} & \small{PWLS-DL (Xu et al., 2012)} & \small{PWLS-ST-$\ell_2$ (Zheng et al., 2018)} & \small{Proposed PWLS-ST-$\ell_1$} \\
 		
 		\raisebox{-.5\height}{\begin{turn}{+90} \small{$25$\% ($246$) views} \end{turn}}~ &
 		\raisebox{-.5\height}{
 			\begin{tikzpicture}
 			\begin{scope}[spy using outlines={rectangle,yellow,magnification=1.6,size=18mm}]
 			\node {\includegraphics[scale=0.56]{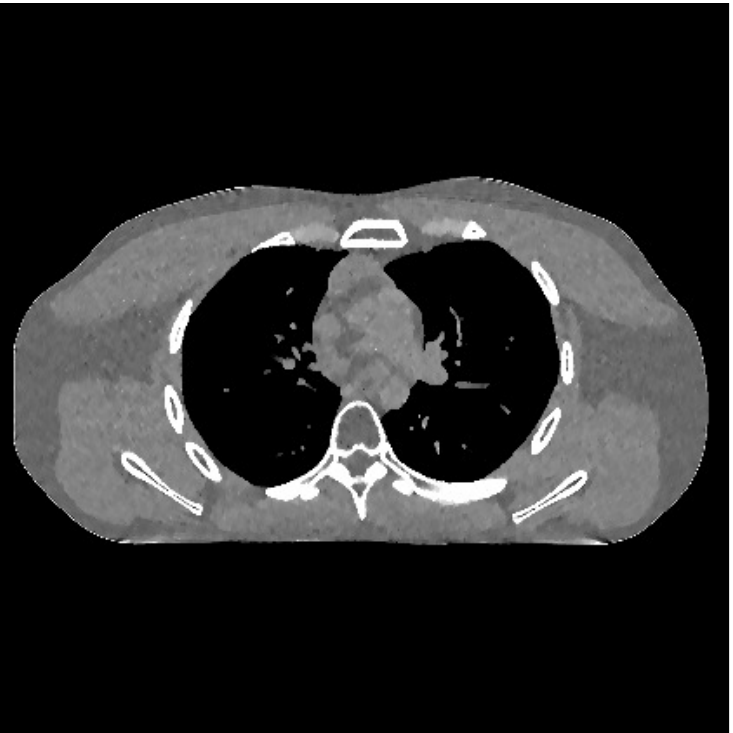}};
 			\spy on (-0.1,0.34) in node [left] at (-0.41,1.72);
 			\node [white] at (1,-1.85) {\small $\mathrm{RMSE} = 30.7$};
 			\end{scope}
 			\end{tikzpicture}} &
 		\raisebox{-.5\height}{
 			\begin{tikzpicture}
 			\begin{scope}[spy using outlines={rectangle,yellow,magnification=1.6,size=18mm}]
 			\node {\includegraphics[scale=0.56]{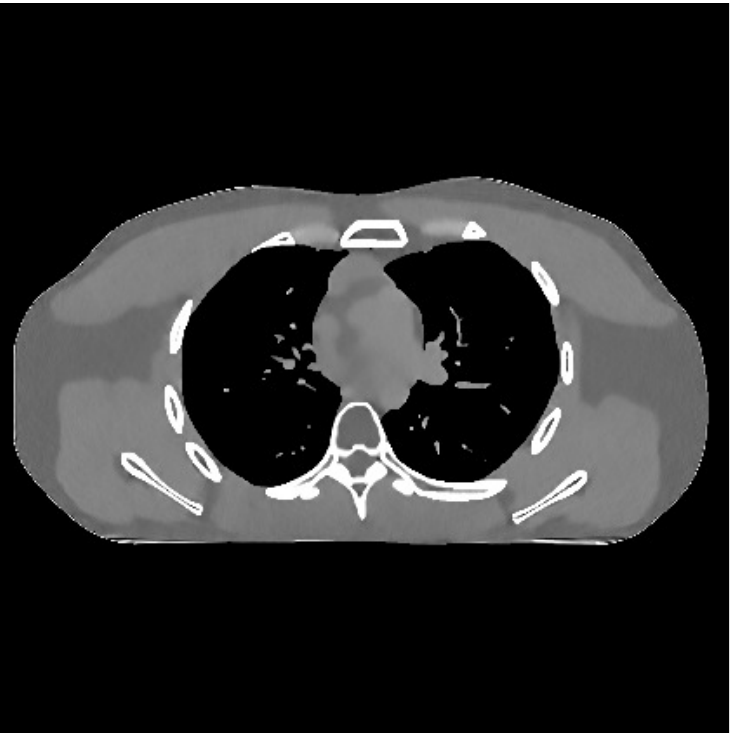}};
 			\spy on (-0.1,0.34) in node [left] at (-0.41,1.72);
 			\node [white] at (1,-1.85) {\small $\mathrm{RMSE} = 24.9$};
 			\end{scope}
 			\end{tikzpicture}} &
 		\raisebox{-.5\height}{
 			\begin{tikzpicture}
 			\begin{scope}[spy using outlines={rectangle,yellow,magnification=1.6,size=18mm}]
 			\node {\includegraphics[scale=0.56]{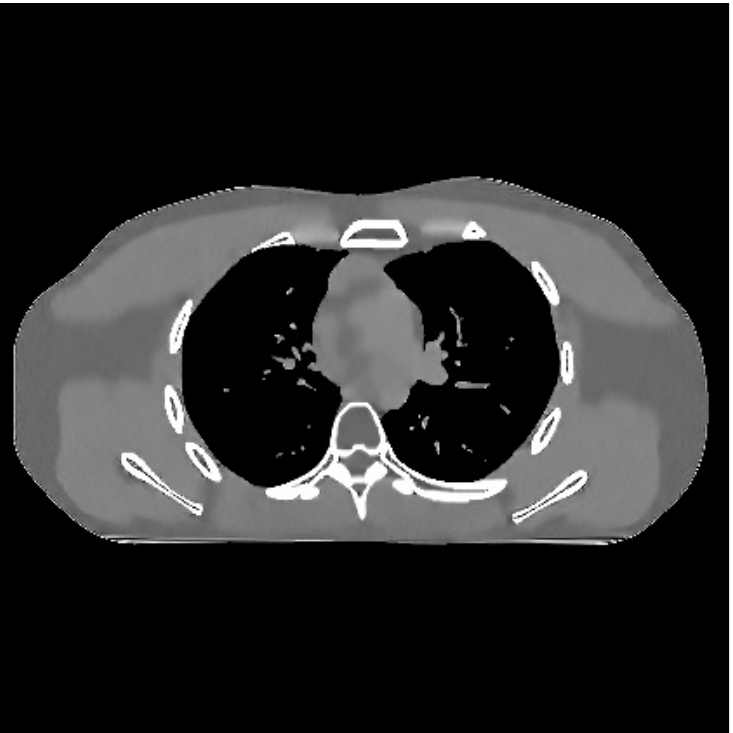}};
 			\spy on (-0.1,0.34) in node [left] at (-0.41,1.72);
 			\node [white] at (1,-1.85) {\small $\mathrm{RMSE} = 26.9$};
 			\end{scope}
 			\end{tikzpicture}} &
 		\raisebox{-.5\height}{
 			\begin{tikzpicture}
 			\begin{scope}[spy using outlines={rectangle,yellow,magnification=1.6,size=18mm}]
 			\node {\includegraphics[scale=0.56]{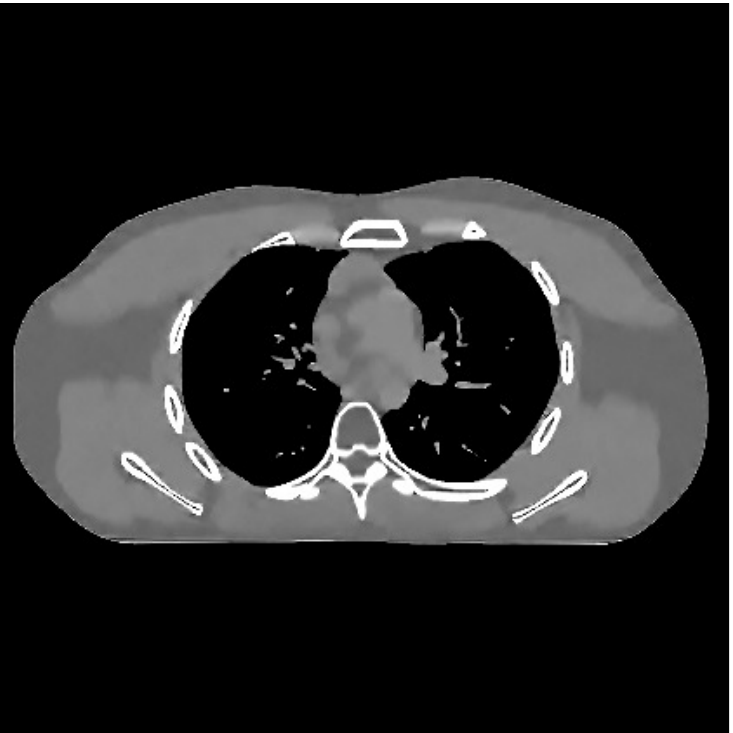}};
 			\spy on (-0.1,0.34) in node [left] at (-0.41,1.72);
 			\node [white] at (1,-1.85) {\small \color{yellow}{$\mathrm{RMSE} = 21.5$}};
 			\end{scope}
 			\end{tikzpicture}} \\

 		\raisebox{-.5\height}{\begin{turn}{+90} \small{$12.5$\% ($123$) views} \end{turn}}~ &
 		\raisebox{-.5\height}{
 			\begin{tikzpicture}
 			\begin{scope}[spy using outlines={rectangle,yellow,magnification=1.6,size=18mm,connect spies}]
 			\node {\includegraphics[scale=0.56]{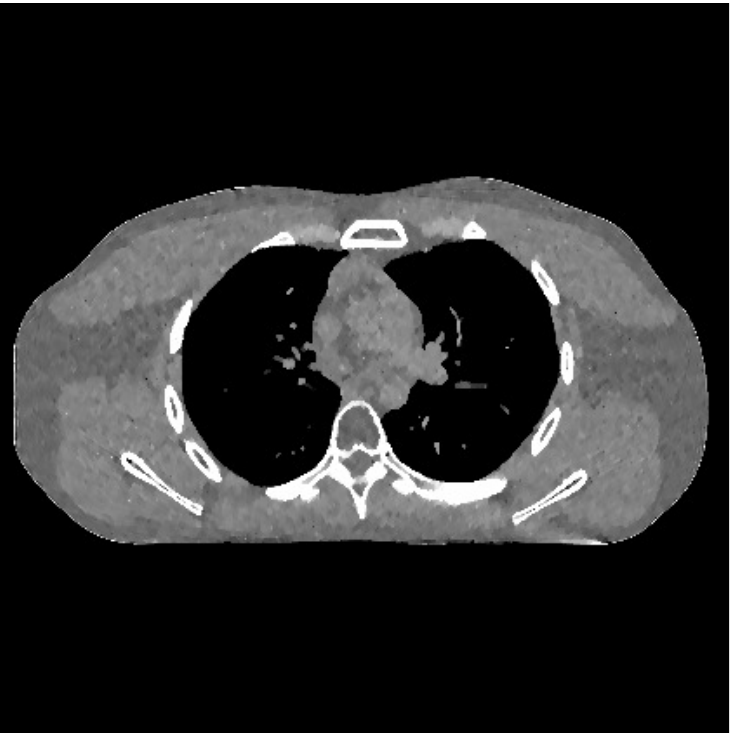}};
 			\spy on (-0.1,0.34) in node [left] at (-0.41,1.8);
 			\spy on (1.5,-0.34) in node [left] at (-0.41,-1.62);
 			\node [white] at (1,-1.85) {\small $\mathrm{RMSE} = 35.0$};
 			\end{scope}
 			\end{tikzpicture}} &
 		\raisebox{-.5\height}{
 			\begin{tikzpicture}
 			\begin{scope}[spy using outlines={rectangle,yellow,magnification=1.6,size=18mm,connect spies}]
 			\node {\includegraphics[scale=0.56]{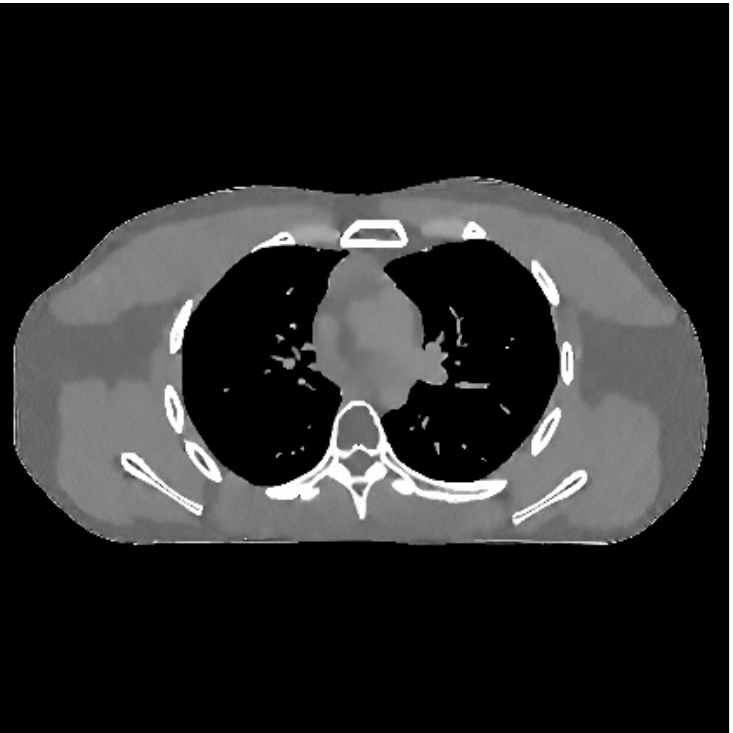}};
 			\spy on (-0.1,0.34) in node [left] at (-0.41,1.8);
 			\spy on (1.5,-0.34) in node [left] at (-0.41,-1.62);
 			\node [white] at (1,-1.85) {\small $\mathrm{RMSE} = 26.9$};
 			\end{scope}
 			\end{tikzpicture}} &
 		\raisebox{-.5\height}{
 			\begin{tikzpicture}
 			\begin{scope}[spy using outlines={rectangle,yellow,magnification=1.6,size=18mm,connect spies}]
 			\node {\includegraphics[scale=0.56]{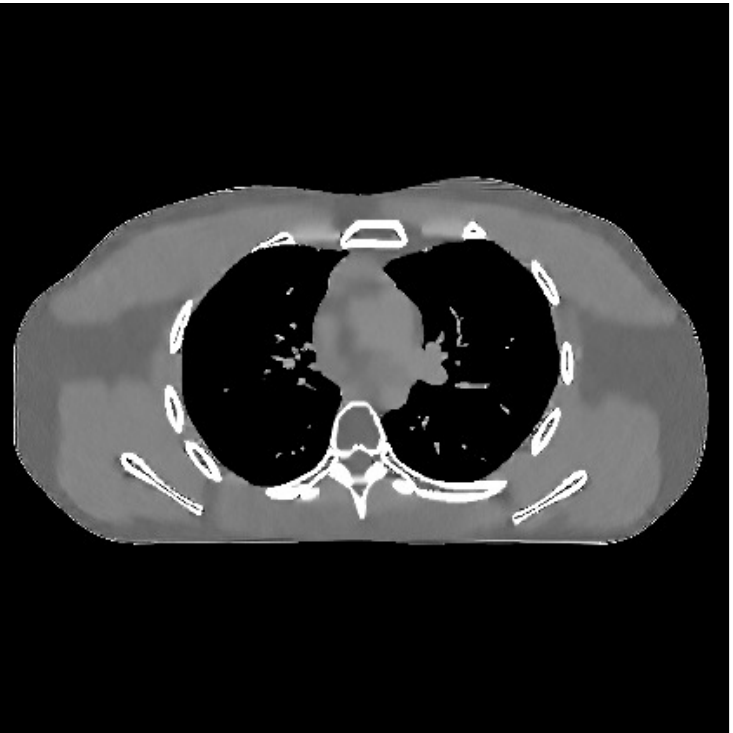}};
 			\spy on (-0.1,0.34) in node [left] at (-0.41,1.8);
 			\spy on (1.5,-0.34) in node [left] at (-0.41,-1.62);
 			\node [white] at (1,-1.85) {\small $\mathrm{RMSE} = 30.9$};
 			\end{scope}
 			\end{tikzpicture}} &
 		\raisebox{-.5\height}{
 			\begin{tikzpicture}
 			\begin{scope}[spy using outlines={rectangle,yellow,magnification=1.6,size=18mm,connect spies}]
 			\node {\includegraphics[scale=0.56]{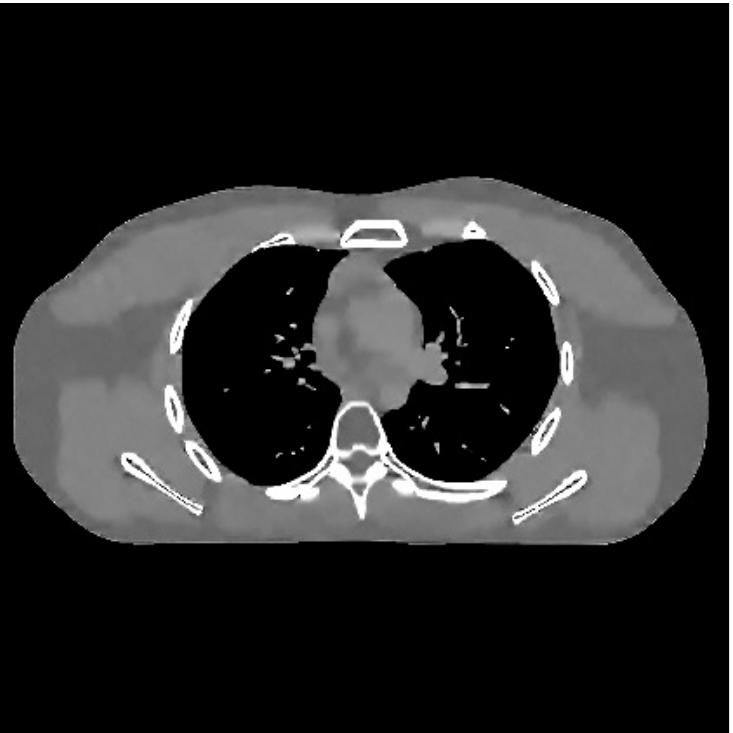}};
 			\spy on (-0.1,0.34) in node [left] at (-0.41,1.8);
 			\spy on (1.5,-0.34) in node [left] at (-0.41,-1.62);
 			\node [white] at (1,-1.85) {\small \color{yellow}{$\mathrm{RMSE} = 25.8$}};
 			\end{scope}
 			\end{tikzpicture}}
 		\\
 		
 		&  \multicolumn{4}{c}{(a) XCAT phantom data}	\vspace{0.1in}
 		\\
 		
 		\raisebox{-.5\height}{\begin{turn}{+90} \small{$25$\% ($246$) views} \end{turn}}~ &
 		
 		\raisebox{-.5\height}{
 			\begin{tikzpicture}
 			\begin{scope}[spy using outlines={rectangle,yellow,magnification=1.9,size=8mm, connect spies}]
 			\node {\includegraphics[width=0.23\textwidth]{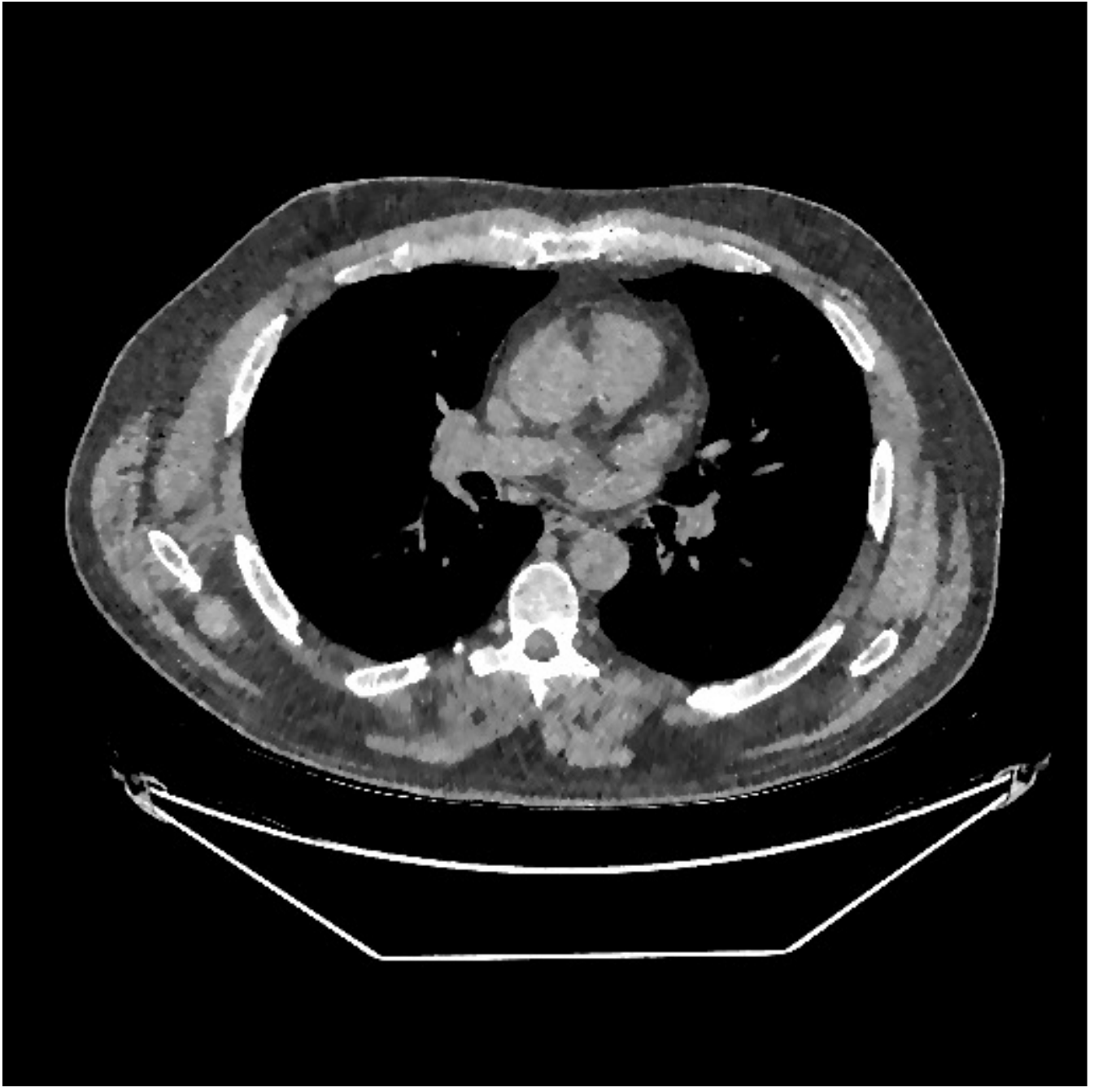}};
 			\spy on (-0.65,1.1) in node [left] at (-1.2,1.65);
 			\spy on (0.12,1.15) in node [left] at (2,1.65);
 			\spy on (-1.2,-0.4) in node [left] at (-1.2,-1.65);			
 			\spy on (0.22,-0.75) in node [left] at (2,-1.65);
 			\end{scope}
 			\end{tikzpicture}} &
 		
 		\raisebox{-.5\height}{
 			\begin{tikzpicture}
 			\begin{scope}[spy using outlines={rectangle,yellow,magnification=1.9,size=8mm, connect spies}]
 			\node {\includegraphics[width=0.23\textwidth]{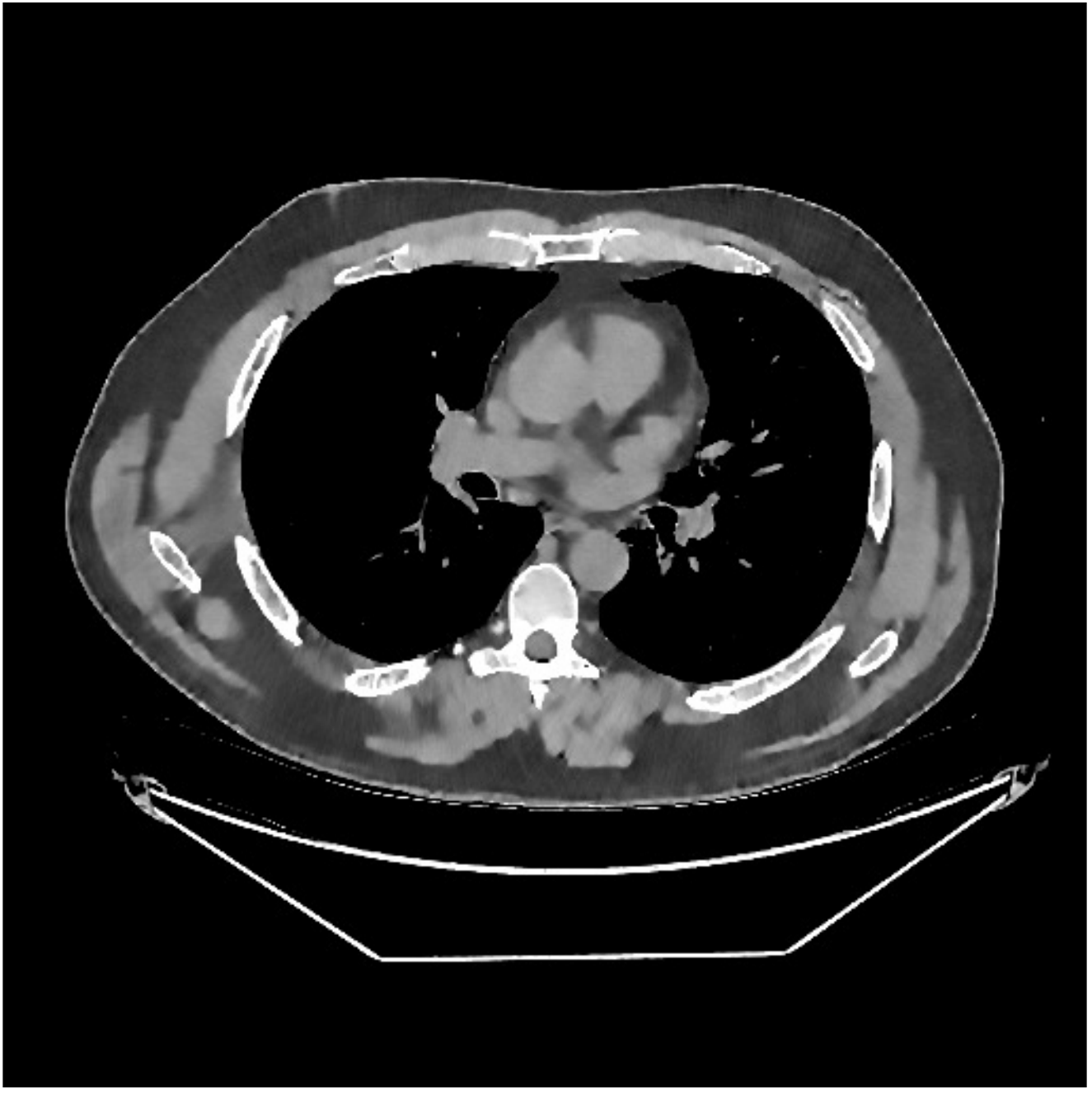}};
 			\spy on (-0.65,1.1) in node [left] at (-1.2,1.65);
 			\spy on (0.12,1.15) in node [left] at (2,1.65);
 			\spy on (-1.2,-0.4) in node [left] at (-1.2,-1.65);			
 			\spy on (0.22,-0.75) in node [left] at (2,-1.65);
 			\end{scope}
 			\end{tikzpicture}} &
 		
 		\raisebox{-.5\height}{
 			\begin{tikzpicture}
 			\begin{scope}[spy using outlines={rectangle,yellow,magnification=1.9,size=8mm, connect spies}]
 			\node {\includegraphics[width=0.23\textwidth]{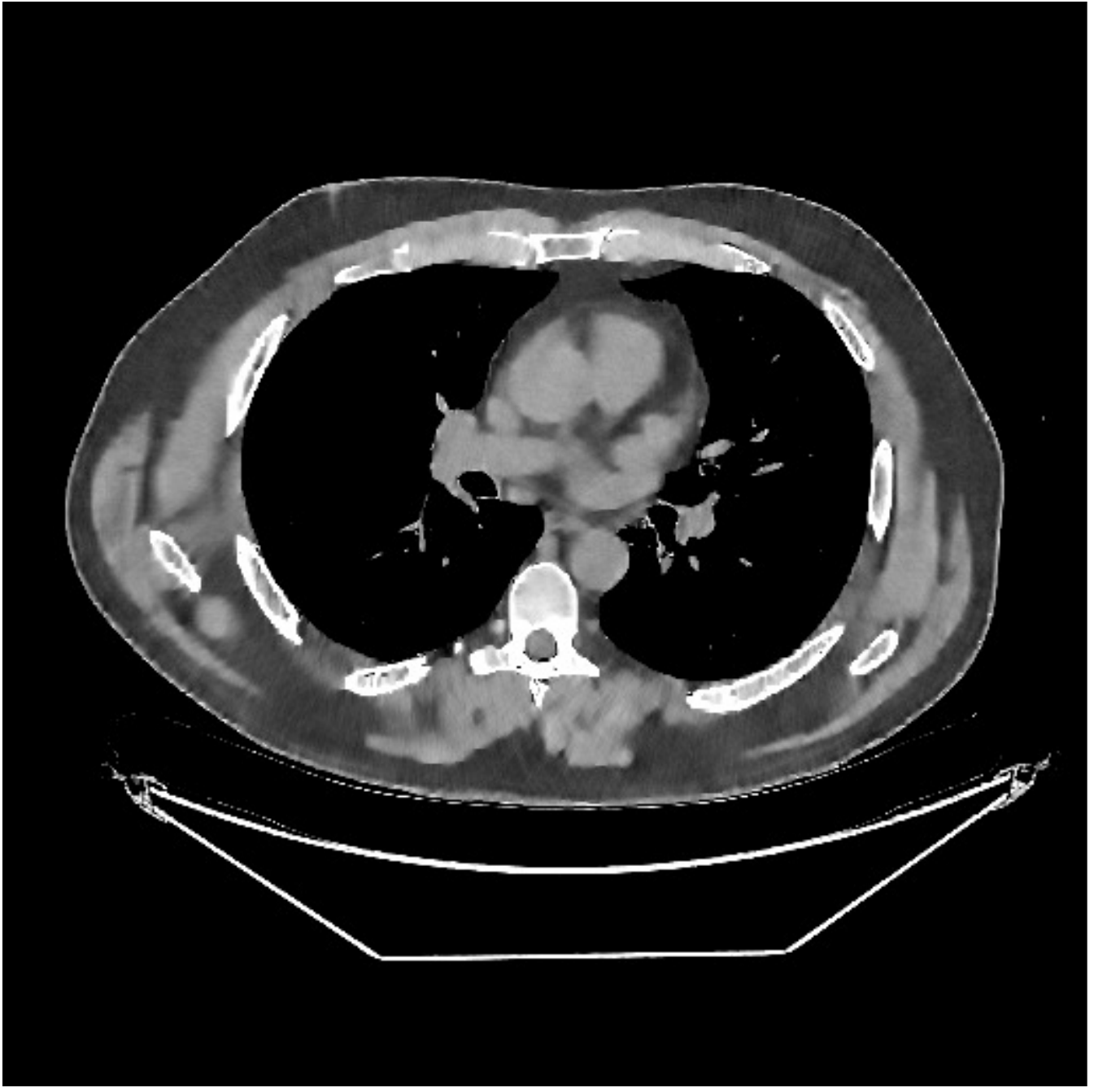}};
 			\spy on (-0.65,1.1) in node [left] at (-1.2,1.65);
 			\spy on (0.12,1.15) in node [left] at (2,1.65);
 			\spy on (-1.2,-0.4) in node [left] at (-1.2,-1.65);			
 			\spy on (0.22,-0.75) in node [left] at (2,-1.65);
 			\end{scope}
 			\end{tikzpicture}} &
 		
 		\raisebox{-.5\height}{
 			\begin{tikzpicture}
 			\begin{scope}[spy using outlines={rectangle,yellow,magnification=1.9,size=8mm, connect spies}]
 			\node {\includegraphics[width=0.23\textwidth]{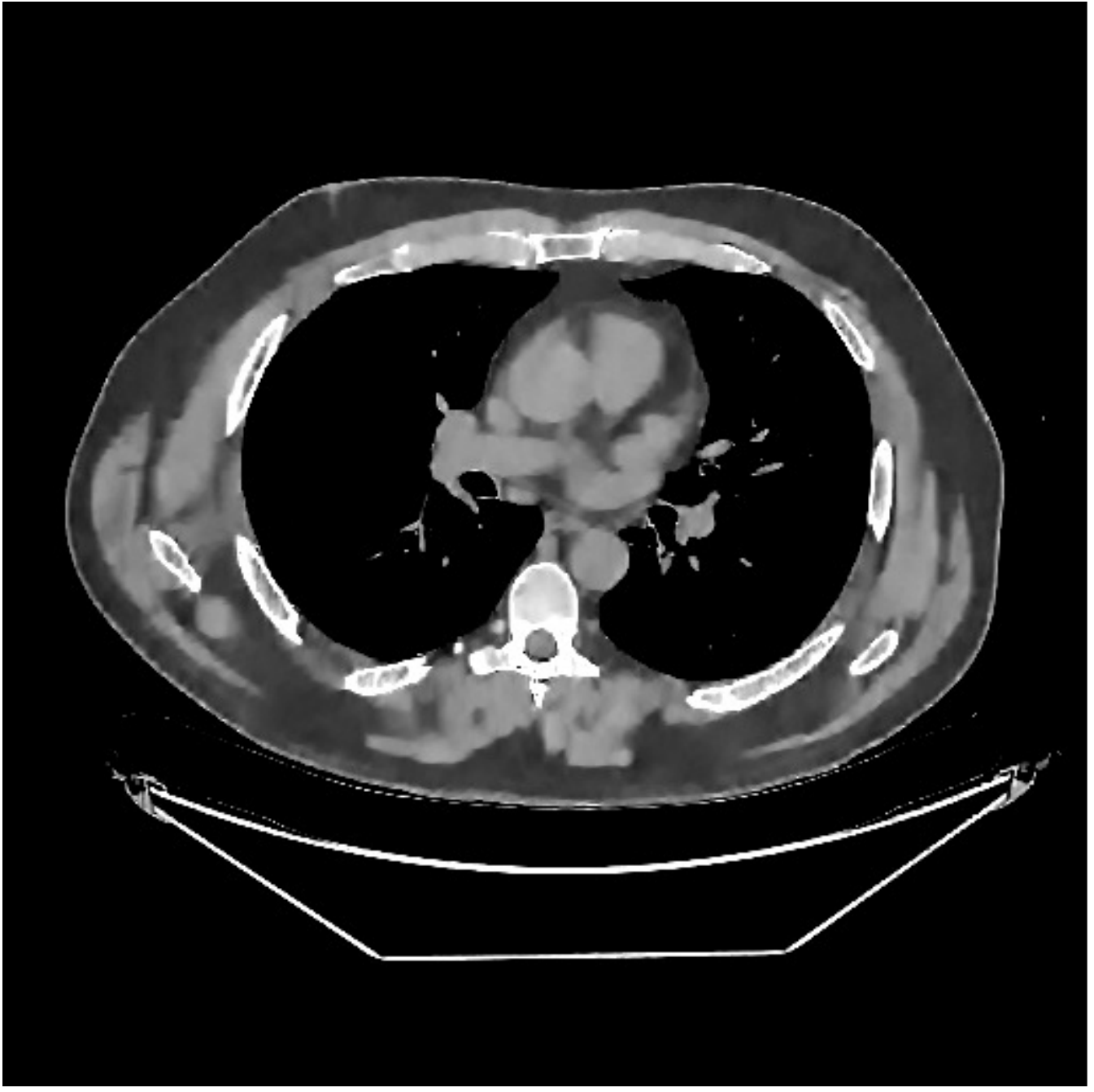}};
 			\spy on (-0.65,1.1) in node [left] at (-1.2,1.65);
 			\spy on (0.12,1.15) in node [left] at (2,1.65);
 			\spy on (-1.2,-0.4) in node [left] at (-1.2,-1.65);			
 			\spy on (0.22,-0.75) in node [left] at (2,-1.65);
 			\end{scope}
 			\end{tikzpicture}} \\
 		
 		\raisebox{-.5\height}{\begin{turn}{+90} \small{$12.5$\% ($123$) views} \end{turn}}~ &

 		\raisebox{-.5\height}{
 			\begin{tikzpicture}
 			\begin{scope}[spy using outlines={rectangle,yellow,magnification=1.9,size=8mm, connect spies}]
 			\node {\includegraphics[width=0.23\textwidth]{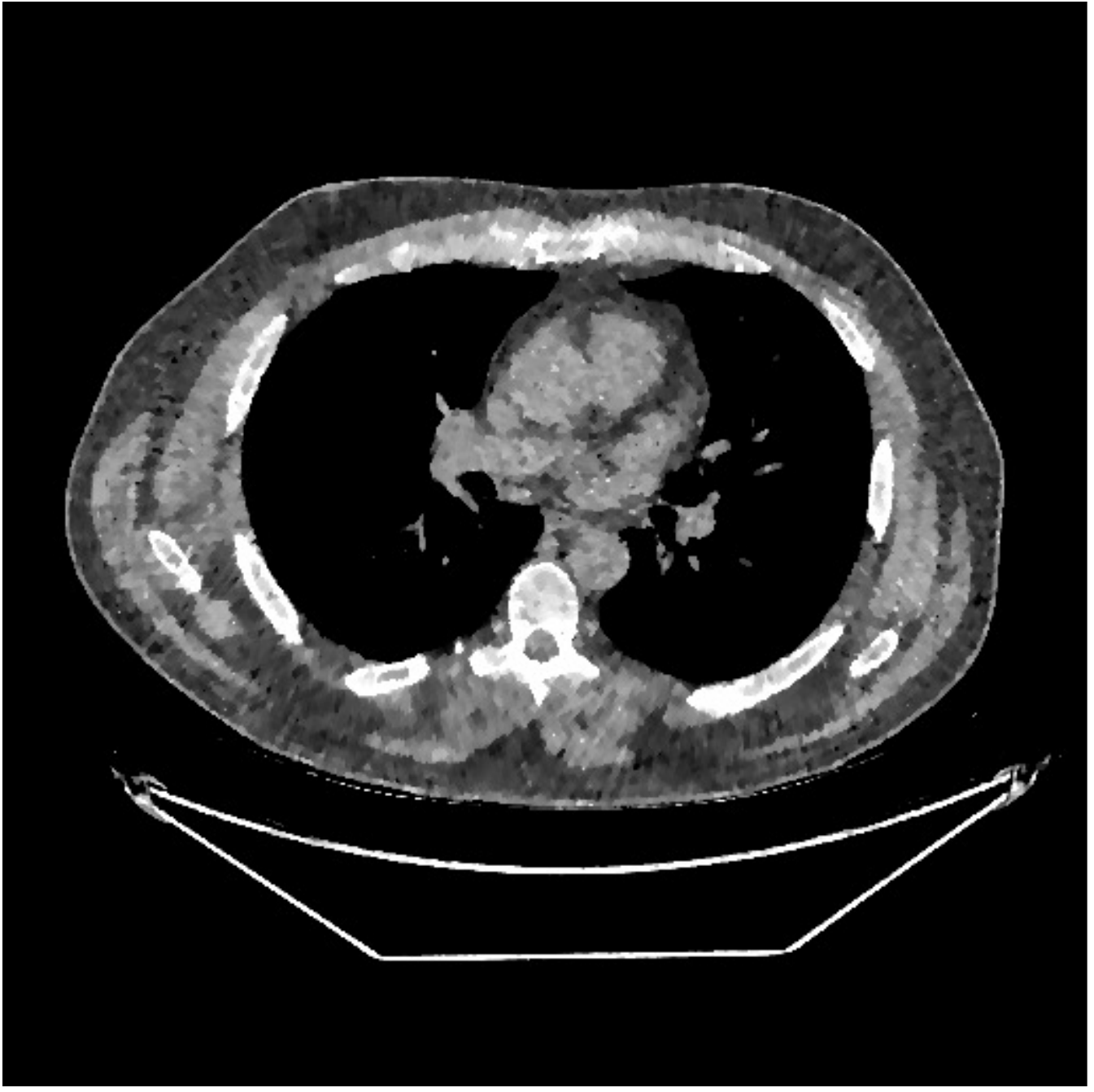}};
 			\spy on (-0.65,1.1) in node [left] at (-1.2,1.65);
 			\spy on (0.12,1.15) in node [left] at (2,1.65);
 			\spy on (-1.2,-0.4) in node [left] at (-1.2,-1.65);			
 			\spy on (0.22,-0.75) in node [left] at (2,-1.65);
 			\end{scope}
 			\end{tikzpicture}} &
 		
 		\raisebox{-.5\height}{
 			\begin{tikzpicture}
 			\begin{scope}[spy using outlines={rectangle,yellow,magnification=1.9,size=8mm, connect spies}]
 			\node {\includegraphics[width=0.23\textwidth]{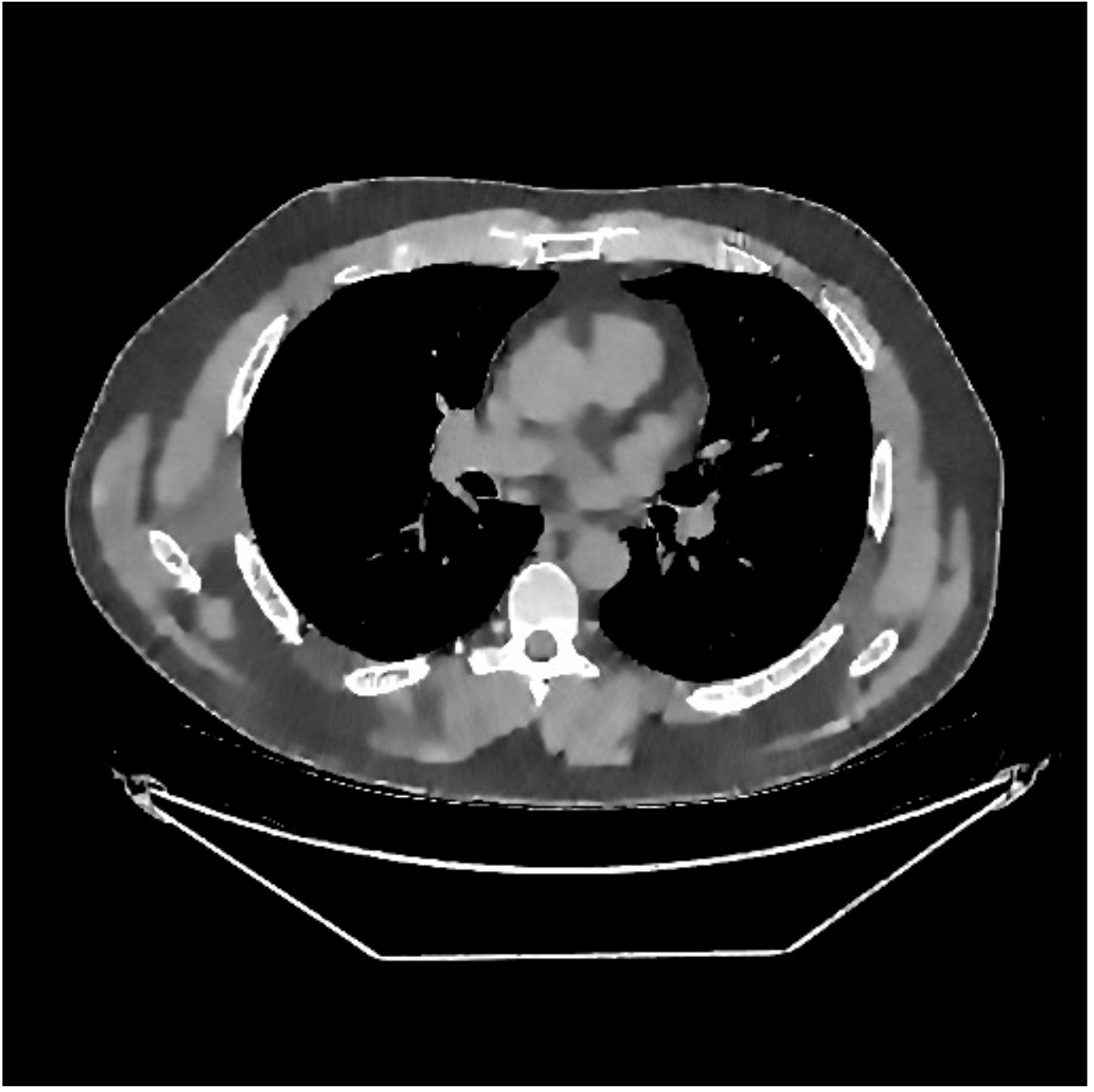}};
 			\spy on (-0.65,1.1) in node [left] at (-1.2,1.65);
 			\spy on (0.12,1.15) in node [left] at (2,1.65);
 			\spy on (-1.2,-0.4) in node [left] at (-1.2,-1.65);			
 			\spy on (0.22,-0.75) in node [left] at (2,-1.65);
 			\end{scope}
 			\end{tikzpicture}} &
 		
 		\raisebox{-.5\height}{
 			\begin{tikzpicture}
 			\begin{scope}[spy using outlines={rectangle,yellow,magnification=1.9,size=8mm, connect spies}]
 			\node {\includegraphics[width=0.23\textwidth]{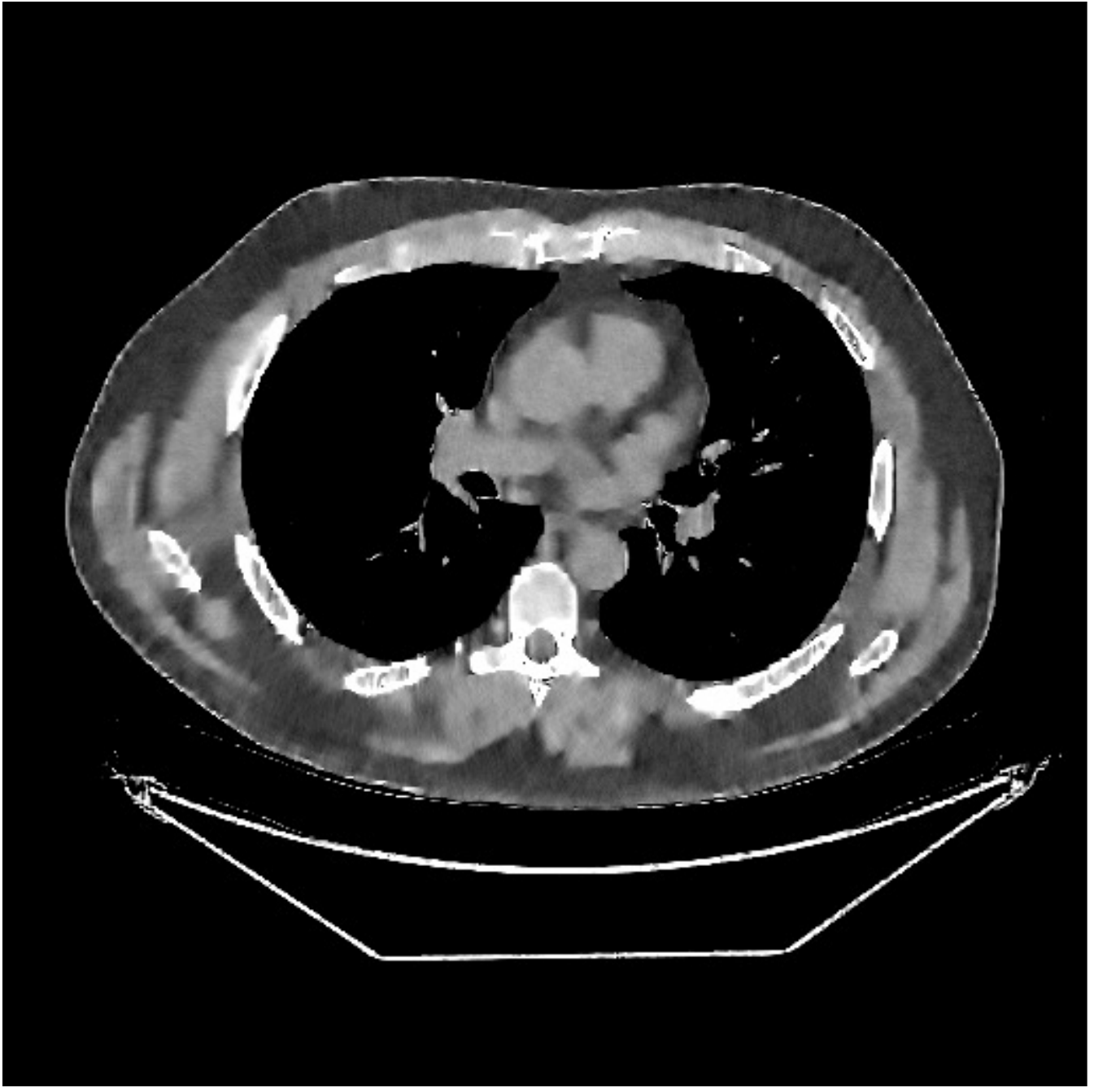}};
 			\spy on (-0.65,1.1) in node [left] at (-1.2,1.65);
 			\spy on (0.12,1.15) in node [left] at (2,1.65);
 			\spy on (-1.2,-0.4) in node [left] at (-1.2,-1.65);			
 			\spy on (0.22,-0.75) in node [left] at (2,-1.65);
 			\end{scope}
 			\end{tikzpicture}} &
 		
 		\raisebox{-.5\height}{
 			\begin{tikzpicture}
 			\begin{scope}[spy using outlines={rectangle,yellow,magnification=1.9,size=8mm, connect spies}]
 			\node {\includegraphics[width=0.23\textwidth]{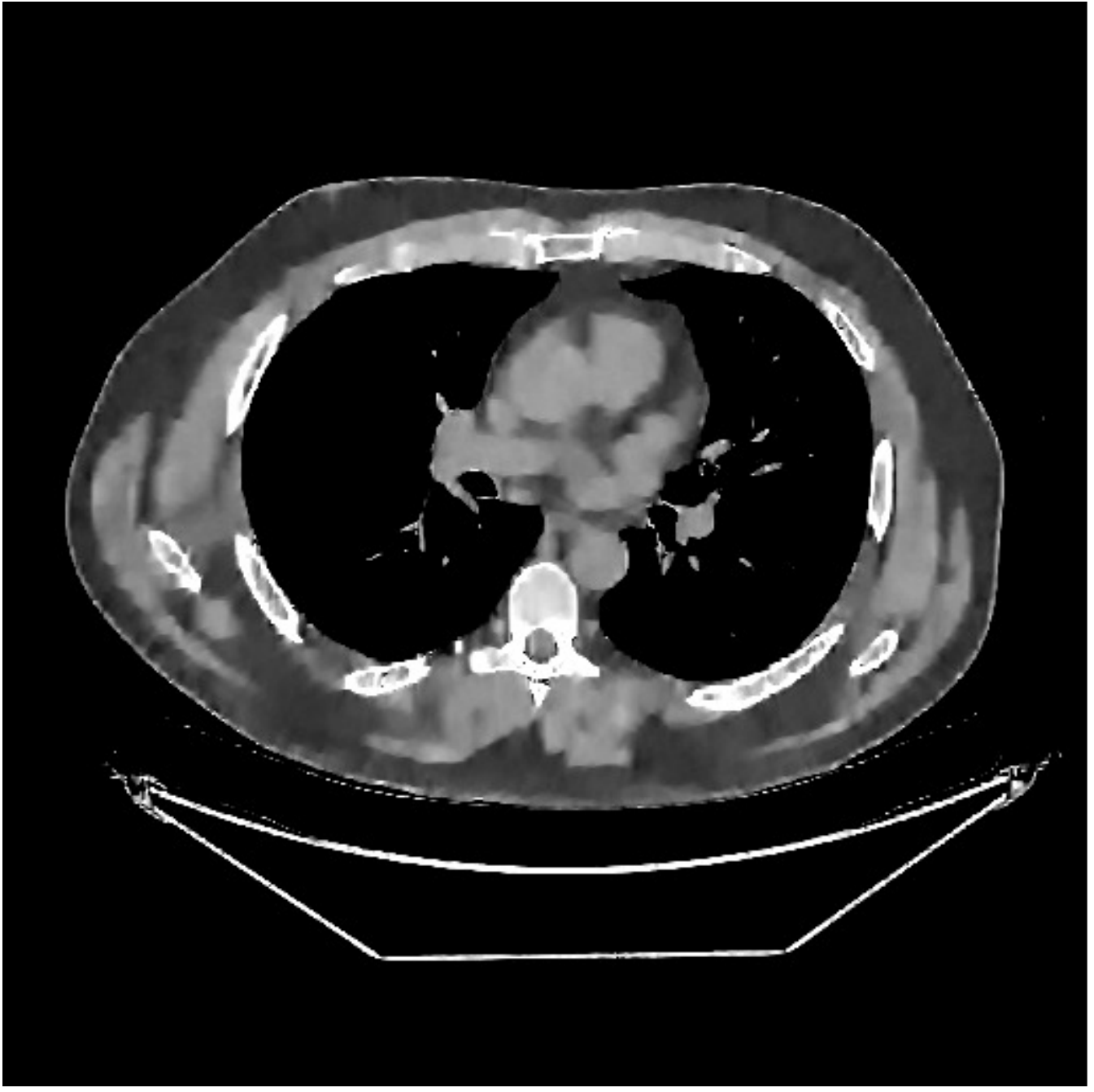}};
 			\spy on (-0.65,1.1) in node [left] at (-1.2,1.65);
 			\spy on (0.12,1.15) in node [left] at (2,1.65);
 			\spy on (-1.2,-0.4) in node [left] at (-1.2,-1.65);			
 			\spy on (0.22,-0.75) in node [left] at (2,-1.65);
 			\end{scope}
 			\end{tikzpicture}}
 		\\
 		
 		& \multicolumn{4}{c}{(b) GE clinical data}	
 		
 	\end{tabular}
 	
 	\caption{Comparison of 2D reconstructed images from different MBIR methods with different number of views (2D fan-beam CT geometry; display window is $[800, 1200]$ HU). 	See the reference images and reconstructed images via FBP in Fig.~\ref{fig:fbpconvnet:ref} and Fig.~\ref{fig:2D_fbp}, respectively. See the error maps between the reconstructed images and the ground truth for (a) in Fig.~\ref{fig:2DCTrecon_err} in the supplement.
	}
 	\label{fig:2DCTrecon}
 	 	\vspace{-0.1in}
 \end{figure*}

 
 

 

 \begin{figure*}[!t]
 	\centering
 	\small\addtolength{\tabcolsep}{-7.5pt}

 	\begin{tabular}{ccccc}
 		
 		{} & \small{FBP} & \small{PWLS-EP} & \small{PWLS-ST-$\ell_2$ (Zheng et al., 2018)} & \small{Proposed PWLS-ST-$\ell_1$} \\
 		
 		\raisebox{-.5\height}{\begin{turn}{+90} \small{$25$\% ($246$) views} \end{turn}}~ &
 		\raisebox{-.5\height}{
 			\begin{tikzpicture}
 			\begin{scope}[spy using outlines={rectangle,yellow,magnification=1.55,size=18mm, connect spies}]
 			\node {\includegraphics[scale=0.55]{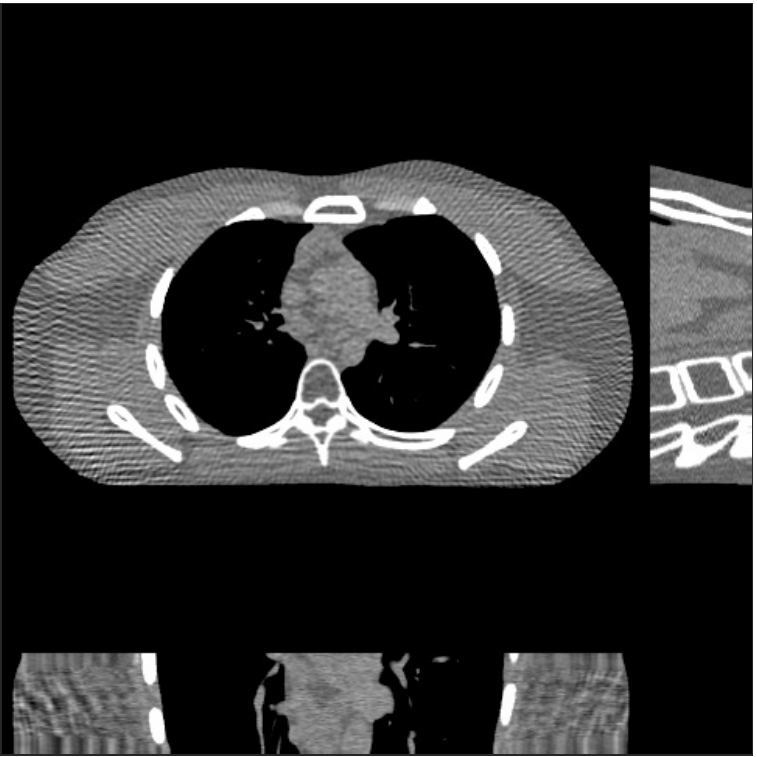}};
 			\spy on (0.9, 0.1) in node [left] at (-0.41,1.92);
 			\node [white] at (1,1.85) {\small $\mathrm{RMSE} = 58.0$};
 			\end{scope}
 			\end{tikzpicture}} &
 		\raisebox{-.5\height}{
 			\begin{tikzpicture}
 			\begin{scope}[spy using outlines={rectangle,yellow,magnification=1.55,size=18mm, connect spies}]
 			\node {\includegraphics[scale=0.55]{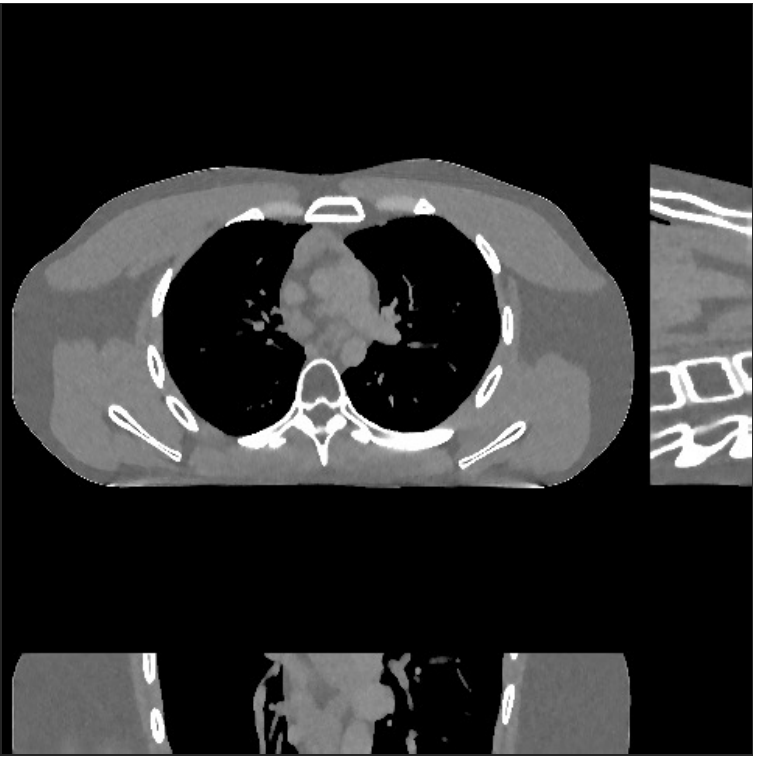}};
 			\spy on (0.9, 0.1) in node [left] at (-0.41,1.92);
 			\node [white] at (1,1.85) {\small $\mathrm{RMSE} = 29.3$};
 			\end{scope}
 			\end{tikzpicture}} &
 		\raisebox{-.5\height}{
 			\begin{tikzpicture}
 			\begin{scope}[spy using outlines={rectangle,yellow,magnification=1.55,size=18mm, connect spies}]
 			\node {\includegraphics[scale=0.55]{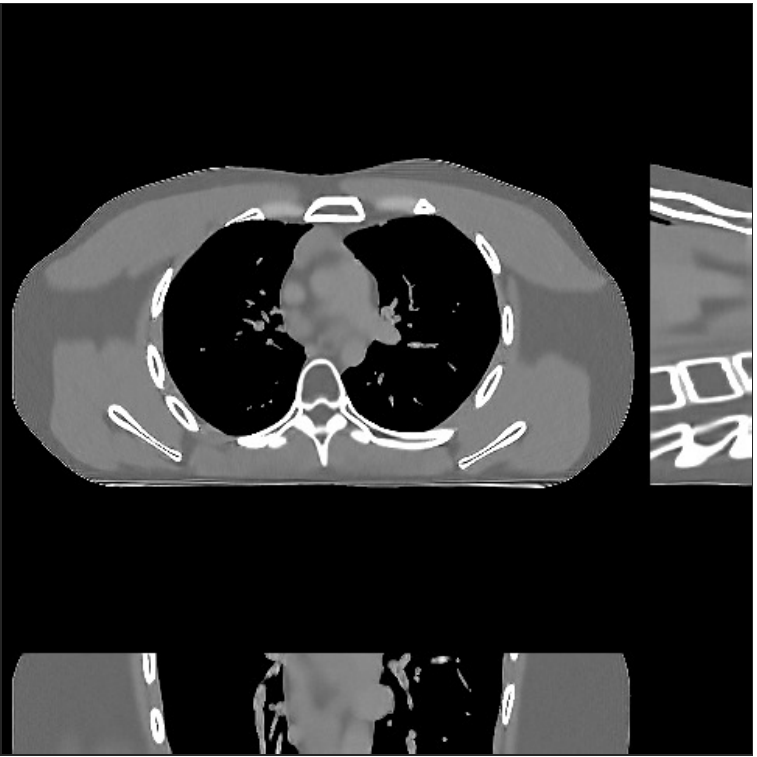}};
 			\spy on (0.9, 0.1) in node [left] at (-0.41,1.92);
 			\node [white] at (1,1.85) {\small $\mathrm{RMSE} = 27.2$};
 			\end{scope}
 			\end{tikzpicture}} &
 		\raisebox{-.5\height}{
 			\begin{tikzpicture}
 			\begin{scope}[spy using outlines={rectangle,yellow,magnification=1.55,size=18mm, connect spies}]
 			\node {\includegraphics[scale=0.55]{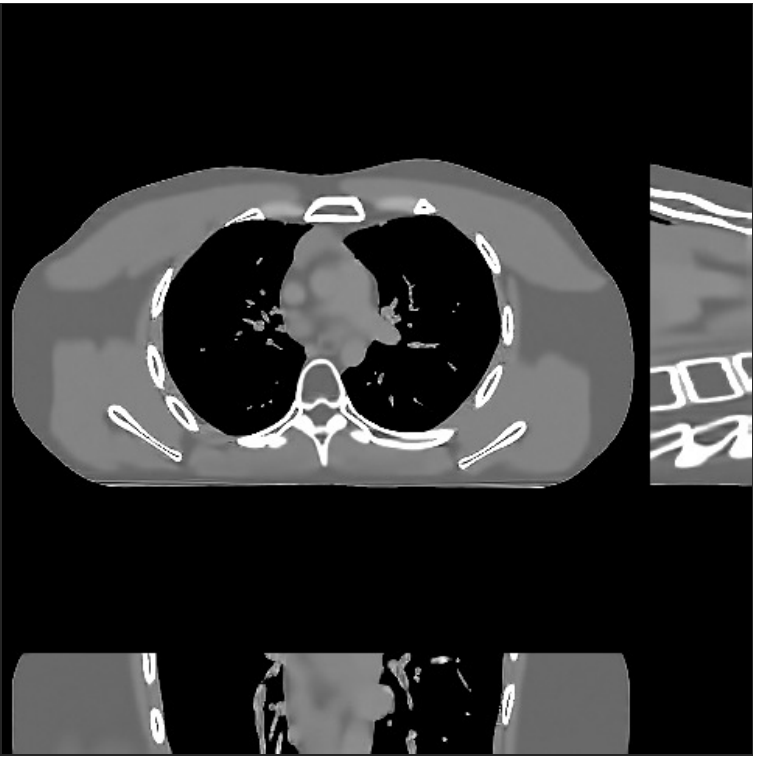}};
 			\spy on (0.9, 0.1) in node [left] at (-0.41,1.92);
 			\node [white] at (1,1.85) {\small \color{yellow}{$\mathrm{RMSE} = 22.2$}};
 			\end{scope}
 			\end{tikzpicture}} \\

 		\raisebox{-.5\height}{\begin{turn}{+90} \small{$12.5$\% ($123$) views} \end{turn}}~ &
 		\raisebox{-.5\height}{
 			\begin{tikzpicture}
 			\begin{scope}[spy using outlines={rectangle,yellow,magnification=1.55,size=18mm, connect spies}]
 			\node {\includegraphics[scale=0.55]{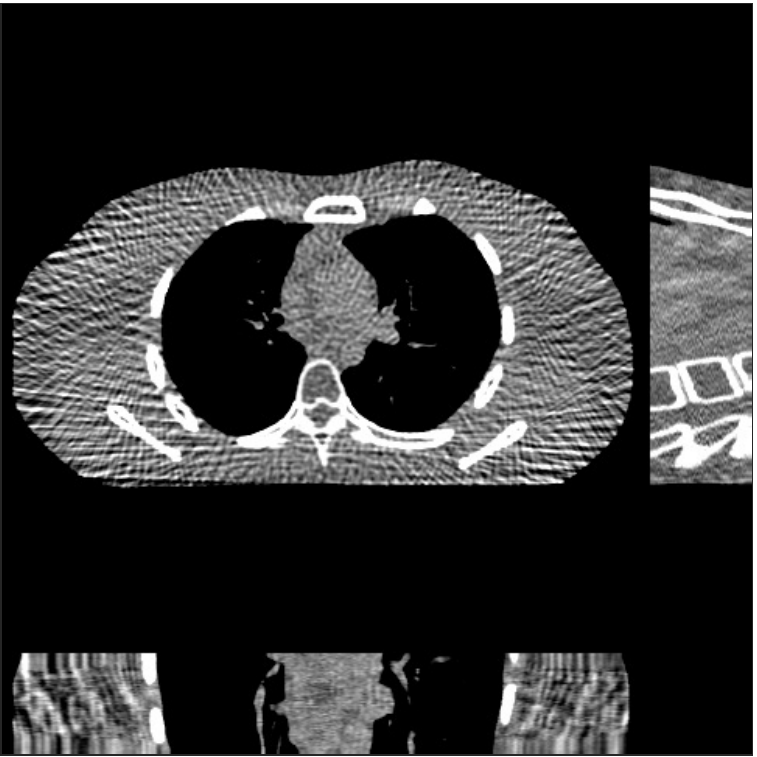}};
 			\spy on (0.9, 0.1) in node [left] at (-0.41,1.92);
 			\node [white] at (1,1.85) {\small $\mathrm{RMSE} = 80.2$};
 			\end{scope}
 			\end{tikzpicture}} &
 		\raisebox{-.5\height}{
 			\begin{tikzpicture}
 			\begin{scope}[spy using outlines={rectangle,yellow,magnification=1.55,size=18mm, connect spies}]
 			\node {\includegraphics[scale=0.55]{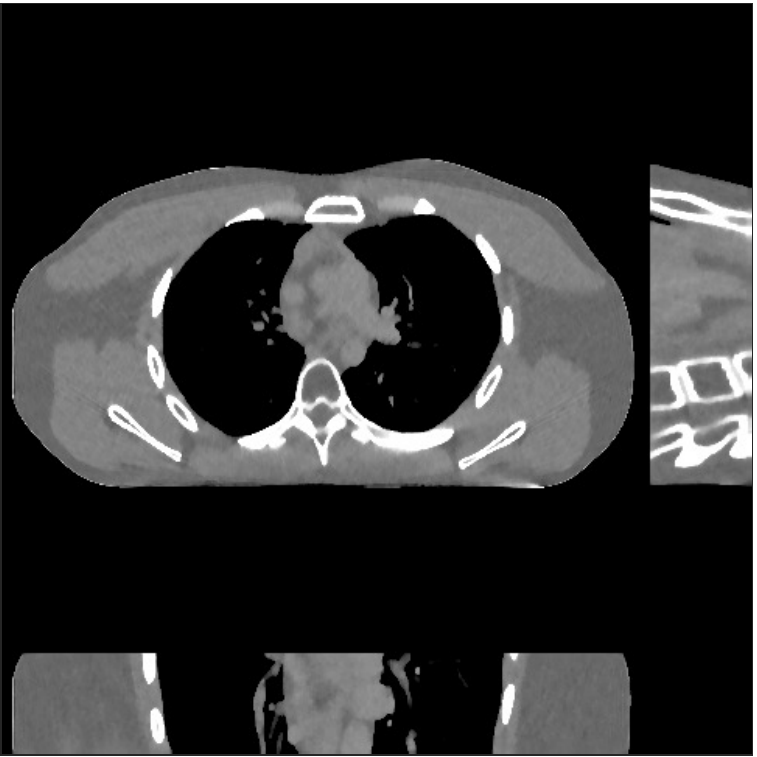}};
 			\spy on (0.9, 0.1) in node [left] at (-0.41,1.92);
 			\node [white] at (1,1.85) {\small $\mathrm{RMSE} = 36.9$};
 			\end{scope}
 			\end{tikzpicture}} &
 		\raisebox{-.5\height}{
 			\begin{tikzpicture}
 			\begin{scope}[spy using outlines={rectangle,yellow,magnification=1.55,size=18mm, connect spies}]
 			\node {\includegraphics[scale=0.55]{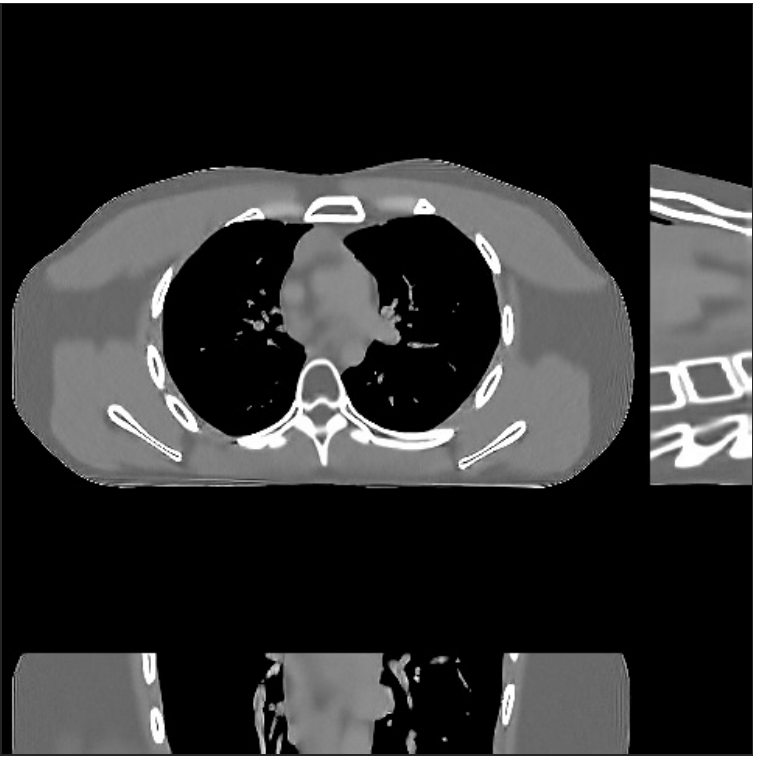}};
 			\spy on (0.9, 0.1) in node [left] at (-0.41,1.92);
 			\node [white] at (1,1.85) {\small $\mathrm{RMSE} = 30.2$};
 			\end{scope}
 			\end{tikzpicture}} &
 		\raisebox{-.5\height}{
 			\begin{tikzpicture}
 			\begin{scope}[spy using outlines={rectangle,yellow,magnification=1.55,size=18mm, connect spies}]
 			\node {\includegraphics[scale=0.55]{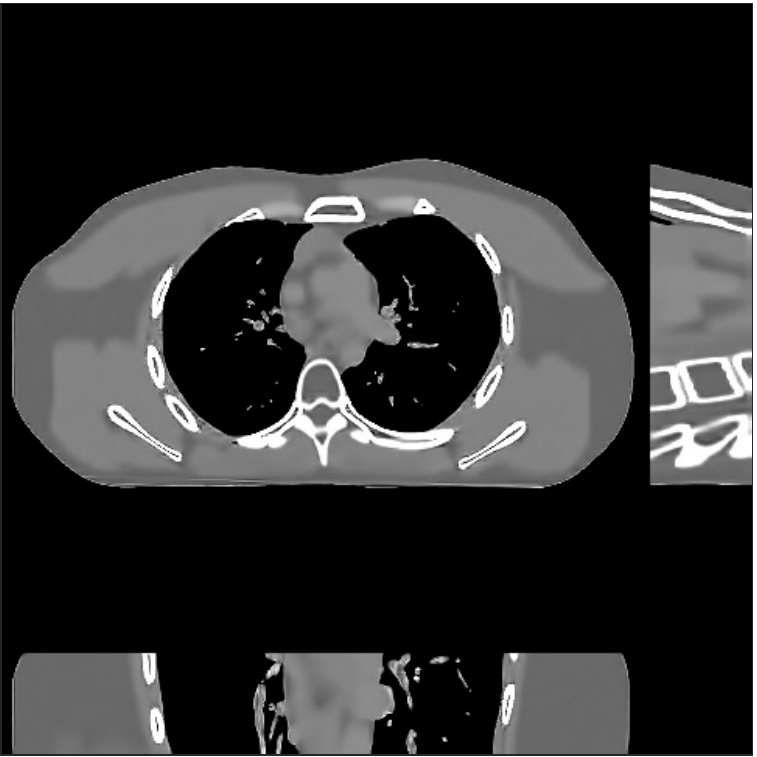}};
 			\spy on (0.9, 0.1) in node [left] at (-0.41,1.92);
 			\node [white] at (1,1.85) {\small \color{yellow}{$\mathrm{RMSE} = 25.6$}};
 			\end{scope}
 			\end{tikzpicture}} 
 		
 	\end{tabular}
 	\caption{Comparison of 3D reconstructed images from different X-ray CT reconstruction methods with different number of views (axial 3D cone-beam geometry; display window is $[800, 1200]$ HU; displayed for the central axial, sagittal, and coronal planes; see the ground truth in Fig.~\ref{fig:3Dtruth} in the supplement.) 
	}
 	\label{fig:3DCTrecon}
 \end{figure*}

 \subsubsection{2D Fan-Beam - Image Reconstruction}
 \label{subsubsec:2d_recon}

This section describes parameters used in reconstruction experiments with the XCAT phantom data and the clinical chest data.
In XCAT phantom experiments, we initialized the PWLS-EP method with FBP reconstructions, and ran the relaxed OS-LALM \cite{Nien&Fessler:16:TMI} for $100$ iterations with $12$ ordered subsets. We chose the regularization parameter (balancing the data fitting term and the regularizer) as $2^{15.5}$ and $2^{15.0}$ for $246$ and $123$ views, respectively. 
For both PWLS-ST-$\ell_1$ and PWLS-ST-$\ell_2$ methods, we used a patch size $8 \!\times\! 8$ with a $1 \!\times\! 1$ overlapping stride. We used converged PWLS-EP reconstructions for initialization and set a stopping criterion as meeting the maximum number of iterations, e.g., $\mathrm{Iter} \!=\! 1000$. For the image update, we set $\mathrm{Iter}_\mathrm{ADMM} = 2$ ($2$ PCG iterations \cite{Ramani&Fessler:12MI}) for PWLS-ST-$\ell_1$; and set $12$ relaxed OS-LALM iterations without ordered subsets for PWLS-ST-$\ell_2$. 
For PWLS-ST-$\ell_1$, we tuned $\nu, \mu$ using the condition number based selection schemes, i.e., $\kappa_{\mathrm{des},\nu}$ in \R{eq:kappa_nu} and $\kappa_{\mathrm{des},\mu}$ in \R{eq:kappa_mu}. We finely tuned the parameters $\lambda, \gamma$ to achieve good image quality. 
For PWLS-ST-$\ell_1$, we chose $\{ \lambda, \gamma/\lambda, \kappa_{\mathrm{des},\nu}, \kappa_{\mathrm{des},\mu} \}$ as follows:
$\{ 1.3 \!\times\! 10^6, 80, 30, 30 \}$ for $246$ views; $\{ 9 \!\times\! 10^5, 80, 30, 30 \}$ for $123$ views.
For PWLS-ST-$\ell_2$, we chose $\{ \lambda, \gamma \}$ as follows: $\{ 3 \!\times\! 10^{5}, 20 \}$ for $246$ views; $\{ 1.6 \!\times\! 10^{5}, 20 \}$ for $123$ views.
Note that $\lambda$ and $\gamma$ are in HU.
For PWLS-DL, we chose a maximum sparsity level of $25$, set an error tolerance as $60$, and set a regularization parameter as $1.3 \!\times\! 10^5$ and $7 \!\times\! 10^4$ for $246$ and $123$ views, respectively. Similar to the PWLS-ST method, we finely tuned these parameters to achieve good image quality.

In clinical data reconstruction, unless stated otherwise, we used the same learned models, trained networks, and reconstruction parameter sets listed above. 
We initialized all methods with FBP reconstructions. For PWLS-EP, we ran the relaxed OS-LALM for $50$ iterations with $6$ ordered subsets, and chose the regularization parameter as $2^{2.5}$ and $2^{2}$ for $246$ and $123$ views, respectively.
For PWLS-ST-$\ell_1$, we used the identical $\kappa_{\mathrm{des},\nu}$, $\kappa_{\mathrm{des},\mu}$, $\gamma/\lambda$ values chosen in the XCAT phantom experiments. 
To automatically select the regularization parameter $\lambda$, we used the guideline described in Section~\ref{sec:param},
and it is chosen as approximately $1.2 \!\times\! 10^{-2}$ and $8 \!\times\! 10^{-3}$ for $246$ and $123$ views, respectively.
For PWLS-ST-$\ell_2$, we chose $\{ \lambda, \gamma \}$ as follows: $\{ 10^{-3}, 35 \}$ for $246$ views; $\{ 5 \!\times\! 10^{-4}, 40 \}$ for $123$ views. 
For PWLS-DL, we chose a maximum sparsity level of $25$, set an error tolerance as $85$, and set a regularization parameter as $7 \!\times\! 10^{-4}$ and $6 \!\times\! 10^{-4}$ for $246$ and $123$ views, respectively.

\subsubsection{3D Cone-Beam - Imaging}
\label{subsubsec:3d_imaging}

In the 3D CT experiments, we simulated an axial cone-beam CT scan using an $840 \times 840 \times 96$ volume from the XCAT phantom (air cropped, $\Delta_x=\Delta_y=0.4883$ mm and $\Delta_z=0.625$ mm). 
We generated sinograms of size  $888$ (detector channels) $\times 64$ (detector rows) $\times \{246, 123$\} ($984$ is the number of full views) using GE LightSpeed cone-beam geometry corresponding to a monoenergetic source with $\rho_0 = 10^5$ incident photons per ray and no scatter, and $\sigma^2 = 5^2$. 
We reconstructed a $420 \times 420 \times 96$ volume with a coarser grid, where $\Delta_x=\Delta_y=0.9766$ mm and $\Delta_z=0.625$ mm.
We defined a cylinder ROI for the 3D case, which consisted of the central $64$ of $96$ axial slices and a circular (around center) region in each slice. The diameter of the circle was $420$ pixels, which is the width of each slice.

\subsubsection{3D Cone-Beam - Training and Image Reconstruction} 
\label{subsubsec:3d_recon}

Similar to the 2D experiments, we pre-learned square STs using $8 \times 8 \times 8$ patches (with an overlapping stride $2 \times 2 \times 2$) extracted from a $420 \times 420 \times 54$ volume of the XCAT phantom, which is different from the volume used for testing.
Initialized with the 3D DCT, we ran the transform learning algorithm \cite{Ravishankar&Bresler:15TSP} for $1000$ iterations with $\tau \!=\! 5.63 \!\times\! 10^{15}$, $\gamma' = 100$ and $\xi = 1$.

	For the PWLS-EP method, initialized with FBP reconstructions, we ran the relaxed OS-LALM for $100$ iterations with $12$ subsets and regularization parameter of $2^{14}$, for both $246$ and $123$ views.
	For both PWLS-ST-$\ell_1$ and PWLS-ST-$\ell_2$, we chose an $8 \times 8 \times 8$ patch size with a patch stride $3 \times 3 \times 3$. Initialized with converged PWLS-EP reconstructions, we chose a maximum number of iterations $\mathrm{Iter} \!=\! 500$ as the stopping criterion.
	For the image update, we set $\mathrm{Iter}_\mathrm{ADMM}$ as $2$ (we empirically found that $2$ PCG iterations provide reasonable convergence behavior, see  Fig.~\ref{fig:pcg}) for PWLS-ST-$\ell_1$, and set $2$ relaxed OS-LALM iterations with $4$ ordered subsets for PWLS-ST-$\ell_2$ \cite{Zheng&etal:17arXiv}.
	For PWLS-ST-$\ell_1$, we chose $\{ \lambda, \gamma/\lambda, \kappa_{\mathrm{des},\nu}, \kappa_{\mathrm{des},\mu} \}$ as follows: $\{ 10^7, 15, 10, 50 \}$ for $246$ views; $\{ 8 \!\times\! 10^6, 15,  10, 40 \}$ for $123$ views.
	For PWLS-ST-$\ell_2$, we chose $\{ \lambda, \gamma \}$ as follows: $\{ 10^{6}, 18 \}$ for $246$ views; $\{ 8 \!\times\! 10^{5}, 18 \}$ for $123$ views.

%
%

\subsection{Results and Discussion}
\label{subsec:results_discussion}

\subsubsection{Reconstruction Comparisons among Different MBIR Methods} 
\label{subsec: MBIR}

This section compares the reconstruction quality and runtime among the proposed MBIR method, PWLS-ST-$\ell_1$, and other three MBIR methods,  PWLS-EP, PWLS-DL, and PWLS-ST-$\ell_2$.
Table~\ref{tab:CTrecon} shows that, for both 2D and 3D sparse-view CT reconstructions of the XCAT phantom, the proposed PWLS-ST-$\ell_1$ model outperforms PWLS-EP and PWLS-ST-$\ell_2$ in terms of RMSE. 
In addition, PWLS-ST-$\ell_1$ using a square transform (of size $64\times64$) achieves lower RMSE than PWLS-DL using an overcomplete dictionary (of size $64\times256$) for 2D sparse-view reconstructions.
 Fig.~\ref{fig:2DCTrecon}(a) and Fig.~\ref{fig:3DCTrecon} show the reconstructed images for 2D and 3D phantom experiments, with different reconstruction models and different number of views. (See the corresponding error maps in the supplement.)
 The proposed PWLS-ST-$\ell_1$ consistently gives more accurate image reconstructions compared to other MBIR methods. 
 Specifically, PWLS-ST-$\ell_1$ has smaller errors in the heart region (see zoom-ins in Fig.~\ref{fig:2DCTrecon}(a)) of 2D reconstructions than PWLS-DL and PWLS-ST-$\ell_2$. 
 In addition, compared to PWLS-ST-$\ell_1$, PWLS-DL and PWLS-ST-$\ell_2$ have some ringing artifacts around the edges with high transition, e.g., edges between air and soft tissues. (See a comparison of profiles of PWLS-ST-$\ell_1$ and PWLS-ST-$\ell_2$ in the supplement.) 
 In particular, PWLS-ST-$\ell_2$ and PWLS-DL give more visible ringing artifacts for 2D reconstruction from fewer views, and PWLS-ST-$\ell_2$ has these ringing artifacts for 3D reconstructions regardless of the number of views (see zoom-ins in Fig.~\ref{fig:3DCTrecon}). Table~\ref{tab:runtime} reports runtimes of different MBIR methods in  reconstructing the $123$-views XCAT phantom scan.
 (FBPConvNet is a non-MBIR method and its runtime for processing a $512\times 512$ image is approximately one second with a TITAN Xp GPU.)
While providing better reconstruction quality, 
the proposed Algorithm~\ref{alg:PWLS-ST-l1} of PWLS-ST-$\ell_1$ 
has shorter runtime compared to the algorithms of PWLS-DL and PWLS-ST-$\ell_2$ in Section~\ref{sec:exp}.
Similar to the PWLS-EP algorithm, 
the reconstruction time of the PWLS-DL, \mbox{PWLS-ST-$\ell_2$}, and \mbox{PWLS-ST-$\ell_1$} algorithms 
can be further reduced by using ordered subsets \cite{erdogan:99:osa}.
 

 Fig.~\ref{fig:2DCTrecon}(b) shows that when tested on the clinical scan data, the proposed \mbox{PWLS-ST-$\ell_1$} method improves
reconstruction quality in terms of noise and artifacts removal (e.g., see zoom-ins for soft-issue regions), and edge preservation (e.g., see zoom-ins for bone regions), compared to PWLS-EP and PWLS-ST-$\ell_2$.
 Compared to PWLS-DL, \mbox{PWLS-ST-$\ell_1$} achieves comparable image quality, but requires less computational complexity.


 The benefit of the proposed PWLS-ST-$\ell_1$ over PWLS-ST-$\ell_2$ can be explained when there exist some outliers for some $\mb{z}^{(i)}$: $\| \widetilde{\mb{\Psi}} \mb{x}  - \mb{z}^{(i)} \|_1$ in \R{sys:L1trsf:x} gives equal emphasis to all sparse codes -- from small to large coefficients that generally correspond to edges in low- and high-contrast regions, respectively -- in estimating $\mb{x}$; however, PWLS-ST-$\ell_2$ adjusts $\mb{x}$ to mainly minimize the outliers, i.e., it may not pay enough attention to reconstruct regions with small coefficients.
 The histogram results in Fig.~\ref{fig:histogram} reveal model mismatch of PWLS-ST-$\ell_2$ over the iterations. 
 Fig.~\ref{fig:2DCTrecon}, Fig.~\ref{fig:3DCTrecon}, and Table~\ref{tab:CTrecon} show that PWLS-ST-$\ell_1$ can moderate model mismatch, and provides more accurate reconstruction than PWLS-ST-$\ell_2$.

\begin{table}[!t]	
	\centering
	\renewcommand{\arraystretch}{1.1}
	
	\caption{RMSE (HU) of different 2D and 3D X-ray CT reconstructions with different number of projection views (XCAT phantom experiments)}	
	\label{tab:CTrecon}
	\begin{tabular}{C{0.6cm}|C{0.7cm}|C{0.9cm}C{0.9cm}C{0.9cm}C{0.9cm}C{0.9cm}}
		\hline \hline
		{}  & Views$^a$      & FBP    &  PWLS-EP &  PWLS-DL  &  PWLS-ST-$\ell_2$  & PWLS-ST-$\ell_1$   \\ 
		\hline
		\multirow{2}*{2D$^b$} & $246$      &  $60.5$     & $30.7$  &  $24.9$        &   $26.9$      &   $\mathbf{21.5}$  \\   
		\cline{2-7}
		& $123$     & $82.7$      & $35.0$     & $26.9$    &   $30.9$      &   $\mathbf{25.8}$  \\ 
		\hline
		\multirow{2}*{3D$^c$} & $246$      &  $58.0$       & $29.3$   & -   &  $27.2$      &   $\mathbf{22.2}$  \\   
		\cline{2-7}
		& $123$     & $80.2$      & $36.9$    & -    &   $30.2$      &   $\mathbf{25.6}$  \\ 
		\hline\hline
	\end{tabular}
	\vspace{-0.1in}
	\medskip
	\begin{myquote}{0.1in}
		$^a$The $246$ and $123$ projection views correspond to $25$\% and $12.5$\% of the full views, $984$, respectively.
		\\
		$^b$For the 2D CT experiments, fan-beam geometry was used.
		\\
		$^c$For the 3D CT experiments, axial cone-beam geometry was used.
	\end{myquote}
\end{table}

\begin{table}[!t]	
	\centering
	\renewcommand{\arraystretch}{1.1}
	
	\caption{Comparisons of runtime among different MBIR methods (2D fan-beam, XCAT phantom experiments, and $12.5$\% ($123$) views)}	
	\label{tab:runtime}
	
	\begin{tabular}{C{1.6cm}C{1.6cm}C{1.8cm}C{1.8cm}}
		\hline \hline
		PWLS-EP$^a$ & 
		PWLS-DL$^b$ & 
		PWLS-ST-$\ell_2$$^b$ &
		PWLS-ST-$\ell_1$$^c$ (Alg.~\ref{alg:PWLS-ST-l1}) \\
		\hline
		$2$~minutes & $1133$~minutes & $95$~minutes  & $80$~minutes  \\ 
		\hline\hline
	\end{tabular}
	
	\medskip
	\begin{myquote}{0.1in}
		$^a$The PWLS-EP method used $100$ iterations with $12$ ordered subsets.
		\\
		$^b$The PWLS-DL and PWLS-ST-$\ell_2$ methods used $1000$ outer iterations.
		\\
		$^c$The PWLS-ST-$\ell_1$ method used $1000$ outer iterations.
		\\
		The runtimes were recorded by Matlab implementations on two $2.6$ GHz CPUs with $12$-core Intel Xeon E5-2690 v3 processors.
	\end{myquote}
\end{table}

\begin{figure}[!h]
	\centering
	\small\addtolength{\tabcolsep}{-5pt}
	\begin{tabular}{c}
		\includegraphics[scale=0.42,clip, trim=0.5em 0em 2em 2.2em]{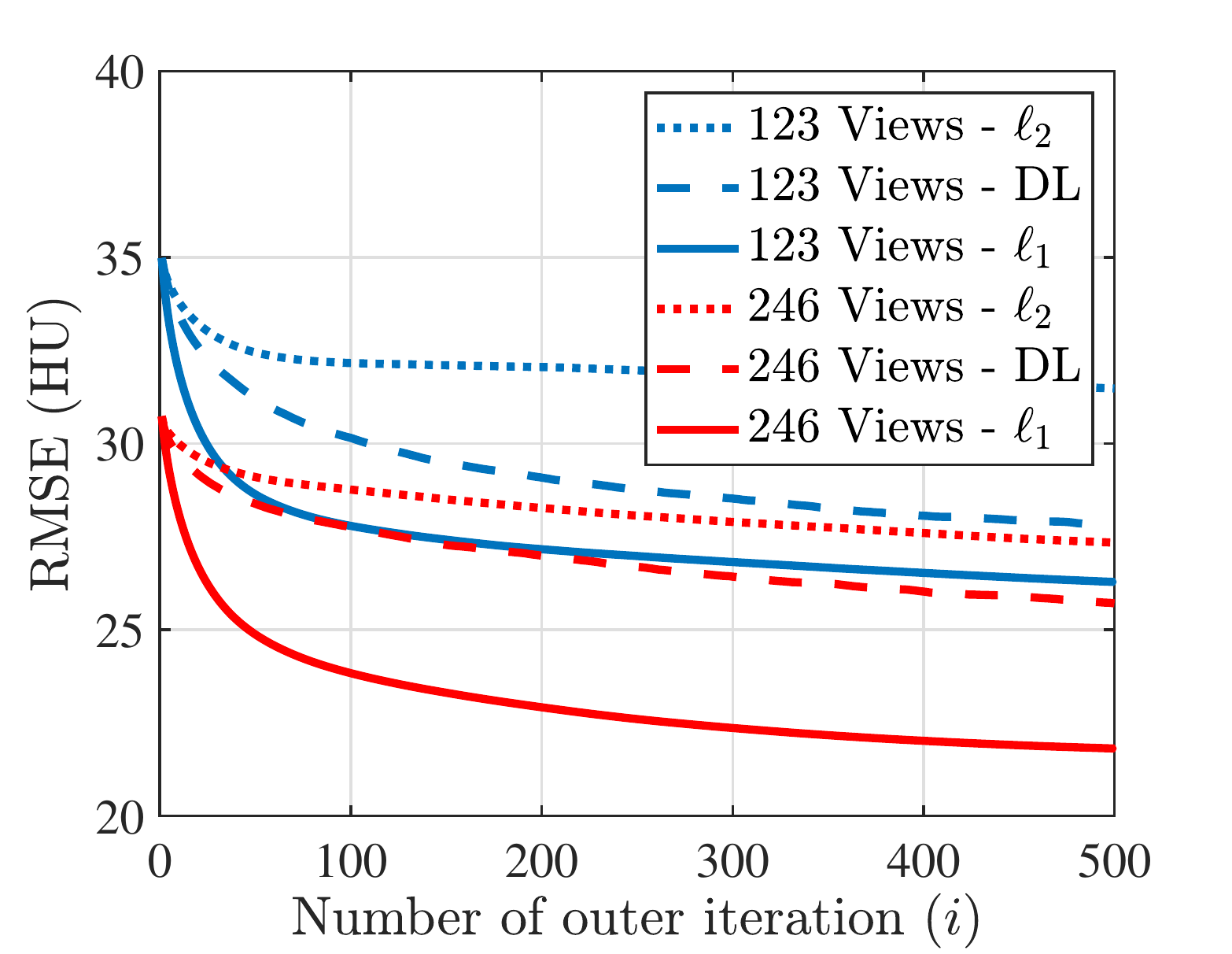} 
		\\
		\small{(a) 2D fan-beam CT experiments}
		\\
		\includegraphics[scale=0.42,clip, trim=0.5em 0.4em 2em 2.2em]{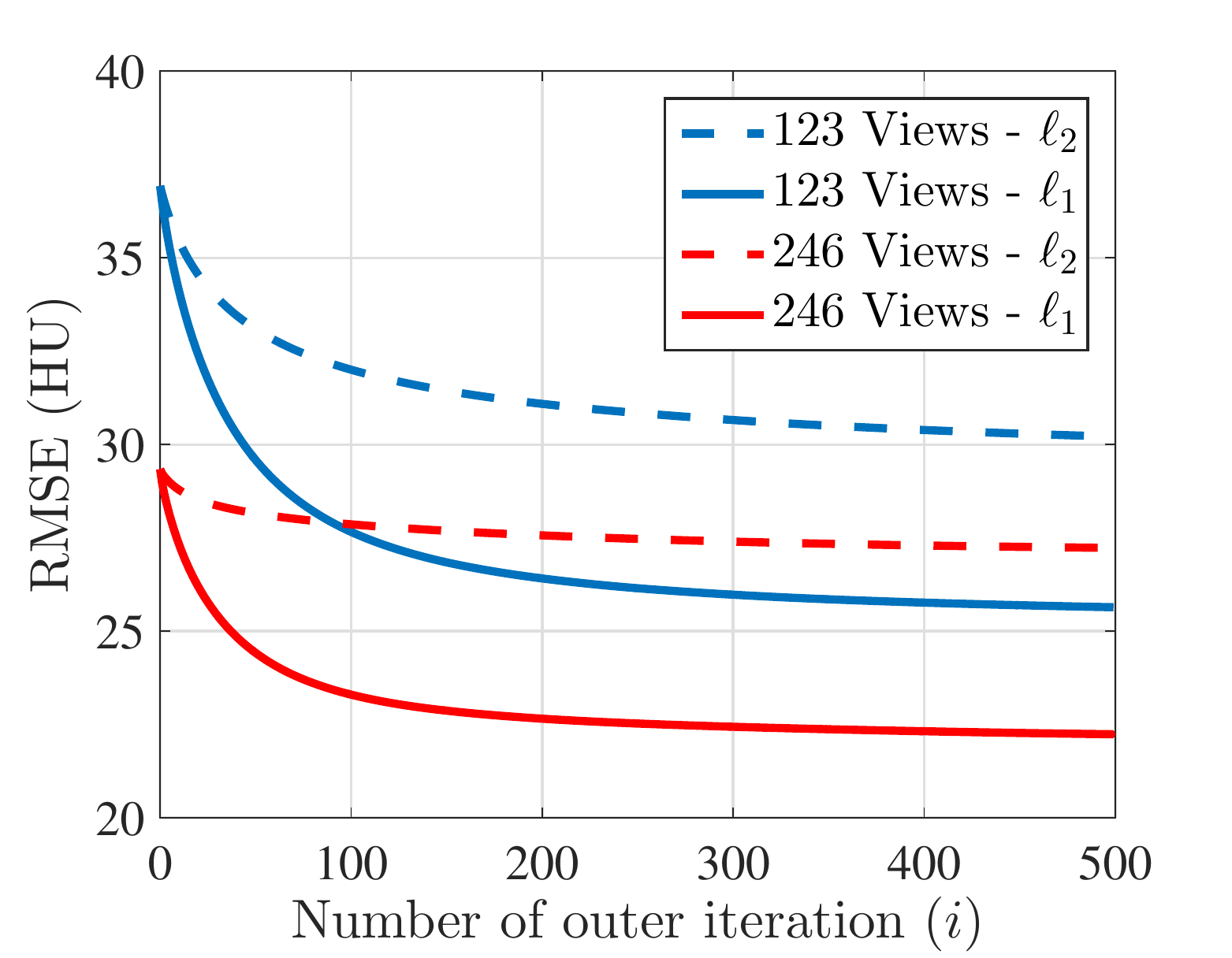} 
		\\
		\small{(b) 3D axial cone-beam CT experiments}
	\end{tabular}
	\caption{RMSE convergence behavior for PWLS-ST-$\ell_1$, PWLS-ST-$\ell_2$, and/or PWLS-DL in different CT geometries and projection views.}
	\label{fig:convg}
\end{figure}

 \begin{figure*}[!t]
 	\centering  
 	\small\addtolength{\tabcolsep}{-9.5pt}
 	
 	\begin{tabular}{ccc|c}
 		
 		{} & 
 		\small{FBPConvNet (Jin et al., 2017)} & 
 		\small{Proposed PWLS-ST-$\ell_1$} &
 		\small{Reference} 
 		\\
 		
 		
 		\raisebox{-.5\height}{
 			\begin{turn}{+90} \small{$25$\% ($246$) views} \end{turn}
 		}~ 
 		&
 		
 		\raisebox{-.5\height}{
 			\begin{tikzpicture}
 			\begin{scope}[spy using outlines={rectangle,yellow,magnification=1.9,size=12mm, connect spies}]
 			\node {\includegraphics[width=0.23\textwidth]{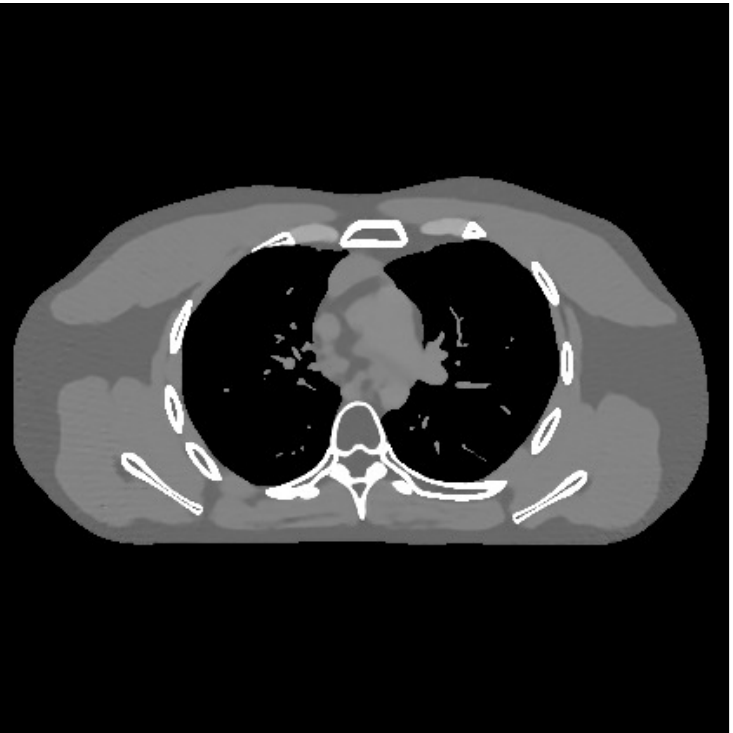}};		
 			\spy on (-1.75,-0.3) in node [left] at (-.8,-1.45);			
 			\spy on (0,0.2) in node [left] at (2,1.45);	
 			\node [white] at (1,-1.85) {\small $\mathrm{RMSE} = 14.6$};
 			\end{scope}
 			\end{tikzpicture}
 		} 
 		&
 		
 		\raisebox{-.5\height}{
 			\begin{tikzpicture}
 			\begin{scope}[spy using outlines={rectangle,yellow,magnification=1.9,size=12mm, connect spies}]
 			\node {\includegraphics[width=0.23\textwidth]{./Fig/246l1_hyper_lam13e5_zeta80_30kap1_30kap2_iter2_learn110-eps-converted-to}};
 			\spy on (-1.75,-0.3) in node [left] at (-.8,-1.45);			
 			\spy on (0,0.2) in node [left] at (2,1.45);	
 			\node [white] at (1,-1.85) {\small {$\mathrm{RMSE} = 21.5$}};
 			\end{scope}
 			\end{tikzpicture}
 		} 
 		&
 		
 		\raisebox{-.5\height}{
 			\begin{tikzpicture}
 			\begin{scope}[spy using outlines={rectangle,yellow,magnification=1.9,size=12mm, connect spies}]
 			\node {\includegraphics[width=0.23\textwidth]{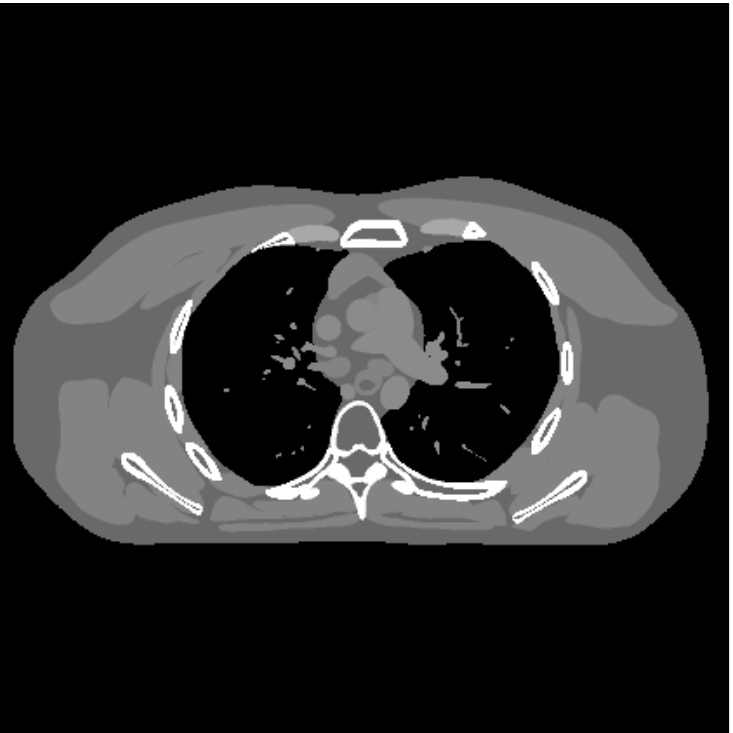}};	
 			\spy on (-1.75,-0.3) in node [left] at (-.8,-1.45);			
 			\spy on (0,0.2) in node [left] at (2,1.45);	
 			\end{scope}
 			\end{tikzpicture}
 		} 
 		\\
 		
 		\raisebox{-.5\height}{
 			\begin{turn}{+90} \small{$12.5$\% ($123$) views} \end{turn}
 		}~ 
 		&
 		
 		\raisebox{-.5\height}{
 			\begin{tikzpicture}
 			\begin{scope}[spy using outlines={rectangle,yellow,magnification=1.9,size=12mm, connect spies}]
 			\node {\includegraphics[width=0.23\textwidth]{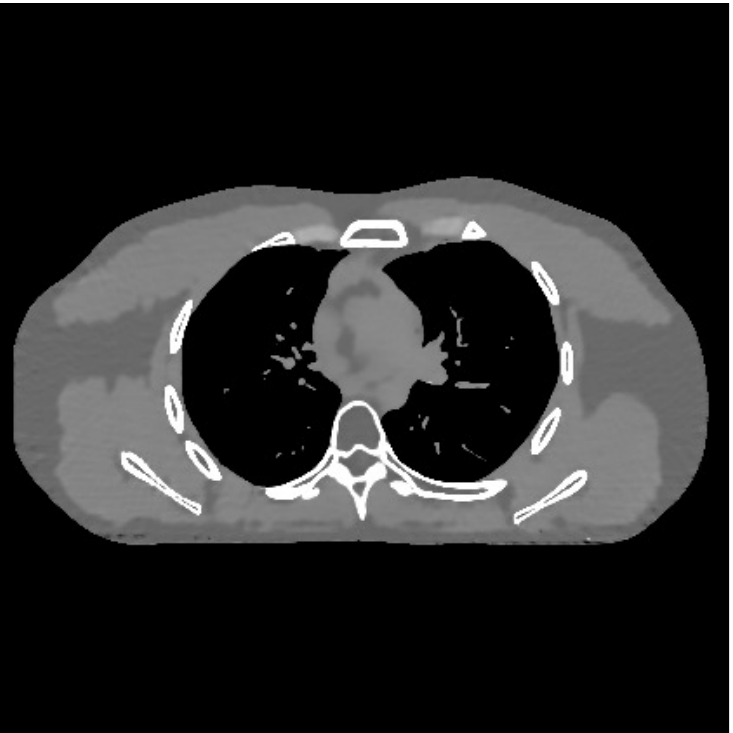}};		
 			\spy on (-1.75,-0.3) in node [left] at (-.8,-1.45);			
 			\spy on (0,0.2) in node [left] at (2,1.45);	
 			\node [white] at (1,-1.85) {\small $\mathrm{RMSE} = 23.9$};
 			\end{scope}
 			\end{tikzpicture}
 		} 
 		&
 		
 		\raisebox{-.5\height}{
 			\begin{tikzpicture}
 			\begin{scope}[spy using outlines={rectangle,yellow,magnification=1.9,size=12mm, connect spies}]
 			\node {\includegraphics[width=0.23\textwidth]{./Fig/123l1_hyper_lam9e5_zeta80_30kap1_30kap2_iter2_learn110-eps-converted-to}};
 			\spy on (-1.75,-0.3) in node [left] at (-.8,-1.45);			
 			\spy on (0,0.2) in node [left] at (2,1.45);
 			\node [white] at (1,-1.85) {\small {$\mathrm{RMSE} = 25.8$}};
 			\end{scope}
 			\end{tikzpicture}
 		} 
 		&
 		\\
 		
 		& 
 		\multicolumn{3}{c}{(a)~XCAT phantom data}
 		\\
 		
 		\raisebox{-.5\height}{
 			\begin{turn}{+90} \small{$25$\% ($246$) views} \end{turn}
 		}~ 
 		&
 		
 		\raisebox{-.5\height}{
 			\begin{tikzpicture}
 			\begin{scope}[spy using outlines={rectangle,yellow,magnification=1.9,size=8mm, connect spies}]
 			\node {\includegraphics[width=0.23\textwidth]{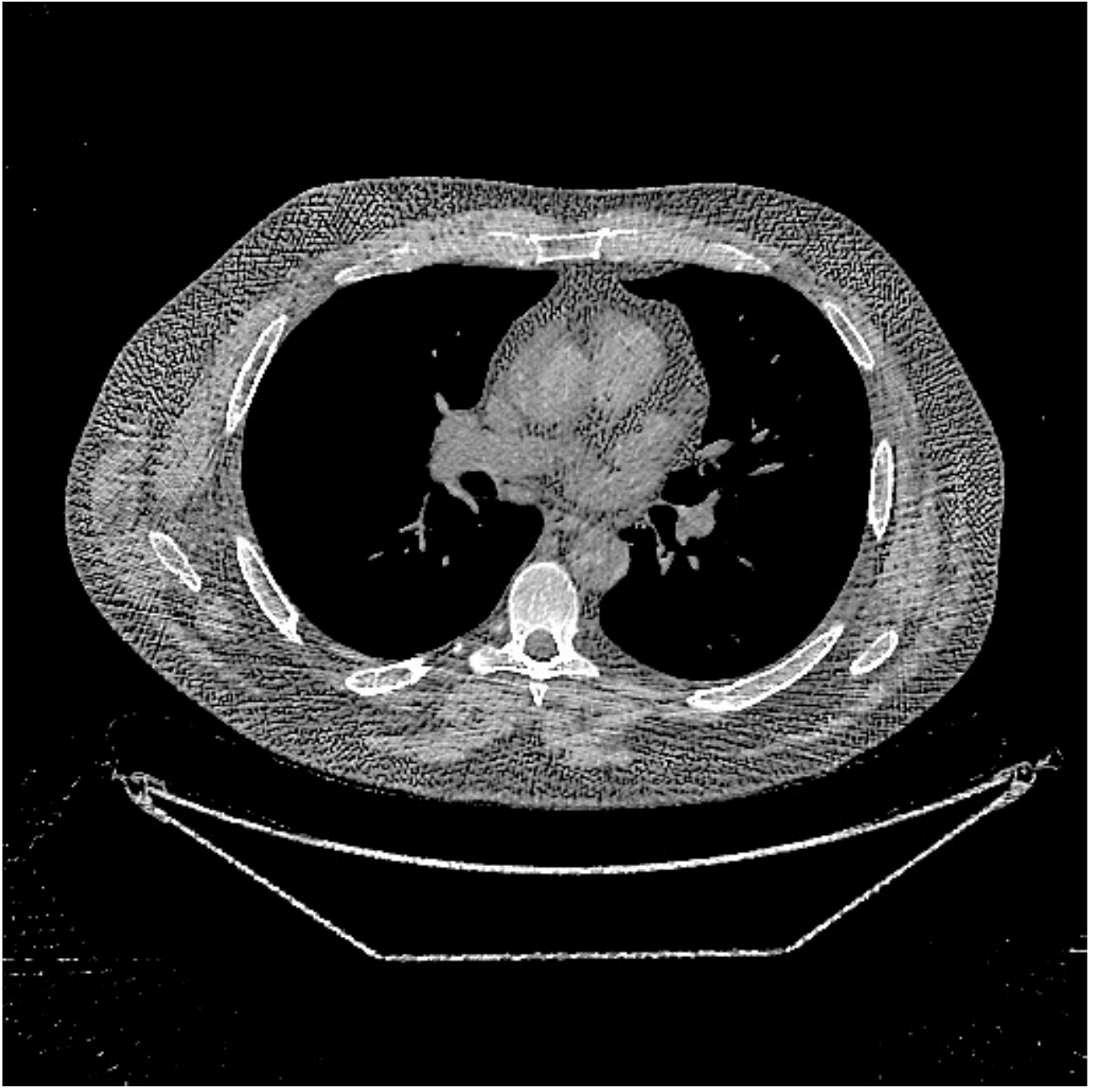}};	
 			\spy on (-0.65,1.1) in node [left] at (-1.2,1.65);
 			\spy on (0.12,1.15) in node [left] at (2,1.65);
 			\spy on (-1.2,-0.4) in node [left] at (-1.2,-1.65);			
 			\spy on (0.22,-0.75) in node [left] at (2,-1.65);
 			\end{scope}
 			\end{tikzpicture}
 		} 
 		&
 		
 		\raisebox{-.5\height}{
 			\begin{tikzpicture}
 			\begin{scope}[spy using outlines={rectangle,yellow,magnification=1.9,size=8mm, connect spies}]
 			\node {\includegraphics[width=0.23\textwidth]{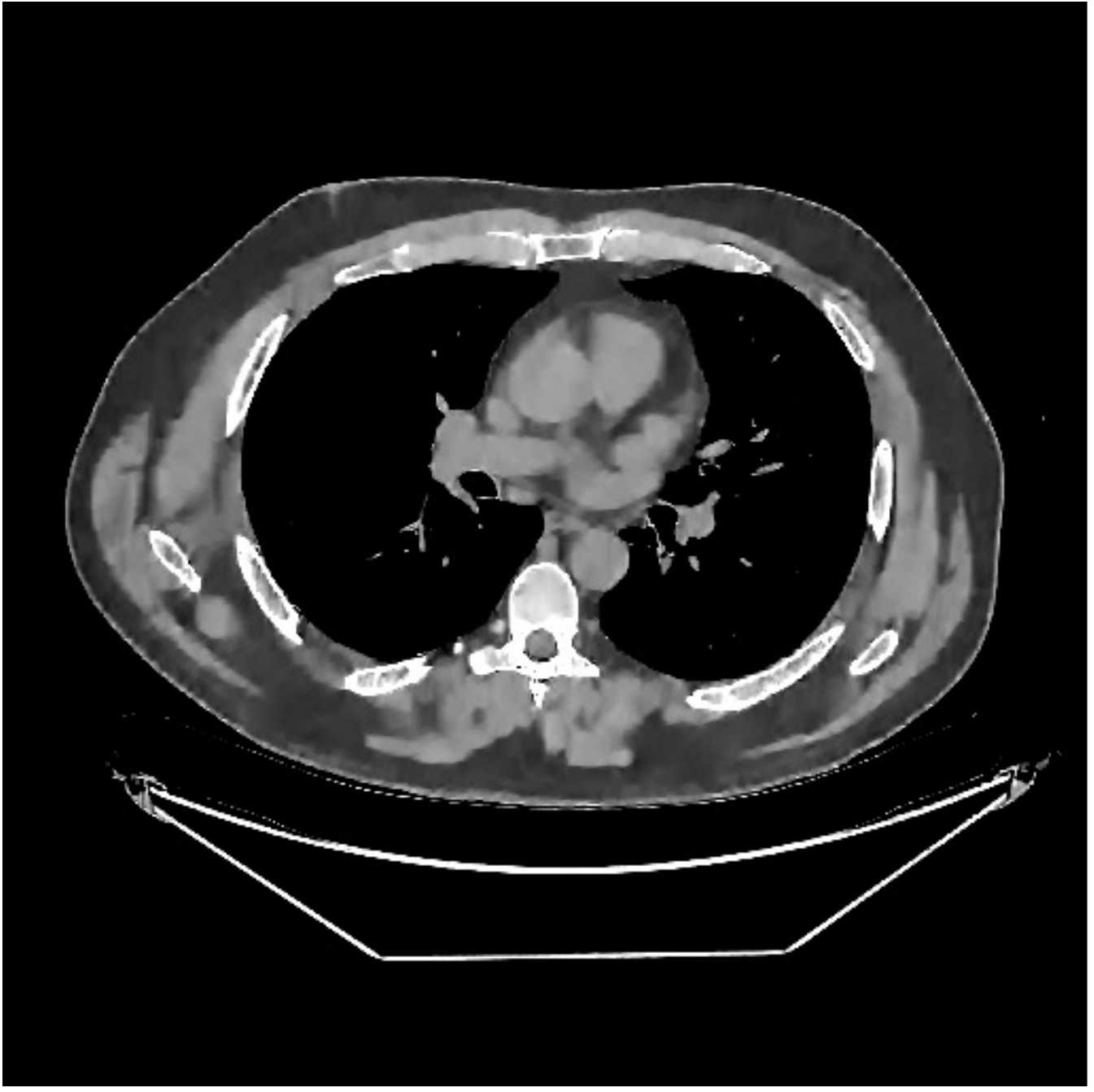}};
 			\spy on (-0.65,1.1) in node [left] at (-1.2,1.65);
 			\spy on (0.12,1.15) in node [left] at (2,1.65);
 			\spy on (-1.2,-0.4) in node [left] at (-1.2,-1.65);			
 			\spy on (0.22,-0.75) in node [left] at (2,-1.65);
 			\end{scope}
 			\end{tikzpicture}
 		} 
 		&
 		
 		\raisebox{-.5\height}{
 			\begin{tikzpicture}
 			\begin{scope}[spy using outlines={rectangle,yellow,magnification=1.9,size=8mm, connect spies}]
 			\node {\includegraphics[width=0.23\textwidth]{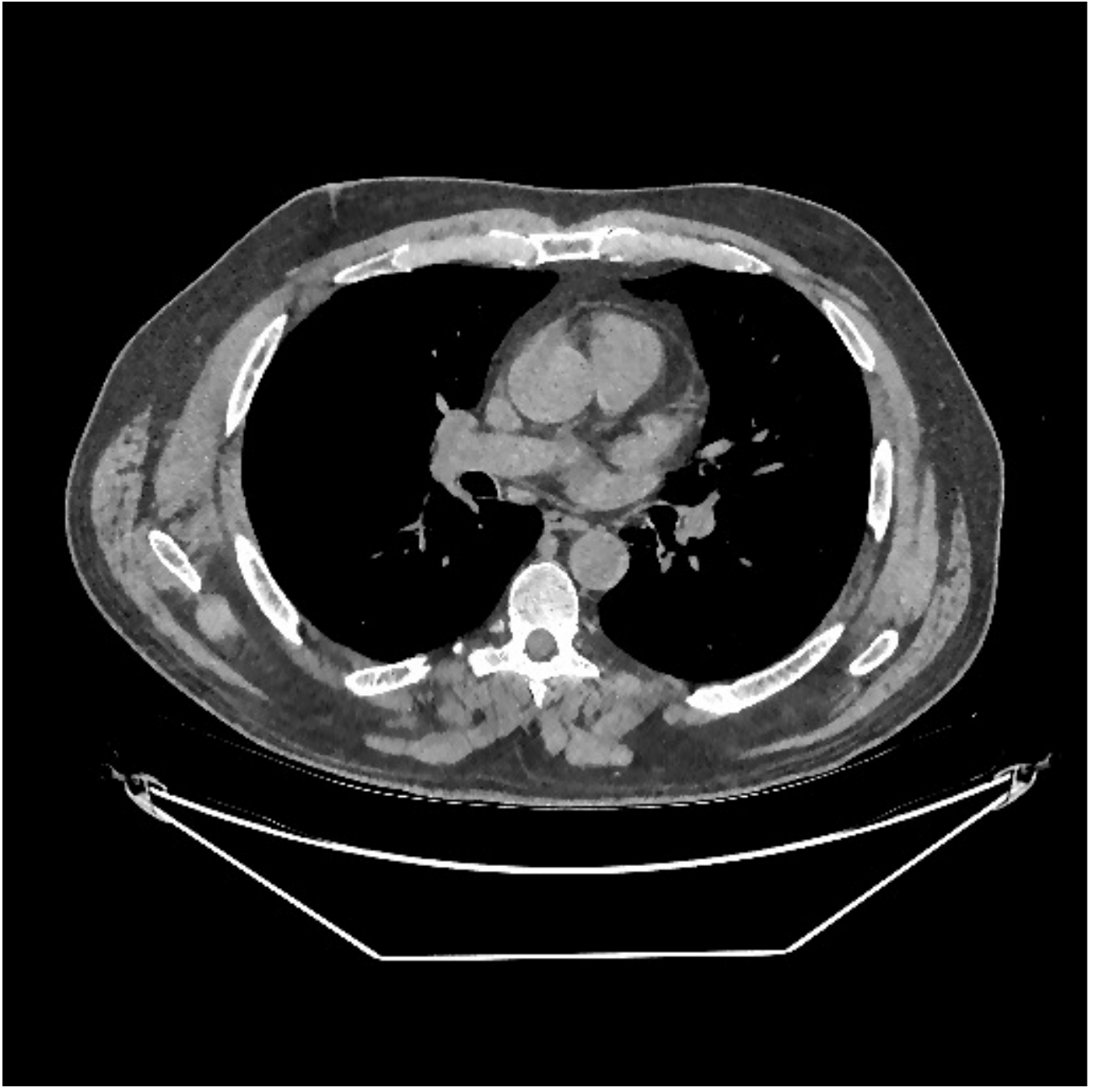}};
 			\spy on (-0.65,1.1) in node [left] at (-1.2,1.65);
 			\spy on (0.12,1.15) in node [left] at (2,1.65);
 			\spy on (-1.2,-0.4) in node [left] at (-1.2,-1.65);			
 			\spy on (0.22,-0.75) in node [left] at (2,-1.65);		 	
 			\end{scope}
 			\end{tikzpicture}
 		} 
 		\\

 		\raisebox{-.5\height}{
 			\begin{turn}{+90} \small{$12.5$\% ($123$) views} \end{turn}
 		}~ 
 		&
 		
 		\raisebox{-.5\height}{
 			\begin{tikzpicture}
 			\begin{scope}[spy using outlines={rectangle,yellow,magnification=1.9,size=8mm, connect spies}]
 			\node {\includegraphics[width=0.23\textwidth]{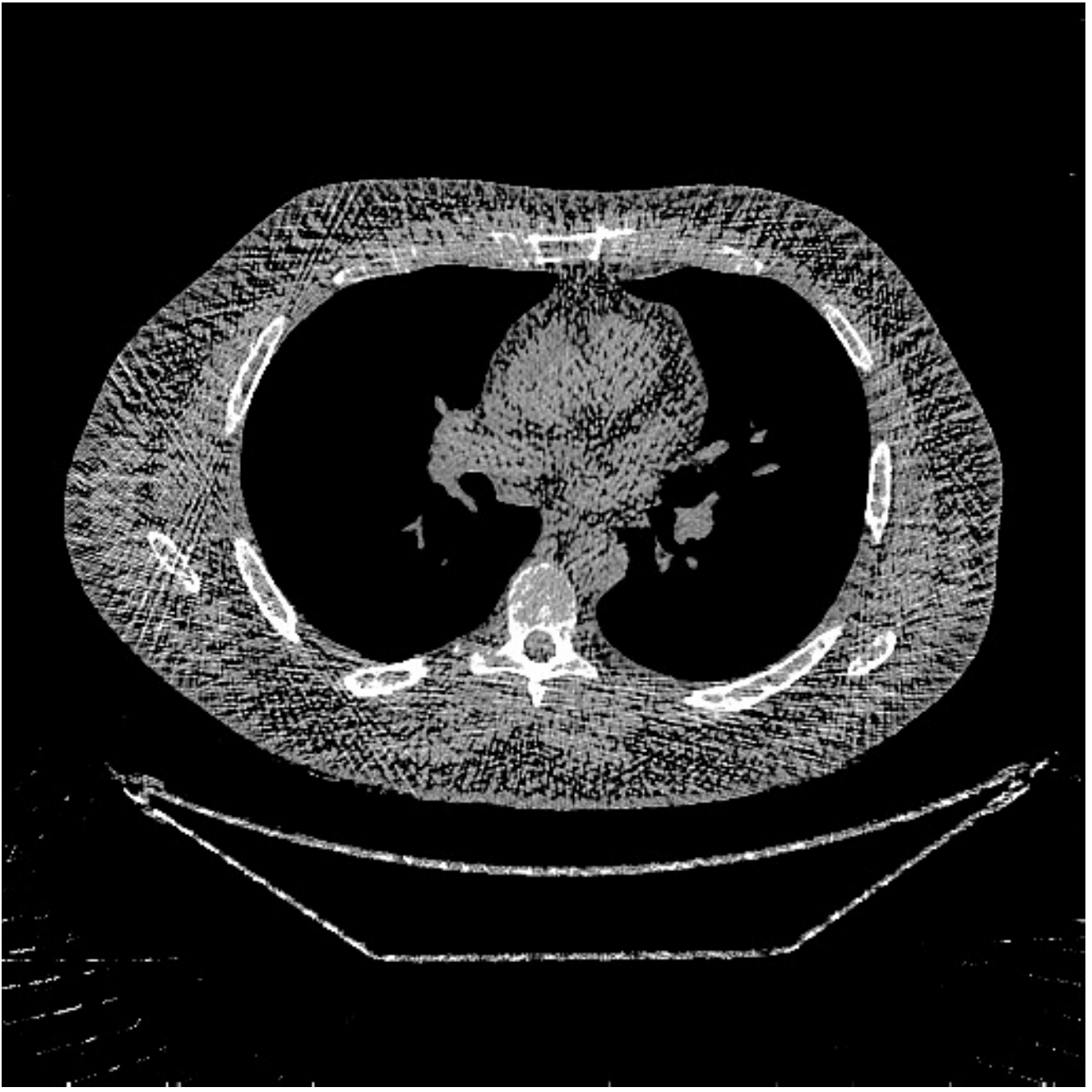}};		
 			\spy on (-0.65,1.1) in node [left] at (-1.2,1.65);
 			\spy on (0.12,1.15) in node [left] at (2,1.65);
 			\spy on (-1.2,-0.4) in node [left] at (-1.2,-1.65);			
 			\spy on (0.22,-0.75) in node [left] at (2,-1.65);			
 			\end{scope}
 			\end{tikzpicture}
 		} 
 		&
 		
 		\raisebox{-.5\height}{
 			\begin{tikzpicture}
 			\begin{scope}[spy using outlines={rectangle,yellow,magnification=1.9,size=8mm, connect spies}]
 			\node {\includegraphics[width=0.23\textwidth]{./2D_GE_Data/GEdata_FBPInit_123L1_lam8e-03_zeta80_30kap1_30kap2_iter2_1000outer_learn110.pdf}};
 			\spy on (-0.65,1.1) in node [left] at (-1.2,1.65);
 			\spy on (0.12,1.15) in node [left] at (2,1.65);
 			\spy on (-1.2,-0.4) in node [left] at (-1.2,-1.65);			
 			\spy on (0.22,-0.75) in node [left] at (2,-1.65);			
 			\end{scope}
 			\end{tikzpicture}
 		} 
 		&		
 		\\
 		
 		&
 		\multicolumn{3}{c}{(b)~GE clinical data}
 		\\

 	\end{tabular}
 	
 	\caption{Generalization capability comparisons between the proposed PWLS-ST-$\ell_1$ MBIR method and FBPConvNet \dquotes{denoising} method \cite{Jin&etal:17TIP} with different number of views (2D fan-beam CT geometry; display window is $[800, 1200]$ HU). 
 			For the clinical data, we applied PWLS-EP reconstruction to full-view ($984$) sinogram to generate the reference image.}
 	\label{fig:fbpconvnet:ref}
 \end{figure*}

\subsubsection{Algorithm Convergence Rate}

Our main concern in convergence rates of Algorithm~\ref{alg:PWLS-ST-l1} lies with an inaccurate preconditioner (e.g., circulant one) particularly for the 3D sparse-view CT reconstructions. 
To see the effects of using a loose preconditioner in Algorithm~\ref{alg:PWLS-ST-l1}, we compared the convergence rates of the 3D case with those of 2D (Fig.~\ref{fig:convg}(a) and Fig.~\ref{fig:convg}(b)). 
In the first $100$ iterations, Algorithm~\ref{alg:PWLS-ST-l1} converges faster in 2D experiments than 3D experiments. However, after $100$ iterations, the convergence rates of Algorithm~\ref{alg:PWLS-ST-l1} are similar in both 2D and 3D reconstructions.
In addition, more PCG (with a circulant preconditioner) iterations does not significantly accelerate Algorithm~\ref{alg:PWLS-ST-l1} (see Fig~\ref{fig:pcg}).
These empirically observations imply that, in the 3D sparse-view CT reconstructions, Algorithm~\ref{alg:PWLS-ST-l1} using a circulant preconditioner ($2$ PCG iterations) is a reasonable choice.

 \begin{figure*}[!t]
 	\centering  
 	\small\addtolength{\tabcolsep}{-7.5pt}
 	
 	\begin{tabular}{ccccc}
 		
 		\raisebox{-.5\height}{\begin{turn}{+90} \small{$246$ views} \end{turn}}~ &
 		
 		\raisebox{-.5\height}{
 			\begin{tikzpicture}
 			\begin{scope}[spy using outlines={rectangle,yellow,magnification=1.9,size=8mm, connect spies}]
 			\node {\includegraphics[width=0.23\textwidth]{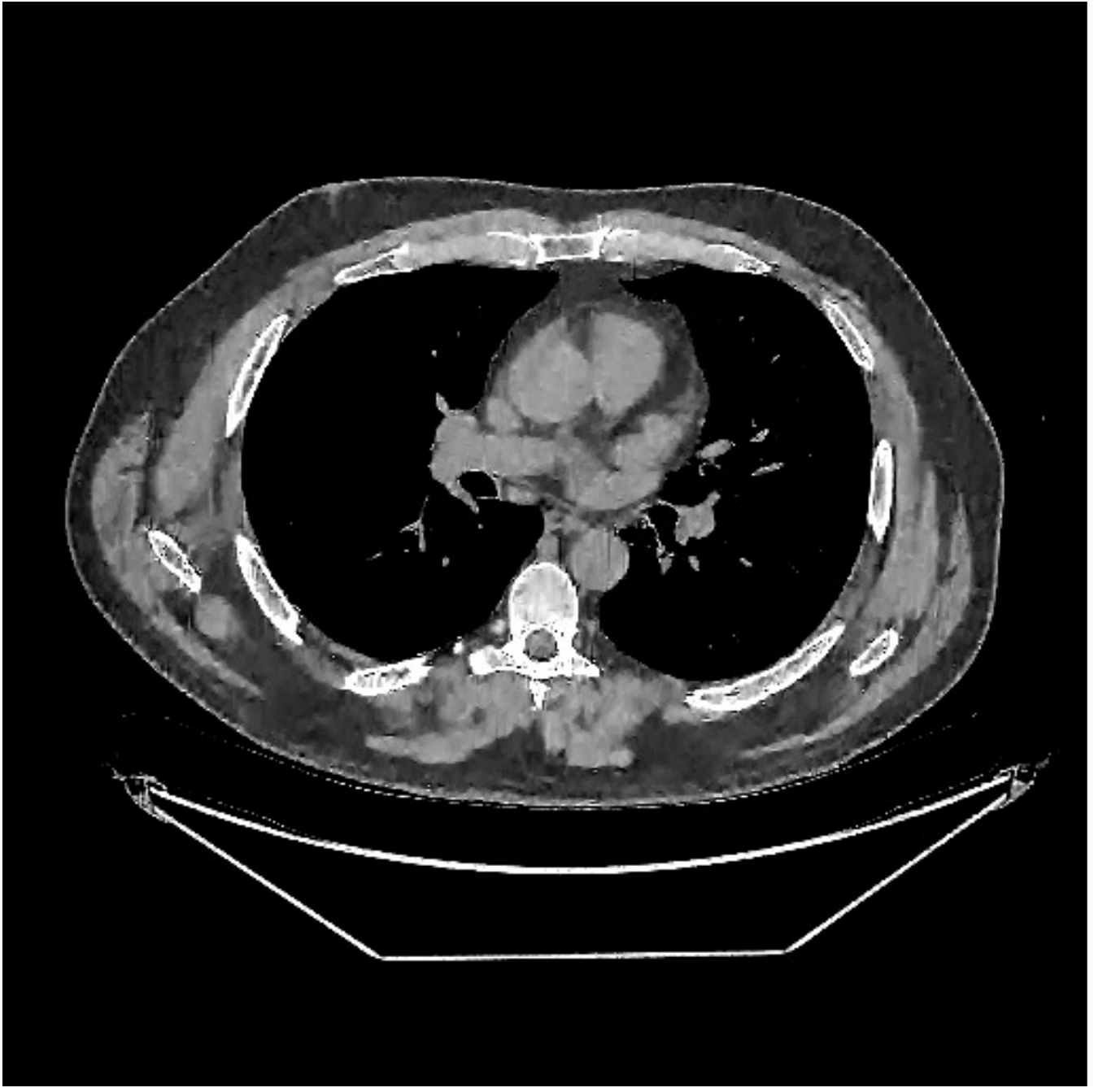}};
 			\spy on (-0.65,1.1) in node [left] at (-1.2,1.65);
 			\spy on (0.12,1.15) in node [left] at (2,1.65);
 			\spy on (-1.2,-0.4) in node [left] at (-1.2,-1.65);			
 			\spy on (0.22,-0.75) in node [left] at (2,-1.65);
 			\end{scope}
 			\end{tikzpicture}} &
 		
 		\raisebox{-.5\height}{
 			\begin{tikzpicture}
 			\begin{scope}[spy using outlines={rectangle,yellow,magnification=1.9,size=8mm, connect spies}]
 			\node {\includegraphics[width=0.23\textwidth]{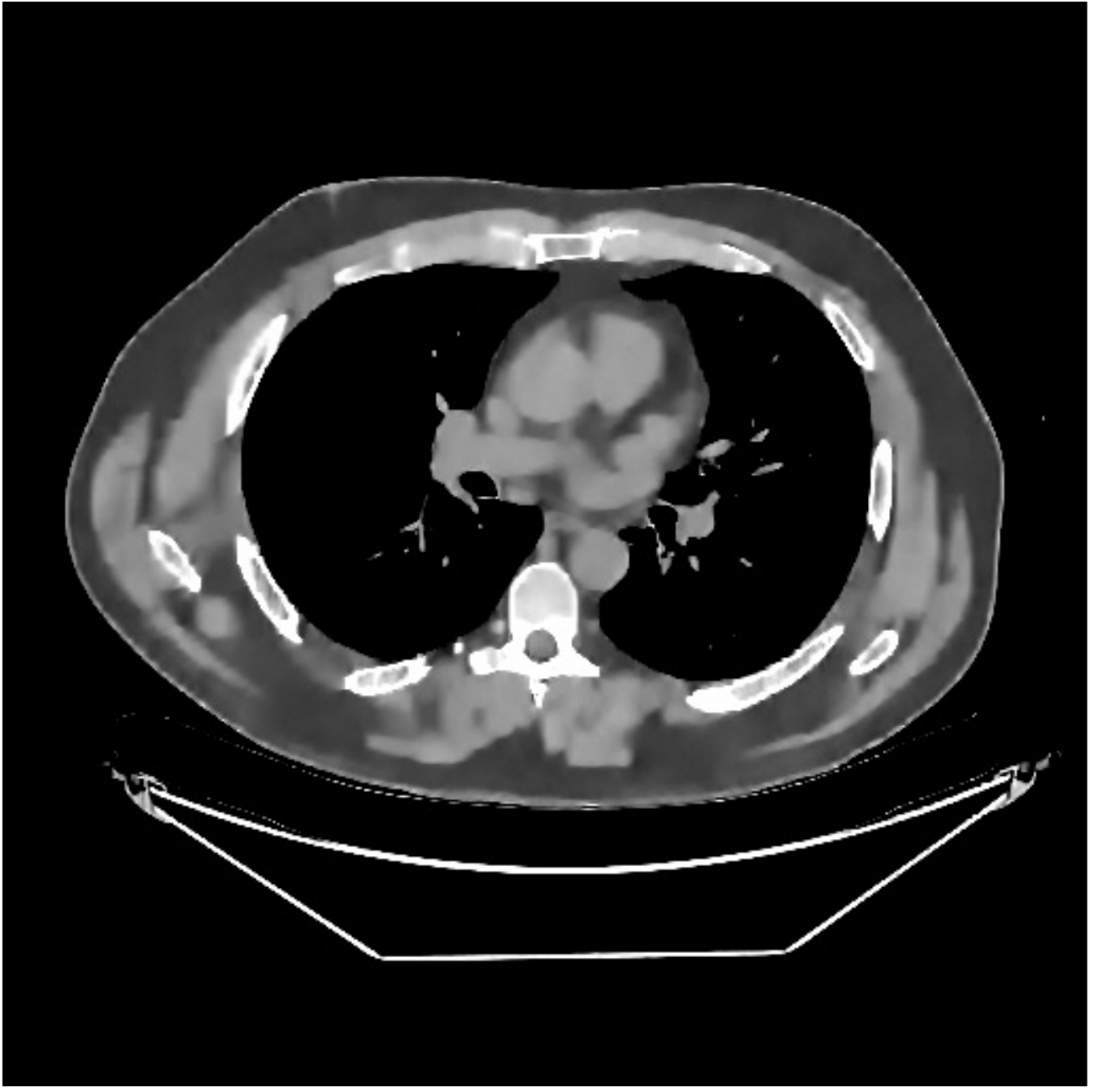}};
 			\spy on (-0.65,1.1) in node [left] at (-1.2,1.65);
 			\spy on (0.12,1.15) in node [left] at (2,1.65);
 			\spy on (-1.2,-0.4) in node [left] at (-1.2,-1.65);			
 			\spy on (0.22,-0.75) in node [left] at (2,-1.65);
 			\end{scope}
 			\end{tikzpicture}}&
 		
 		\raisebox{-.5\height}{
 			\begin{tikzpicture}
 			\begin{scope}[spy using outlines={rectangle,yellow,magnification=1.9,size=8mm, connect spies}]
 			\node {\includegraphics[width=0.23\textwidth]{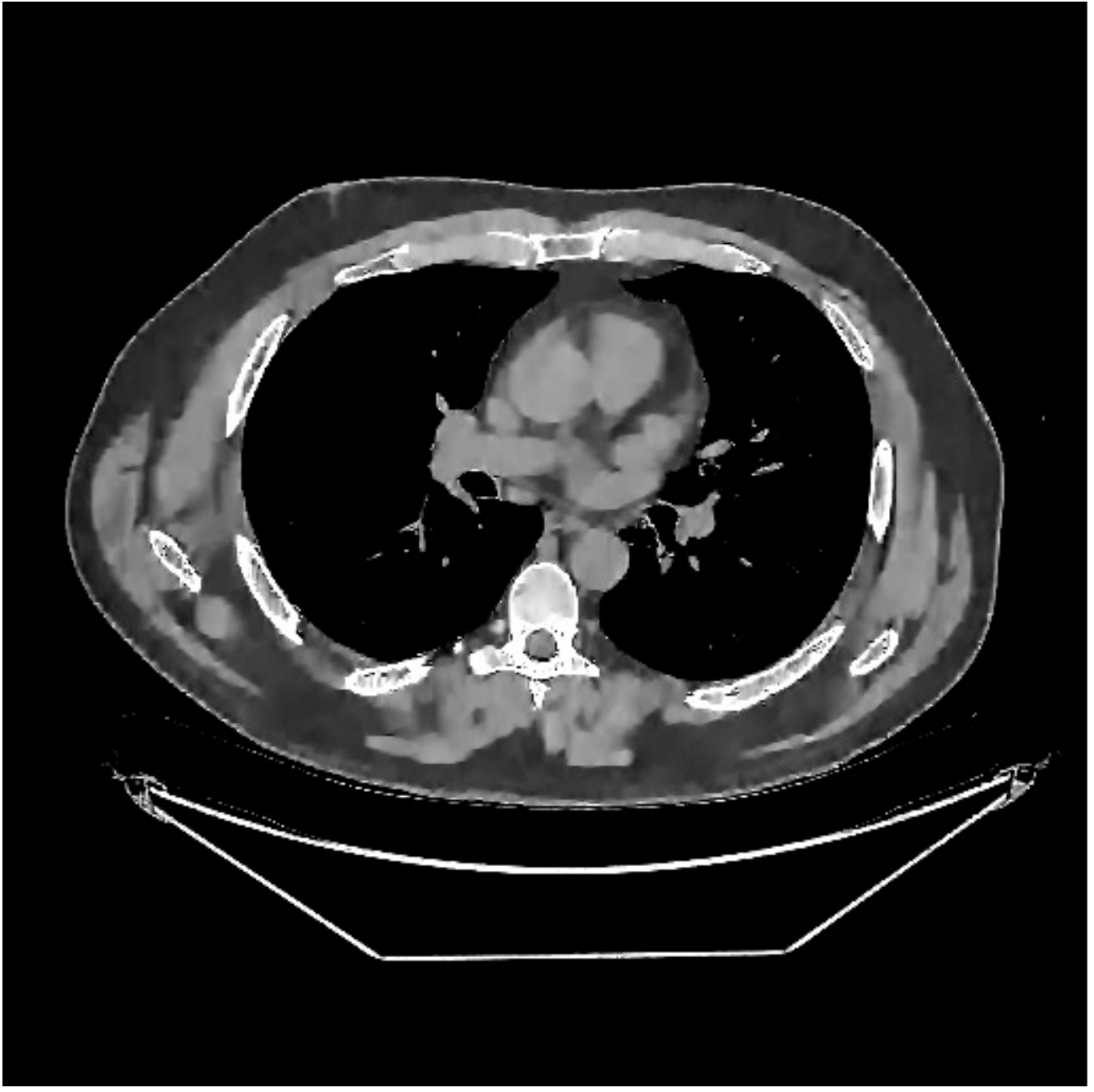}};
 			\spy on (-0.65,1.1) in node [left] at (-1.2,1.65);
 			\spy on (0.12,1.15) in node [left] at (2,1.65);
 			\spy on (-1.2,-0.4) in node [left] at (-1.2,-1.65);			
 			\spy on (0.22,-0.75) in node [left] at (2,-1.65);
 			\end{scope}
 			\end{tikzpicture}}&
 		
 		\raisebox{-.5\height}{
 			\begin{tikzpicture}
 			\begin{scope}[spy using outlines={rectangle,yellow,magnification=1.9,size=8mm, connect spies}]
 			\node {\includegraphics[width=0.23\textwidth]{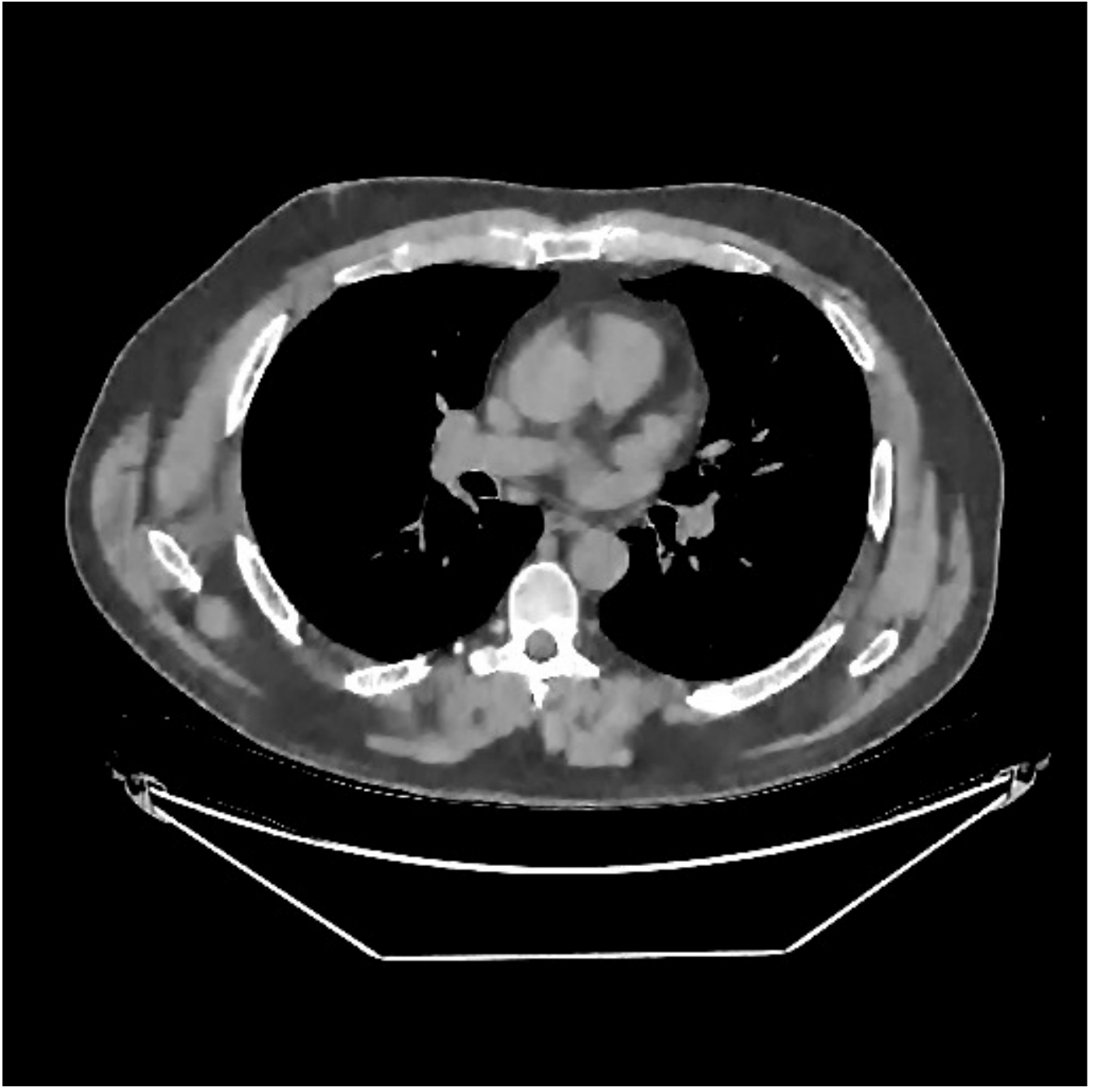}};
 			\spy on (-0.65,1.1) in node [left] at (-1.2,1.65);
 			\spy on (0.12,1.15) in node [left] at (2,1.65);
 			\spy on (-1.2,-0.4) in node [left] at (-1.2,-1.65);			
 			\spy on (0.22,-0.75) in node [left] at (2,-1.65);
 			\end{scope}
 			\end{tikzpicture}} \\
 		
 		{}  
		& 
		\small{(a) $\lambda =  5 \!\times\! 10^{-3}$, $\gamma/\lambda = 80$} 
		& 
		\small{(b) $\lambda =  2 \!\times\! 10^{-2}$, $\gamma/\lambda = 80$} 
		& 
		\small{(c) $\lambda =   10^{-2}$, $\gamma/\lambda = 50$}
		& 
		\small{(d) $\lambda =   10^{-2}$, $\gamma/\lambda = 200$} 
		
 		
 	\end{tabular}
 	
 	\caption{Comparison of reconstructed images from clinical data for the proposed PWLS-ST-$\ell_1$ method with different combinations of regularization parameters $\lambda$ and $\gamma$ (2D fan-beam CT geometry; display window is $[800, 1200]$ HU).}
 	\label{fig:2D_GEdata_sensitivity}
 \end{figure*}

\subsubsection{Generalization Capability Comparisons between a \dquotes{Denoising} Deep NN and the Proposed PWLS-ST-$\ell_1$ Method}
\label{subsubsec:generalization}

This section compares the generalization capabilities between the proposed MBIR method, PWLS-ST-$\ell_1$, and a denoising deep NN, FBPConvNet \cite{Jin&etal:17TIP}, that are trained from the phantom data; in particular, we tested the trained PWLS-ST-$\ell_1$ and FBPConvNet models to phantom and clinical scan data.
The results in Fig.~\ref{fig:fbpconvnet:ref} show that the non-MBIR FBPConvNet method has higher overfitting risks, compared to the proposed PWLS-ST-$\ell_1$ MBIR method.
When tested on clinical scan data, 
PWLS-ST-$\ell_1$ achieves much more accurate reconstruction, compared to FBPConvNet. 
See Fig.~\ref{fig:fbpconvnet:ref}(b).
When tested on phantom data, 
FBPConvNet generates more unnatural features as the number of views reduces,
although it gives lower RMSE values compared to PWLS-ST-$\ell_1$. 
See zoom-ins in Fig.~\ref{fig:fbpconvnet:ref}(a).
The FBPConvNet results above correspond to those in the recent work \cite{Chen&etal:18TMI}
that FBPConvNet \cite{Jin&etal:17TIP} generated some unexpected structures.

\subsubsection{Parameter Selection and Sensitivity of the Proposed PWLS-ST-$\ell_1$ Method}
\label{sec:param}




This section describes our parameter selection strategy for PWLS-ST-$\ell_1$ reconstruction, 
and discusses its parameter sensitivity. 
Our strategy to select its parameters, $\{ \lambda, \gamma/\lambda, \kappa_{\mathrm{des},\nu}, \kappa_{\mathrm{des},\mu} \}$, in the XCAT phantom experiments is given as follows.
We first chose the hard-shrinkage parameter $\gamma/\lambda$ according to the sparsity based guideline described in \cite{Zheng&etal:17arXiv}
(specifically, the percentages of non-zero elements in the sparse codes after some outer iterations of Algorithm~\ref{alg:PWLS-ST-l1} are $ 4-5$\%);
given chosen $\gamma/\lambda$, we ran a coarse grid search for selecting the regularization parameter $\lambda$ and
the desired condition numbers $\kappa_{\mathrm{des},\nu} \!=\! \kappa_{\mathrm{des},\mu}$ in Section~\ref{sec:param_select}.
(In particular, we found that $\kappa_{\mathrm{des},\nu}, \kappa_{\mathrm{des},\mu} \in [10,50]$ are reasonable for fast and stable convergence of Algorithm~\ref{alg:PWLS-ST-l1}.)

Our strategy to select the regularization parameter $\lambda$ for new data, 
e.g., the GE clinical data in Section~\ref{sec:exp:fan:imaging},
is given as follows.
Given fixed CT geometry, we first compute diagonal majorizers for $\mb{A}^T \mb{W} \mb{A}$ in \R{sys:L1trsf} (see details in \cite{Chun&Fessler:18Asilomar}) for both the phantom and clinical data, 
and calculate the mean values of their diagonal elements within a circular ROI.
Next, we apply the ratio of these two mean values to the chosen regularization parameter from the phantom experiments, 
and obtain $\lambda$ for the clinical data experiments.
This procedure aims that the selected $\lambda$ values give similar regularization strength 
-- particularly across the pixels (or voxels) -- to both MBIRs in phantom and clinical data experiments.
We found that the other hyperparameters $\{ \gamma/\lambda, \kappa_{\mathrm{des},\nu}, \kappa_{\mathrm{des},\mu} \}$ 
chosen in the phantom experiments work well in the clinical data experiments.

Fig.~\ref{fig:2D_GEdata_sensitivity} studies 
the influence of regularization parameters $\lambda$ and $\gamma$ on PWLS-ST-$\ell_1$.
Given a fixed hard-shrinkage parameter $\gamma/\lambda$, 
a larger $\lambda$ value better removes noise (or unwanted artifacts), 
but too large $\lambda$ can oversmooth reconstructed images;
compare Fig.~\ref{fig:2D_GEdata_sensitivity}(a) and Fig.~\ref{fig:2D_GEdata_sensitivity}(b). 
Given a fixed regularization parameter $\lambda$, 
a larger $\gamma$ value leads to lower sparsity in sparse codes and achieves better noise reduction, 
but too large $\gamma$ can remove some edges (e.g., in bone regions);
compare Fig.~\ref{fig:2D_GEdata_sensitivity}(c) and Fig.~\ref{fig:2D_GEdata_sensitivity}(d). 
In particular, Fig.~\ref{fig:2D_GE_thres} in the supplement shows that once the $\lambda$ value is properly chosen, 
PWLS-ST-$\ell_1$ is robust to a wide range of $\gamma$ values.

\section{Conclusion}\label{sec:conclusion}

We presented a new MBIR approach for sparse-view CT, PWLS-ST-$\ell_1$ that combines PWLS reconstruction and $\ell_1$ prior with a learned ST.
In addition, we analyzed the empirical MSE for the image update estimator, and the MMSE for the $\ell_2$-norm relaxed image update estimator of the proposed PWLS-ST-$\ell_1$ model: the analysis reveals that as the \dquotes{denoised} sparse codes approach those of the true signal in the learned transform domain, one can obtain better image reconstruction.
We introduced an efficient ADMM-based algorithm for the proposed PWLS-ST-$\ell_1$ model, with a new ADMM parameter selection scheme based on (approximated) condition numbers.
This scheme provided fast and stable convergence in our experiments and helped tuning ADMM parameters of the proposed algorithm for different datasets and different CT imaging geometries.

For sparse-view 2D fan-beam CT and 3D axial cone-beam CT, PWLS-ST-$\ell_1$ significantly improves the reconstruction quality compared to conventional methods, such as FBP and PWLS-EP.
The comparisons between PWLS-ST-$\ell_1$ and the existing PWLS-ST-$\ell_2$ model suggest that, model mismatch exists between the training model~\R{eq:l1LearnSpTrasf} and the $\ell_2$ prior used in PWLS-ST-$\ell_2$. 
The proposed PWLS-ST-$\ell_1$ method using $\ell_1$ prior can moderate this model mismatch, and achieved more accurate reconstructions than PWLS-ST-$\ell_2$.
Our results with the XCAT phantom data and clinical data show that, for sparse-view 2D fan-beam CT, PWLS-ST-$\ell_1$ using a learned \textit{square} ST achieves comparable or better image quality and is much faster compared to the existing PWLS-DL method using learned \textit{overcomplete} dictionary. 
Our results with clinical data indicate that, deep \dquotes{denoising} NNs  (e.g., FBPConvNet \cite{Jin&etal:17TIP}) can have overfitting risks, while MBIR methods trained in an unsupervised way do not suffer from overfitting, and give more stable reconstruction. 

Future work will explore PWLS-ST-$\ell_1$ with the technique of controlling local spatial resolution or noise in the reconstructed images \cite{Fessler&Rogers:96TIP, Cho&Fessler:15TMI} to further reduce blur, particularly around the center of reconstructed images (see \cite[Appx.]{Chun&etal:17Fully3D} and \cite{Zheng&etal:17arXiv}). 
On the algorithmic side, to more rapidly solve the block multi-nonconvex problem \R{sys:L1trsf}, we plan to apply \textit{block proximal gradient method using majorizer} \cite{Chun&Fessler:18cao, chun:18:cdl} that guarantees convergence to critical points, or design a more accurate preconditioner that allows the parameter selection scheme in Section~\ref{sec:param_select}.

\begin{appendices}

\section{Notation} \label{sec:notation}

Bold capital letters represent matrices, and bold lowercase letters are used for vectors (all vectors are column vectors).  Italic type is used for all letters representing  variables, parameters, and elements of matrices and vectors.  
We use $\nm{\cdot}_{p}$ to denote the $\ell_p$-norm and write $\ip{\cdot}{\cdot}$ for the standard inner product on $\bbC^N$.  
The weighted $\ell_2$-norm with a Hermitian positive definite matrix $\mb{A}$ is denoted by $\nm{\cdot}_{\mb{A}} = \nm{ \mb{A}^{1/2} (\cdot) }_2$.
$\nm{\cdot}_{0}$ denotes the $\ell_0$-norm, i.e., the number of nonzeros of a vector.  
The Frobenius norm of a matrix $\mb{A}$ is denoted by $\nm{ \mb{A} }_F$.
$( \cdot )^T$, $( \cdot )^H$ indicate the transpose and complex conjugate transpose (Hermitian transpose), respectively. 
$\sgn(\cdot)$ and $\det(\cdot)$ denote the sign function and determinant of a matrix, respectively.
For self-adjoint matrices $\mb{A},\mb{B} \in \bbC^{N \times N}$, the notation $\mb{B} \preceq \mb{A}$ denotes that $\mb{A}-\mb{B}$ is a positive semi-definite matrix.

\end{appendices}


\section*{Acknowledgments}

The authors thank GE Healthcare for providing the clinical data.

\bibliographystyle{IEEEtran}
\bibliography{referenceBibs_Bobby}

\begin{thebibliography}{10}
\providecommand{\url}[1]{#1}
\csname url@samestyle\endcsname
\providecommand{\newblock}{\relax}
\providecommand{\bibinfo}[2]{#2}
\providecommand{\BIBentrySTDinterwordspacing}{\spaceskip=0pt\relax}
\providecommand{\BIBentryALTinterwordstretchfactor}{4}
\providecommand{\BIBentryALTinterwordspacing}{\spaceskip=\fontdimen2\font plus
\BIBentryALTinterwordstretchfactor\fontdimen3\font minus
  \fontdimen4\font\relax}
\providecommand{\BIBforeignlanguage}[2]{{%
\expandafter\ifx\csname l@#1\endcsname\relax
\typeout{** WARNING: IEEEtran.bst: No hyphenation pattern has been}%
\typeout{** loaded for the language `#1'. Using the pattern for}%
\typeout{** the default language instead.}%
\else
\language=\csname l@#1\endcsname
\fi
#2}}
\providecommand{\BIBdecl}{\relax}
\BIBdecl

\bibitem{Chen&Tang&Leng:08MP}
G.~H. Chen, J.~Tang, and S.~Leng, ``Prior image constrained compressed sensing
  ({PICCS}): a method to accurately reconstruct dynamic {CT} images from highly
  undersampled projection data sets,'' \emph{Med. Phys.}, vol.~35, no.~2, pp.
  660--663, Feb. 2008.

\bibitem{Chun&Talavage:13Fully3D}
I.~Y. Chun and T.~Talavage, ``Efficient compressed sensing statistical
  {X}-ray/{CT} reconstruction from fewer measurements,'' in \emph{Proc.
  $12^{\text{th}}$ Intl. Mtg. on Fully 3D Image Recon. in Rad. and Nuc. Med},
  Lake Tahoe, CA, Jun. 2013, pp. 30--33.

\bibitem{Foucart&Rauhut:book}
S.~Foucart and H.~Rauhut, \emph{A mathematical introduction to compressive
  sensing}.\hskip 1em plus 0.5em minus 0.4em\relax New York, NY: Springer,
  2013.

\bibitem{Adcock&etal:05bookCh}
B.~Adcock, A.~C. Hansen, and B.~Roman, ``The quest for optimal sampling:
  Computationally efficient, structure-exploiting measurements for compressed
  sensing,'' in \emph{Compressed Sensing and its Applications}, ser. Applied
  and Numerical Harmonic Analysis.\hskip 1em plus 0.5em minus 0.4em\relax
  Birkh{ä}user, Cham, 2015, pp. 143--167.

\bibitem{Chun&Adcock:17TIT}
I.~Y. Chun and B.~Adcock, ``Compressed sensing and parallel acquisition,''
  \emph{IEEE Trans. Inf. Theory}, vol.~63, no.~7, pp. 1--23, May 2017.

\bibitem{Sidky&Kao&Pan:06}
E.~Y. Sidky, C.-M. Kao, and X.~Pan, ``Accurate image reconstruction from
  few-views and limited-angle data in divergent-beam {CT},'' \emph{J. X-ray
  Sci. Technol.}, vol.~14, no.~2, pp. 119--139, 2006.

\bibitem{Yu&Wang:09PMB}
H.~Yu and G.~Wang, ``Compressed sensing based interior tomography,''
  \emph{Phys. Med. Biol.}, vol.~54, no.~9, pp. 2791--2805, May 2009.

\bibitem{Bian&etal:10PMB}
J.~Bian, J.~H. Siewerdsen, X.~Han, E.~Y. Sidky, J.~L. Prince, C.~A. Pelizzari,
  and X.~Pan, ``Evaluation of sparse-view reconstruction from
  flat-panel-detector cone-beam {CT},'' \emph{Phys. Med. Biol.}, vol.~55,
  no.~22, p. 6575, Oct. 2010.

\bibitem{Ramani&Fessler:12MI}
S.~Ramani and J.~A. Fessler, ``A splitting-based iterative algorithm for
  accelerated statistical {X}-ray {CT} reconstruction,'' \emph{IEEE Trans. Med.
  Imag.}, vol.~31, no.~3, pp. 677--688, Mar. 2012.

\bibitem{Nie&etal:14PMB}
S.~Niu, Y.~Gao, Z.~Bian, J.~Huang, W.~Chen, G.~Yu, Z.~Liang, and J.~Ma,
  ``Sparse-view {X}-ray {CT} reconstruction via total generalized variation
  regularization,'' \emph{Phys. Med. Biol.}, vol.~59, no.~12, p. 2997, May
  2014.

\bibitem{Chen&etal:17TMI}
H.~Chen, Y.~Zhang, M.~K. Kalra, F.~Lin, P.~Liao, J.~Zhou, and G.~Wang,
  ``Low-dose {CT} with a residual encoder-decoder convolutional neural network
  ({RED}-{CNN}),'' \emph{IEEE Trans. Med. Imag.}, vol.~36, no.~12, pp.
  2524--2535, Jun. 2017.

\bibitem{Kang&Min&Ye:17MP}
E.~Kang, J.~Min, and J.~C. Ye, ``A deep convolutional neural network using
  directional wavelets for low-dose {X}-ray {CT} reconstruction,'' \emph{Med.
  Phys.}, vol.~44, no.~10, pp. e360--e375, Oct. 2017.

\bibitem{Wolterink&etal:17TMI}
J.~M. Wolterink, T.~Leiner, M.~A. Viergever, and I.~Isgum, ``Generative
  adversarial networks for noise reduction in low-dose {CT},'' \emph{IEEE
  Trans. Med. Imag.}, vol.~36, no.~12, pp. 2536--2545, May 2017.

\bibitem{Jin&etal:17TIP}
K.~H. Jin, M.~T. McCann, E.~Froustey, and M.~Unser, ``Deep convolutional neural
  network for inverse problems in imaging,'' \emph{IEEE Trans. Image Process.},
  vol.~26, no.~9, pp. 4509--4522, Sep. 2017.

\bibitem{Ye&Han&Cha:18SJIS}
J.~Ye, Y.~Han, and E.~Cha, ``Deep convolutional framelets: A general deep
  learning framework for inverse problems,'' \emph{SIAM J. Imaging Sci.},
  vol.~11, no.~2, pp. 991--1048, Apr. 2018.

\bibitem{Chen&etal:18TMI}
H.~Chen, Y.~Zhang, W.~Zhang, H.~Sun, P.~Liao, K.~He, J.~Zhou, and G.~Wang,
  ``{LEARN:} {Learned} experts' assessment-based reconstruction network for
  sparse-data {CT},'' \emph{IEEE Trans. Med. Imag.}, vol.~37, no.~6, pp.
  1333--1347, Jun. 2018.

\bibitem{wu:17:ild}
D.~Wu, K.~Kim, G.~E. Fakhri, and Q.~Li, ``Iterative low-dose {CT}
  reconstruction with priors trained by artificial neural network,''
  \emph{{IEEE Trans. Med. Imag.}}, vol.~36, no.~12, pp. {2479--2486}, Dec.
  2017.

\bibitem{Chun&Fessler:18IVMSP}
I.~Y. Chun and J.~A. Fessler, ``Deep {BCD}-net using identical
  encoding-decoding {CNN} structures for iterative image recovery,'' in
  \emph{Proc. IEEE IVMSP Workshop}, Zagori, Greece, Jun. 2018, pp. 1--5.

\bibitem{Chun&etal:18Allerton}
I.~Y. Chun, H.~Lim, Z.~Huang, and J.~A. Fessler, ``Fast and convergent
  iterative signal recovery using trained convolutional neural networkss,'' in
  \emph{Proc. Allerton Conf. on Commun., Control, and Comput.}, Allerton, IL,
  Oct. 2018, pp. 155--159.

\bibitem{Chun&etal:18arXiv:momnet}
\BIBentryALTinterwordspacing
I.~Y. Chun, Z.~Huang, H.~Lim, and J.~A. Fessler, ``{Momentum-Net}: Fast and
  convergent iterative neural network for inverse problems,''
  \emph{\emph{submitted}}, Jul. 2019. [Online]. Available:
  \url{http://arxiv.org/abs/1907.11818}
\BIBentrySTDinterwordspacing

\bibitem{lehtinen:18:nli}
J.~Lehtinen, J.~Munkberg, J.~Hasselgren, S.~Laine, T.~Karras, M.~Aittala, and
  T.~Aila, ``{Noise2Noise:} learning image restoration without clean data,'' in
  \emph{{Proc. Intl. Conf. Mach. Learn}}, 2018, pp. {2971--2980}.

\bibitem{pelt:18:itr}
D.~Pelt, K.~Batenburg, and J.~Sethian, ``Improving tomographic reconstruction
  from limited data using mixed-scale dense convolutional neural networks,''
  \emph{Journal of Imaging}, vol.~4, no.~11, p. 128, 2018.

\bibitem{Yuan&etal:19Fully3D}
N.~Yuan, J.~Zhou, and J.~Qi, ``Low-dose {CT} image denoising without high-dose
  reference images,'' in \emph{Proc. $15^{\text{th}}$ Intl. Mtg. on Fully 3D
  Image Recon. in Rad. and Nuc. Med}, Philadelphia, United States, Jun. 2019,
  p. 110721C.

\bibitem{Chun&Fessler:18cao}
\BIBentryALTinterwordspacing
I.~Y. Chun and J.~A. Fessler, ``Convolutional analysis operator learning:
  {A}cceleration and convergence,'' \emph{{IEEE Trans. Im. Proc.} \emph{(to
  appear)}}, Jan. 2019. [Online]. Available:
  \url{https://arxiv.org/abs/1802.05584}
\BIBentrySTDinterwordspacing

\bibitem{Chun&etal:19SPL}
\BIBentryALTinterwordspacing
I.~Y. Chun, D.~Hong, B.~Adcock, and J.~A. Fessler, ``Convolutional analysis
  operator learning: Dependence on training data,'' \emph{IEEE Signal Process.
  Lett.}, vol.~26, no.~8, pp. 1137--1141, Jun. 2019. [Online]. Available:
  \url{http://arxiv.org/abs/1902.08267}
\BIBentrySTDinterwordspacing

\bibitem{chun:18:cdl}
I.~Y. Chun and J.~A. Fessler, ``Convolutional dictionary learning: acceleration
  and convergence,'' \emph{{IEEE Trans. Im. Proc.}}, vol.~27, no.~4, pp.
  {1697--712}, Apr. 2018.

\bibitem{Chun&Fessler:17SAMPTA}
------, ``Convergent convolutional dictionary learning using adaptive contrast
  enhancement ({CDL-ACE}): Application of {CDL} to image denoising,'' in
  \emph{Proc. $12^{\textmd{th}}$ Sampling Theory and Appl. (SampTA)}, Tallinn,
  Estonia, Jul. 2017, pp. 460--464.

\bibitem{Aharon&Elad&Bruckstein:06TSP}
M.~Aharon, M.~Elad, and A.~Bruckstein, ``\textit{K}-{SVD}: An algorithm for
  designing overcomplete dictionaries for sparse representation,'' \emph{IEEE
  Trans. Signal Process.}, vol.~54, no.~11, pp. 4311--4322, Nov. 2006.

\bibitem{Cai&etal:14ACHA}
J.-F. Cai, H.~Ji, Z.~Shen, and G.-B. Ye, ``Data-driven tight frame construction
  and image denoising,'' \emph{Appl. Comput. Harmon. A.}, vol.~37, no.~1, pp.
  89--105, Oct. 2014.

\bibitem{Ravishankar&Bresler:15TSP}
S.~Ravishankar and Y.~Bresler, ``$\ell_0$ sparsifying transform learning with
  efficient optimal updates and convergence guarantees,'' \emph{IEEE Trans.
  Signal Process.}, vol.~63, no.~9, pp. 2389--2404, May 2015.

\bibitem{Xu&etal:12TMI}
Q.~Xu, H.~Yu, X.~Mou, L.~Zhang, J.~Hsieh, and G.~Wang, ``Low-dose {X}-ray {CT}
  reconstruction via dictionary learning,'' \emph{IEEE Trans. Med. Imag.},
  vol.~31, no.~9, pp. 1682--1697, Sep. 2012.

\bibitem{pfister:14:mbi}
L.~Pfister and Y.~Bresler, ``Model-based iterative tomographic reconstruction
  with adaptive sparsifying transforms,'' in \emph{Proc. SPIE}, vol. 9020,
  2014, pp. {90\,200H--1--90\,200H--11}.

\bibitem{Zhang&etal:16BEO}
C.~Zhang, T.~Zhang, M.~Li, C.~Peng, Z.~Liu, and J.~Zheng, ``Low-dose {CT}
  reconstruction via {L}1 dictionary learning regularization using iteratively
  reweighted least-squares,'' \emph{Biomed. Eng. OnLine}, vol.~15, no.~1,
  p.~66, Jun. 2016.

\bibitem{Zheng&etal:16IVMSP}
X.~Zheng, Z.~Lu, S.~Ravishankar, Y.~Long, and J.~A. Fessler, ``Low dose {CT}
  image reconstruction with learned sparsifying transform,'' in \emph{Proc.
  $2016$ IEEE IVMSP}, Bordeaux, France, Jul. 2016, pp. 1--5.

\bibitem{Zheng&etal:17Fully3D}
X.~Zheng, S.~Ravishankar, Y.~Long, and J.~A. Fessler, ``Union of learned
  sparsifying transforms based low-dose {3D} {CT} image reconstruction,'' in
  \emph{Proc. $14^{\text{th}}$ Intl. Mtg. on Fully 3D Image Recon. in Rad. and
  Nuc. Med}, Xi'an, China, Jun. 2017, pp. 69--72.

\bibitem{Zheng&etal:17arXiv}
------, ``{PWLS-ULTRA}: An efficient clustering and learning-based approach for
  low-dose {3D} {CT} image reconstruction,'' \emph{IEEE Trans. Med. Imag.},
  vol.~37, no.~6, pp. 1498--1510, Jun. 2018.

\bibitem{Lu&etal:13CVPR}
C.~Lu, J.~Shi, and J.~Jia, ``Online robust dictionary learning,'' in
  \emph{Proc. $2013$ IEEE CVPR}, Portland, OR, Jun. 2013, pp. 415--422.

\bibitem{Jiang&etal:15IJCAI}
W.~Jiang, F.~Nie, and H.~Huang, ``Robust dictionary learning with capped
  $\ell_1$-norm,'' in \emph{Proc. $2015$ IJCAI}, Buenos Aires, Argentina, Jul.
  2015, pp. 3590--3596.

\bibitem{Boyd&Parikh&Chu&Peleato&Eckstein:11FTML}
S.~Boyd, N.~Parikh, E.~Chu, B.~Peleato, and J.~Eckstein, ``Distributed
  optimization and statistical learning via the alternating direction method of
  multipliers,'' \emph{Found. \& Trends in Machine Learning}, vol.~3, no.~1,
  pp. 1--122, Jan. 2011.

\bibitem{Segars&etal:08MP}
W.~P. Segars, M.~Mahesh, T.~J. Beck, E.~C. Frey, and B.~M.~W. Tsui, ``Realistic
  {CT} simulation using the {4D} {XCAT} phantom,'' \emph{Med. Phys.}, vol.~35,
  no.~8, pp. 3800--3808, Jul. 2008.

\bibitem{Chun&etal:17Fully3D}
I.~Y. Chun, X.~Zheng, Y.~Long, and J.~A. Fessler, ``Sparse-view {X}-ray {CT}
  reconstruction using $\ell_1$ regularization with learned sparsifying
  transform,'' in \emph{Proc. $14^{\text{th}}$ Intl. Mtg. on Fully 3D Image
  Recon. in Rad. and Nuc. Med}, Xi'an, China, Jun. 2017, pp. 115--119.

\bibitem{Thibault&Bouman&Sauer&Hsieh:06SPIE}
J.~B. Thibault, C.~A. Bouman, K.~D. Sauer, and J.~Hsieh, ``A recursive filter
  for noise reduction in statistical iterative tomographic imaging,'' in
  \emph{Proc. SPIE 6065, Computational Imaging IV}, vol. 6065, Feb. 2006, p.
  60650X.

\bibitem{Chun&Adcock&Talavage:15TMI}
I.~Y. Chun, B.~Adcock, and T.~M. Talavage, ``Efficient compressed sensing
  {SENSE} {pMRI} reconstruction with joint sparsity promotion,'' \emph{IEEE
  Trans. Med. Imag.}, vol.~35, no.~1, pp. 354--368, Jan. 2016.

\bibitem{Fessler&Booth:99IP}
J.~A. Fessler and S.~D. Booth, ``Conjugate-gradient preconditioning methods for
  shift-variant {PET} image reconstruction,'' \emph{IEEE Trans. Image
  Process.}, vol.~8, no.~5, pp. 688--699, May 1999.

\bibitem{Booth&Fessler:95ICIP}
S.~D. Booth and J.~A. Fessler, ``Combined diagonal/{F}ourier preconditioning
  methods for image reconstruction in emission tomography,'' in \emph{Proc.
  $1995$ ICIP}, vol.~2, Washington, DC, Oct. 1995, pp. 441--444.

\bibitem{Fu&etal:13Fully3D}
L.~Fu, Z.~Yu, J.-B. Thibault, B.~De~Man, M.~McGaffin~G., and J.~A. Fessler,
  ``Space-variant channelized preconditioner design for {3D} iterative {CT}
  reconstruction,'' in \emph{Proc. $12^{\text{th}}$ Intl. Mtg. on Fully 3D
  Image Recon. in Rad. and Nuc. Med}, Lake Tahoe, CA, Jun. 2013, pp. 205--208.

\bibitem{Fu&etal:17Fully3D}
L.~Fu, J.~A. Fessler, P.~E. Kinahan, and B.~De~Man, ``Combining non-diagonal
  preconditioning and ordered-subsets for iterative {CT} reconstruction,'' in
  \emph{Proc. $14^{\text{th}}$ Intl. Mtg. on Fully 3D Image Recon. in Rad. and
  Nuc. Med}, Xi'an, China, Jun. 2017, pp. 760--766.

\bibitem{Feldkamp&David&Kress:84JOSAA}
L.~A. Feldkamp, L.~C. Davis, and J.~W. Kress, ``Practical cone beam
  algorithm,'' \emph{J. Opt. Soc. Am. A}, vol.~1, no.~6, pp. 612--9, Jun. 1984.

\bibitem{Cho&Fessler:15TMI}
J.~H. Cho and J.~A. Fessler, ``Regularization designs for uniform spatial
  resolution and noise properties in statistical image reconstruction for 3-{D}
  {X}-ray {CT},'' \emph{IEEE Trans. Med. Imag.}, vol.~2, no.~34, pp. 678--689,
  Feb. 2015.

\bibitem{Nien&Fessler:16:TMI}
H.~Nien and J.~A. Fessler, ``Relaxed linearized algorithms for faster {X-ray}
  {CT} image reconstruction,'' \emph{IEEE Trans. Med. Imag.}, vol.~35, no.~4,
  pp. 1090--1098, Apr. 2016.

\bibitem{erdogan:99:osa}
H.~Erdo\u{g}an and J.~A. Fessler, ``Ordered subsets algorithms for transmission
  tomography,'' \emph{Phys. Med. Biol.}, vol.~44, no.~11, pp. 2835--2851, Nov.
  1999.

\bibitem{Chun&Fessler:18Asilomar}
I.~Y. Chun and J.~A. Fessler, ``Convolutional analysis operator learning:
  Application to sparse-view {CT},'' in \emph{Proc. Asilomar Conf. on Signals,
  Syst., and Comput.}, Pacific Grove, CA, Oct. 2018, pp. 1631--1635.

\bibitem{Fessler&Rogers:96TIP}
J.~A. Fessler and W.~L. Rogers, ``Spatial resolution properties of
  penalized-likelihood image reconstruction methods: {S}pace-invariant
  tomographs,'' \emph{IEEE Trans. Image Process.}, vol.~5, no.~9, pp.
  1346--1358, Sep. 1996.

\bibitem{Zheng&etal:19L1ST}
X.~Zheng, I.~Y. Chun, Z.~Li, Y.~Long, and J.~A. Fessler, ``Sparse-view {X}-ray
  {CT} reconstruction using $\ell_1$ prior with learned transform,'' \emph{IEEE
  Trans. Computational Imaging}, 2019, \emph{submitted}.

\end{thebibliography}

{
	\twocolumn[
	\begin{center}
		\Huge Sparse-View X-Ray CT Reconstruction \\ Using $\ell_1$ Prior with Learned Transform \\ -- Supplementary Material
	\end{center}]
}

\setcounter{table}{0}
\setcounter{section}{0}
\setcounter{figure}{0}
\setcounter{theorem}{0}
\setcounter{equation}{0}
\setcounter{algorithm}{0}
\renewcommand{\thetable}{S.\Roman{table}}
\renewcommand{\thesection}{S.\Roman{section}}
\renewcommand{\thefigure}{S.\arabic{figure}}
\renewcommand\thetheorem{S.\arabic{theorem}}
\renewcommand{\theequation}{S.\arabic{equation}}
\renewcommand{\thealgorithm}{S.\arabic{algorithm}}

This supplement provides additional results to accompany our manuscript \cite{Zheng&etal:19L1ST}.
We use the prefix ``S'' for the numbers in sections, equations, figures, and tables in the supplementary material.

\section{Comparisons of a deep neural network trained with \dquotes{noisy} targets and \dquotes{clean} targets}
\label{sec:noisy_targets}

\begin{figure*}[!t]
	\centering  	
	\subfigure[The input FBP image ($123$ views)]{
		\includegraphics[width=0.48\textwidth]{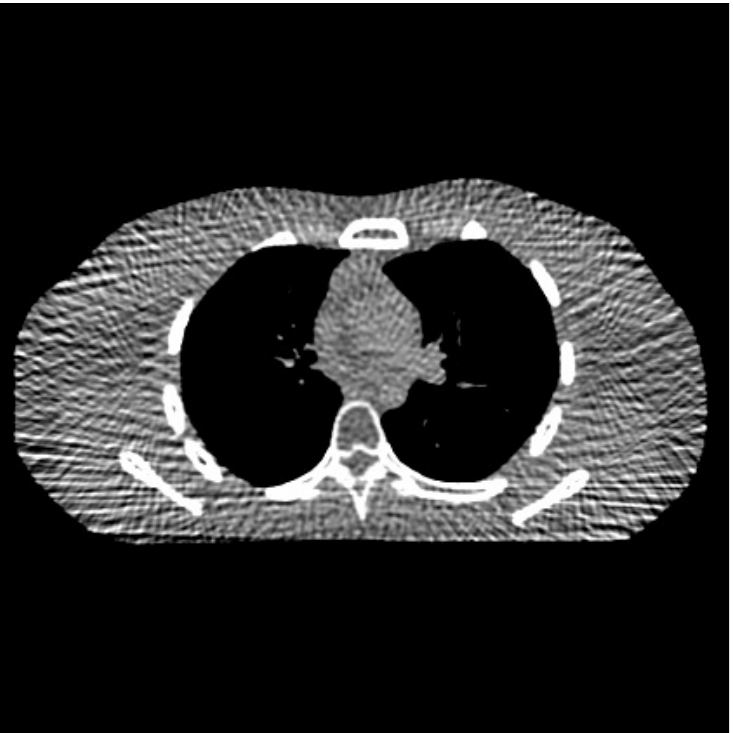}	
	} 
	\subfigure[Ground truth]{
		\includegraphics[width=0.48\textwidth]{./Fig/2D_true-eps-converted-to}	
	}  \\ 
	\subfigure[The FBPConvNet result image using noisy targets]{	\includegraphics[width=0.48\textwidth]{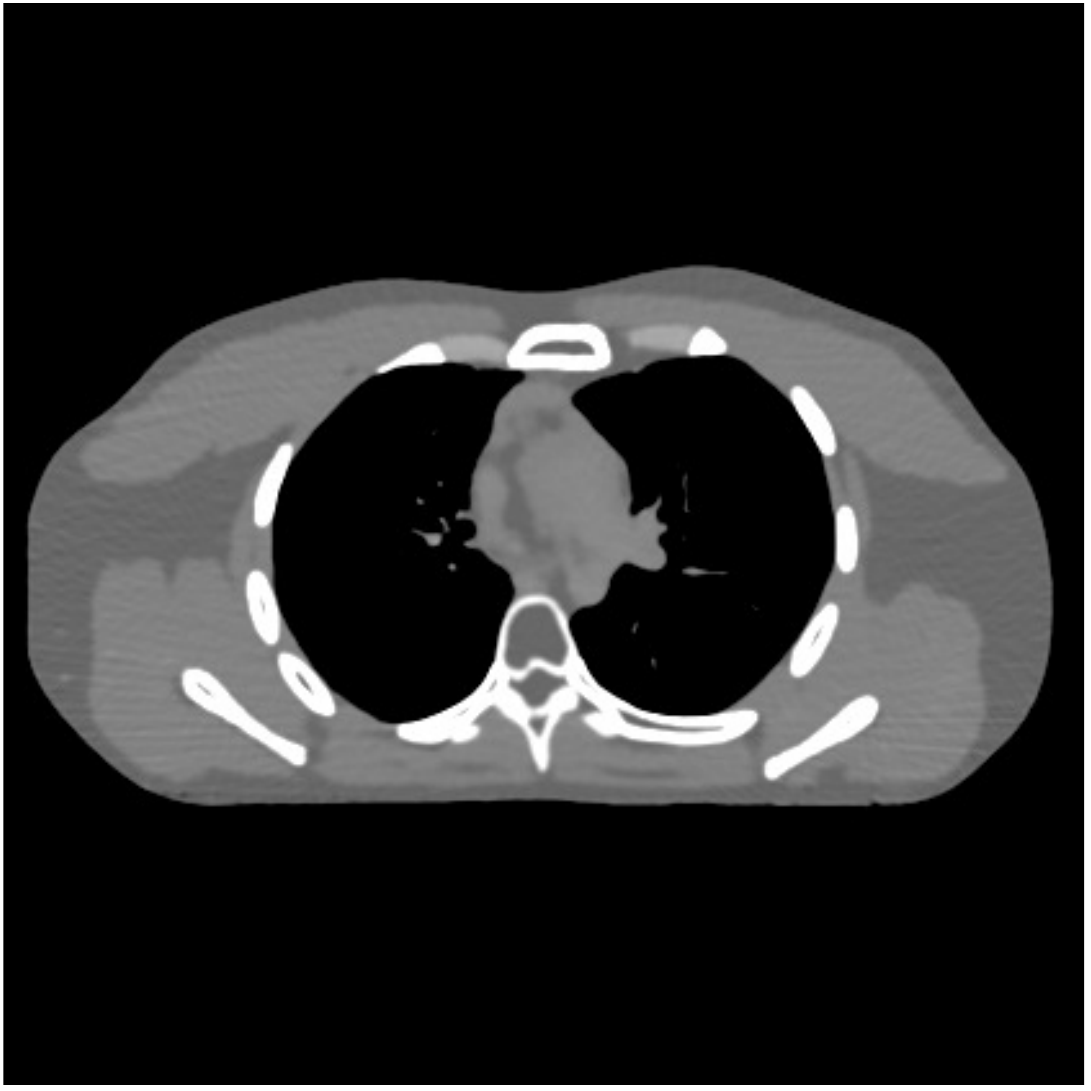}		
	}	
	\subfigure[The FBPConvNet result image using clean targets]{	\includegraphics[width=0.48\textwidth]{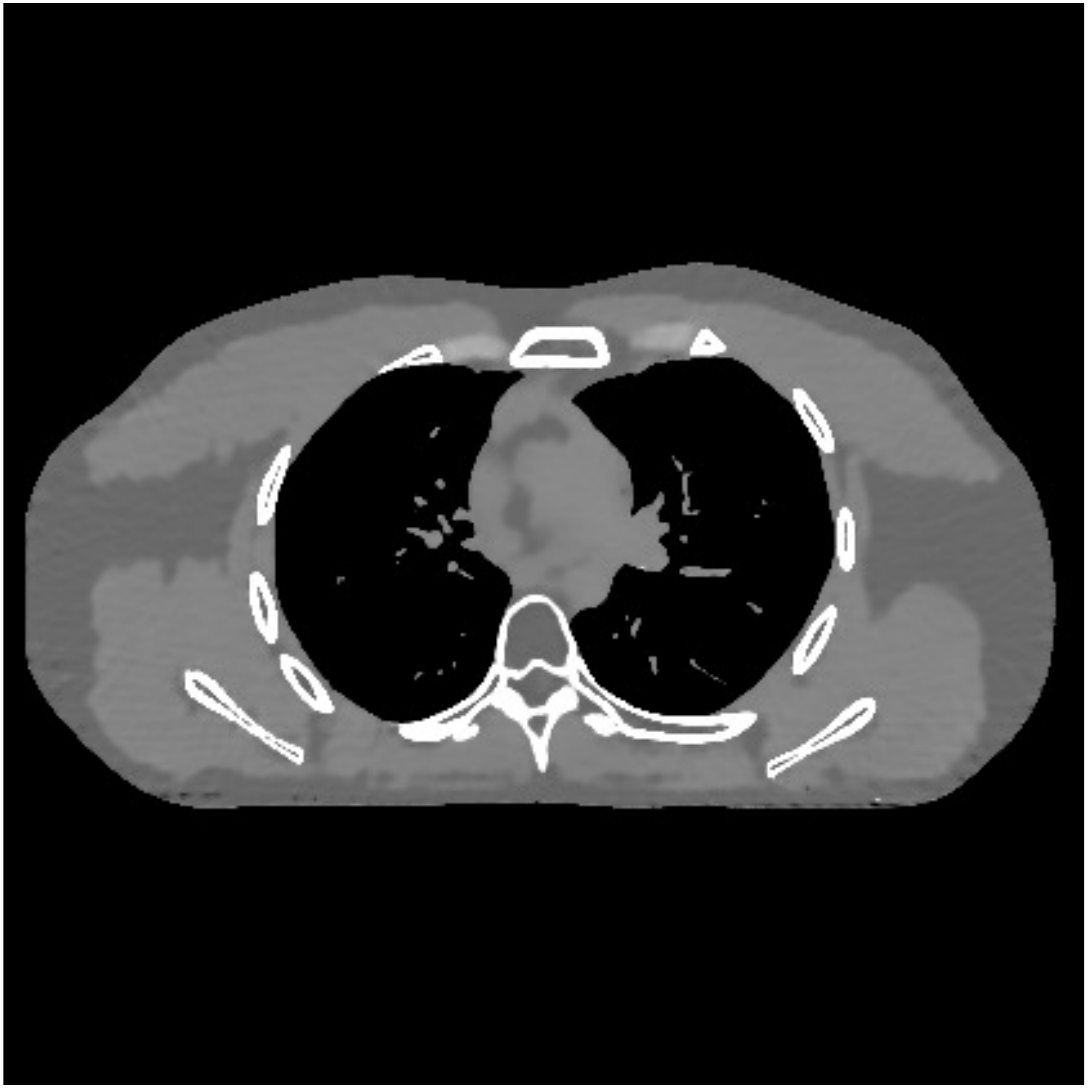}		
	}	
	\caption{Comparison of FBPConvNet result images using noisy targets and clean targets. (a) RMSE = $82.7$ HU;  (b) RMSE = $54.8$ HU; (c) RMSE = $23.9$ HU. Display window is $[800, 1200]$ HU.} 
	\label{fig:n2n}
\end{figure*}

	Based on the \dquotes{Noise2Noise} approach \cite{lehtinen:18:nli}, we trained a FBPConvNet \cite{Jin&etal:17TIP} network with \dquotes{noisy} targets by using pairs of FBP-reconstructed images from $123$-views and full ($984$)-views scans. We trained another FBPConvNet network with \dquotes{clean} targets by using pairs of FBP-reconstructed images from $123$-views scans and ground truth images. See training details in Section~\ref{subsubsec:2d_training}.
	Fig.~\ref{fig:n2n} shows image results of FBPConvNet using noisy targets and clean targets. The image using noisy targets is over-smoothed in bone regions and loses many structural details in lung regions, compared to the one using clean targets. The reason is twofold based on the limitations of the Noise2Noise approach. First, Noise2Noise assumes noise on noisy targets to be zero-mean.  However, it is unclear what distributions the artifacts or noise on noisy targets follow. Second, it is difficult to determine which loss function is optimal or reliable for training with noisy targets. This comparison suggests that for methods trained with supervised learning, one would expect improved results by using clean targets in the training processes.

\section{Additional results}

\begin{figure*}[!t]
	\centering
	\subfigure[Phantom data, $25$\% ($246$) views]{
		\includegraphics[width=0.23\textwidth]{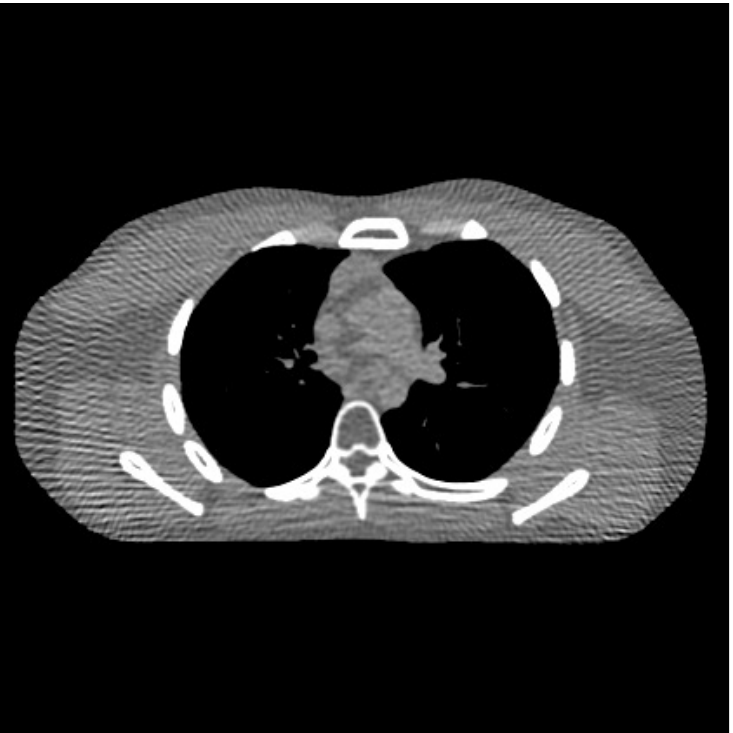}
	}
	\subfigure[Phantom data, $12.5$\% ($123$) views]{
		\includegraphics[width=0.23\textwidth]{./Fig/123_xfbp-eps-converted-to}
	} 
	\subfigure[Clinical data, $25$\% ($246$) views]{
		\includegraphics[width=0.23\textwidth]{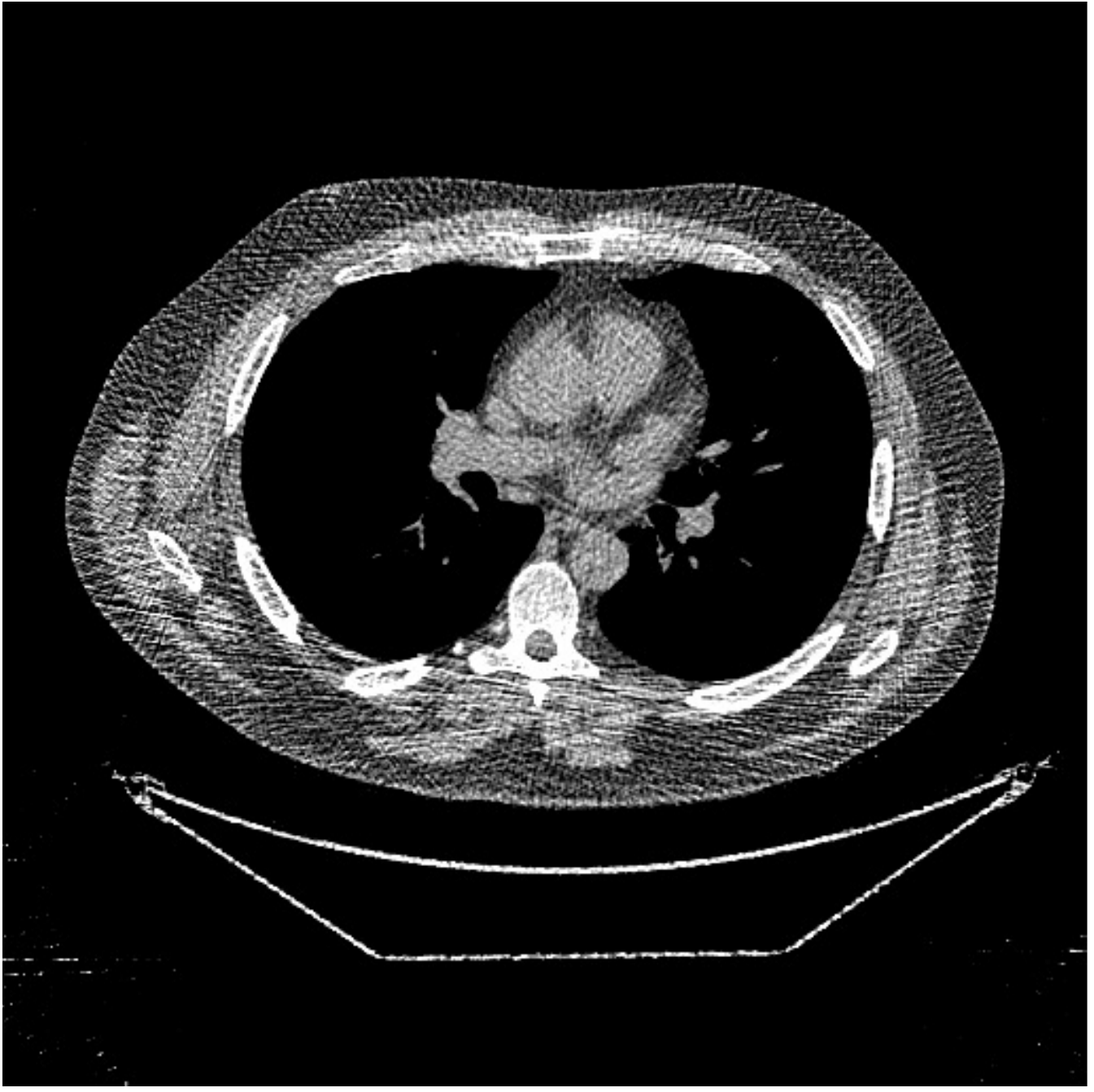}
	}
	\subfigure[Clinical data, $12.5$\% ($123$) views]{
		\includegraphics[width=0.23\textwidth]{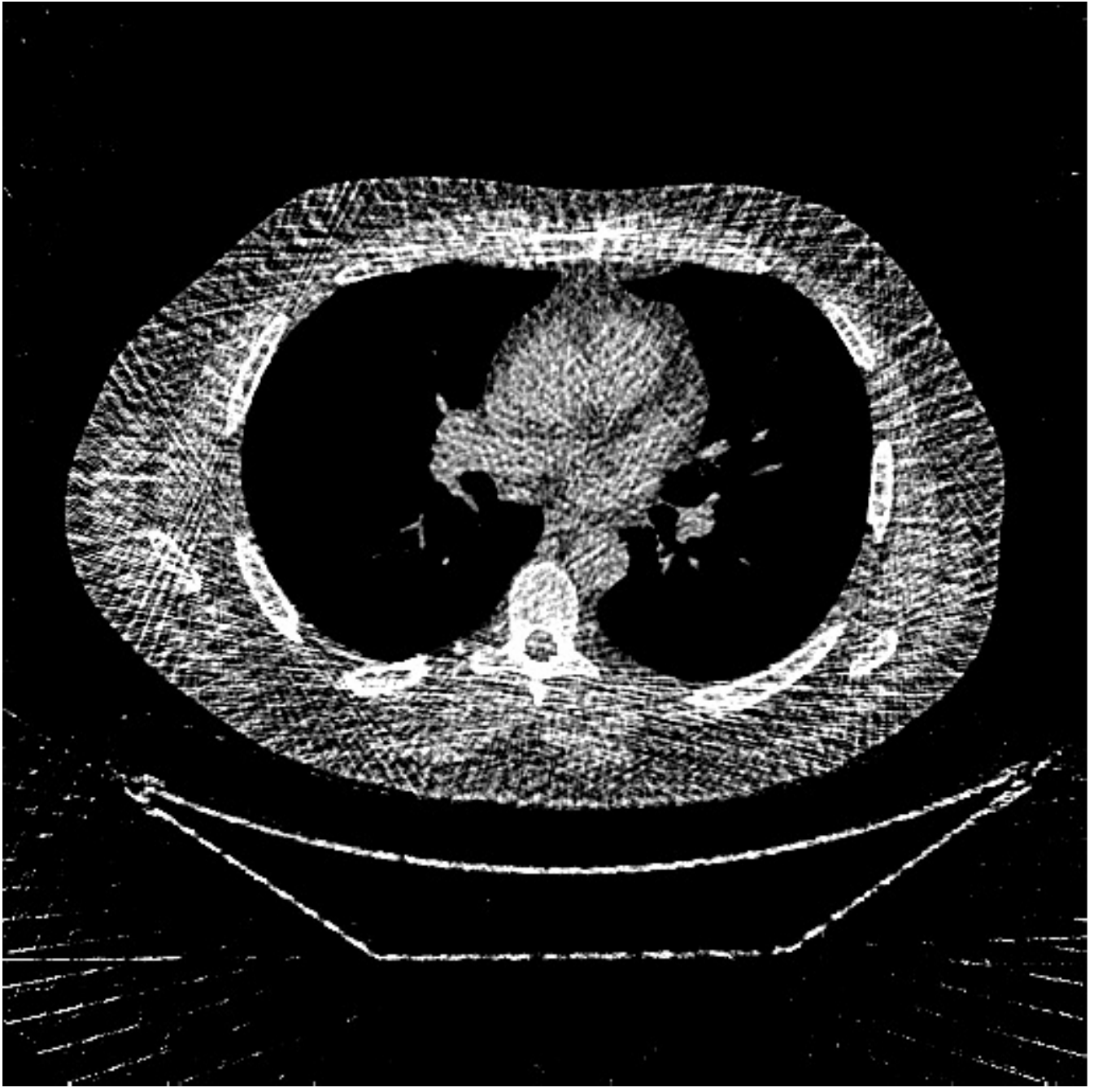}
	}
	\caption{The FBP reconstructions from the phantom data and the clinical data with different number of views. Display window is $[800, 1200]$ HU.} 
	\label{fig:2D_fbp}
	\vspace{0.1in}
\end{figure*}

Fig.~\ref{fig:2D_fbp} shows the FBP reconstructions from the phantom data and the clinical data with $25$\% ($246$) projection views and $12.5$\% ($123$) projection views.

\begin{figure*}[!h]
	\centering  	
	\includegraphics[width=0.35\textwidth]{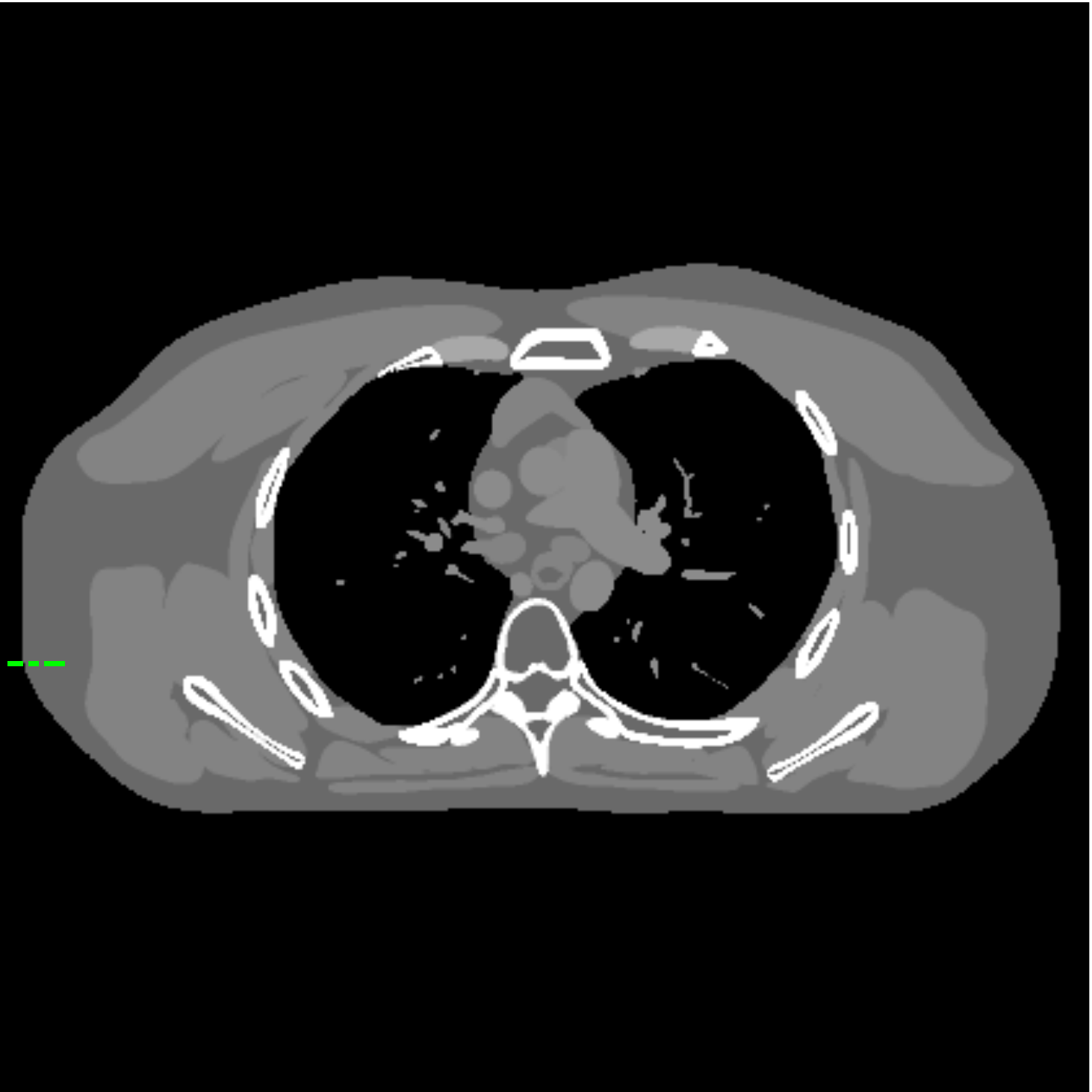} 
	\includegraphics[width=0.43\textwidth]{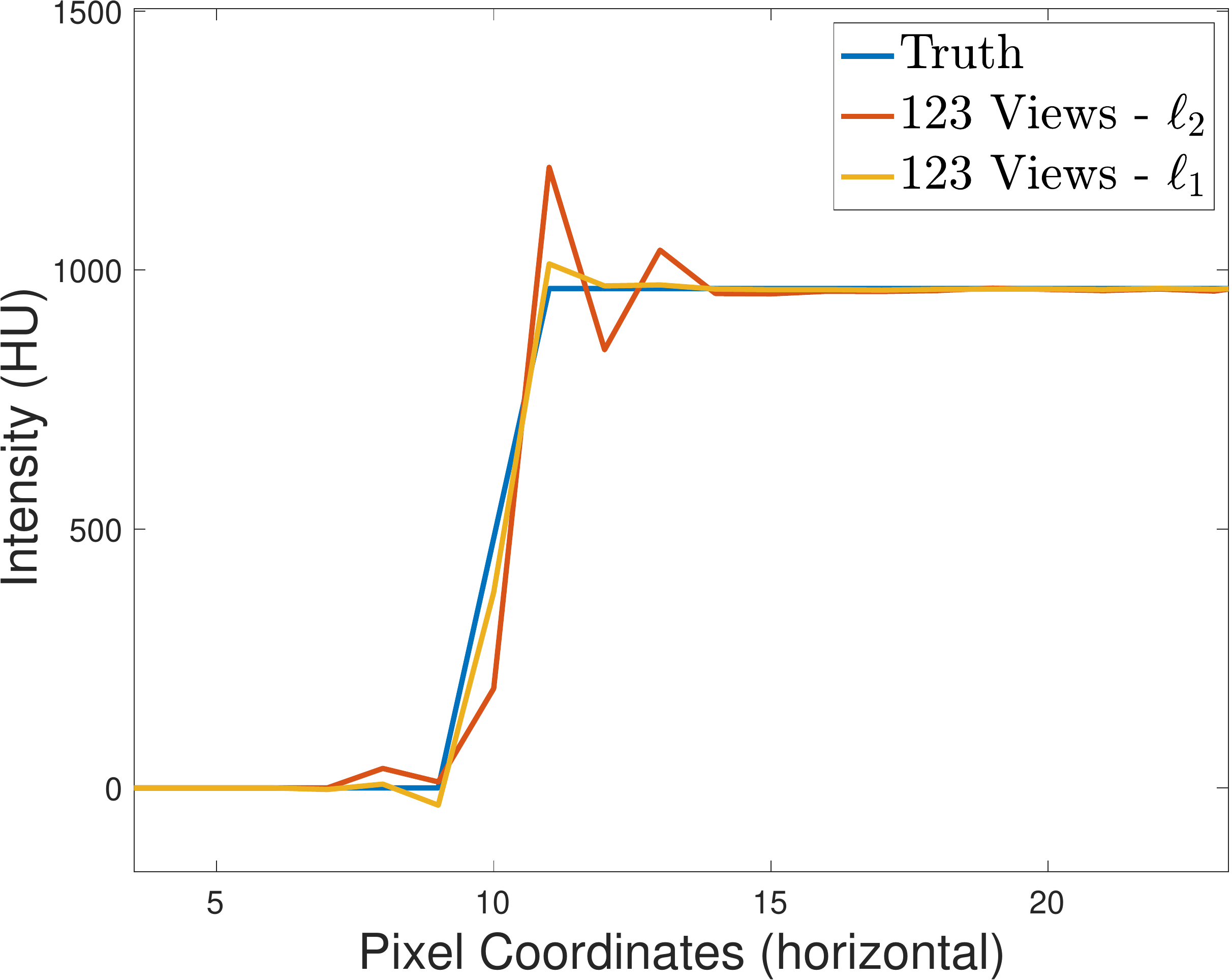}	
	\caption{Profile of XCAT reconstructions for the PWLS-ST-$\ell_2$ and PWLS-ST-$\ell_1$ methods (2D fan-beam and 123 views). The profile location is indicated by a green line.} 
	\label{fig:profile}
\end{figure*}

Fig.~\ref{fig:profile} shows an example of the profiles of PWLS-ST-$\ell_1$ and PWLS-ST-$\ell_2$. PWLS-ST-$\ell_2$ suffers from Gibbs phenomenon due to the model mismatch, and has some ringing artifacts around the edges with high transition.

\begin{figure*}[!h]
	\centering
	\includegraphics[width=0.4\textwidth]{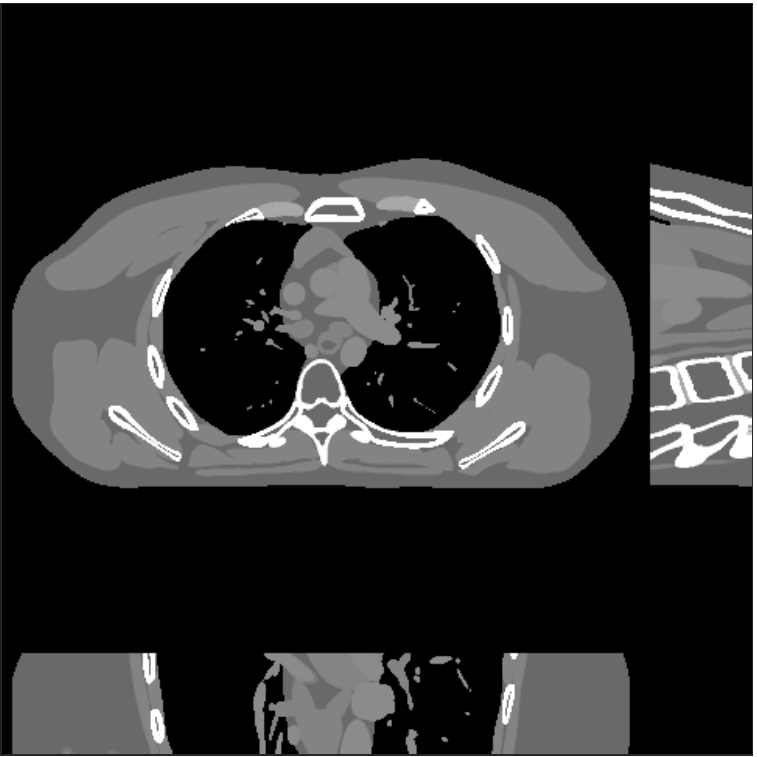}
	\caption{The XCAT phantom in the ROI of size $420 \times 420 \times 64$ used for testing in our 3D experiments. Display window is $[800, 1200]$ HU.} 
	\label{fig:3Dtruth}
\end{figure*}

Fig.~\ref{fig:3Dtruth} shows the XCAT phantom in the ROI of size $420 \times 420 \times 64$ used for testing in our 3D experiments.

\begin{figure*}[!t]
	\centering  	
	\subfigure[PWLS-EP initialization]{
		\includegraphics[width=0.315\textwidth]{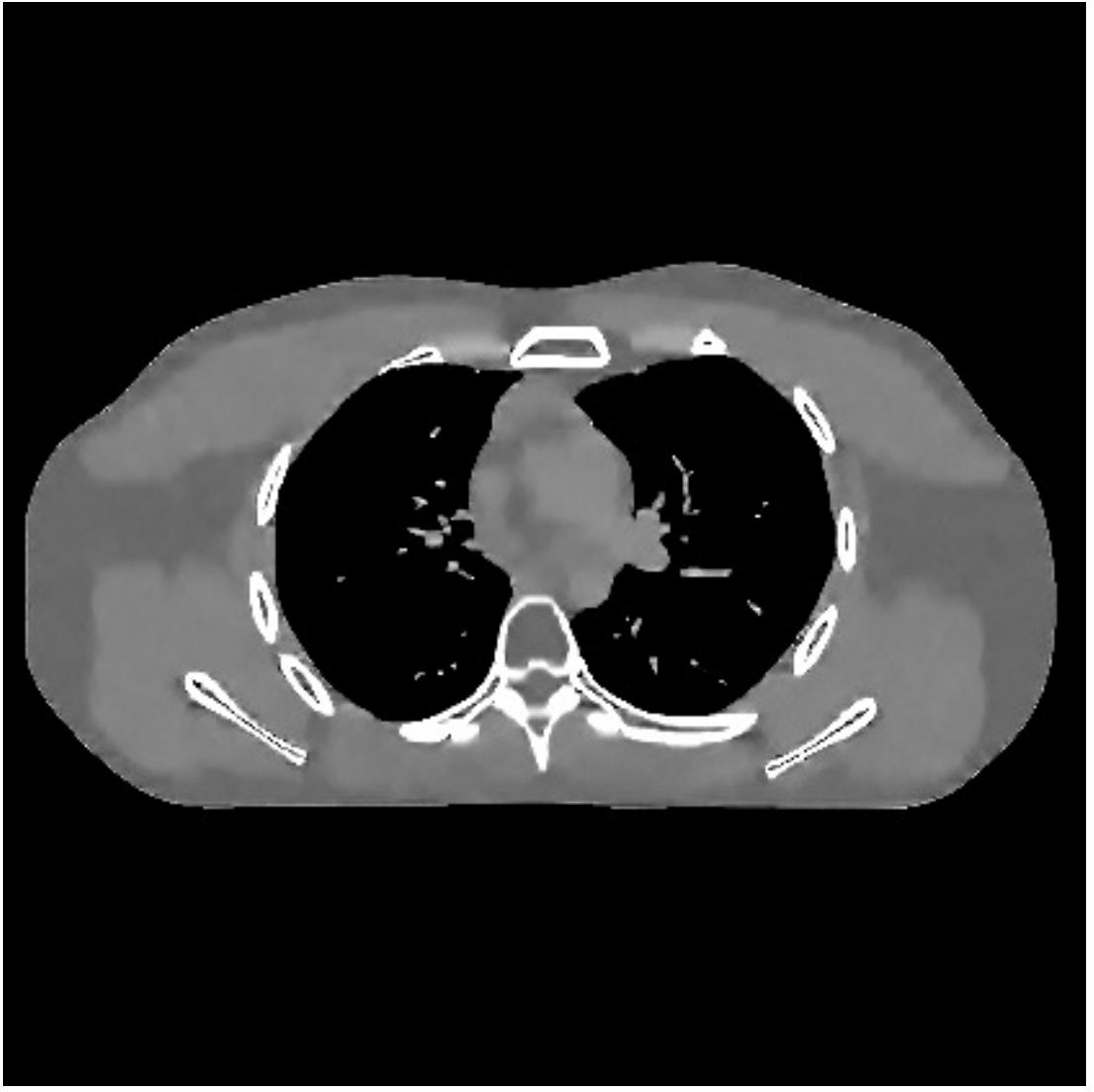}	
	} 
	\subfigure[FBP initialization]{
		\includegraphics[width=0.315\textwidth]{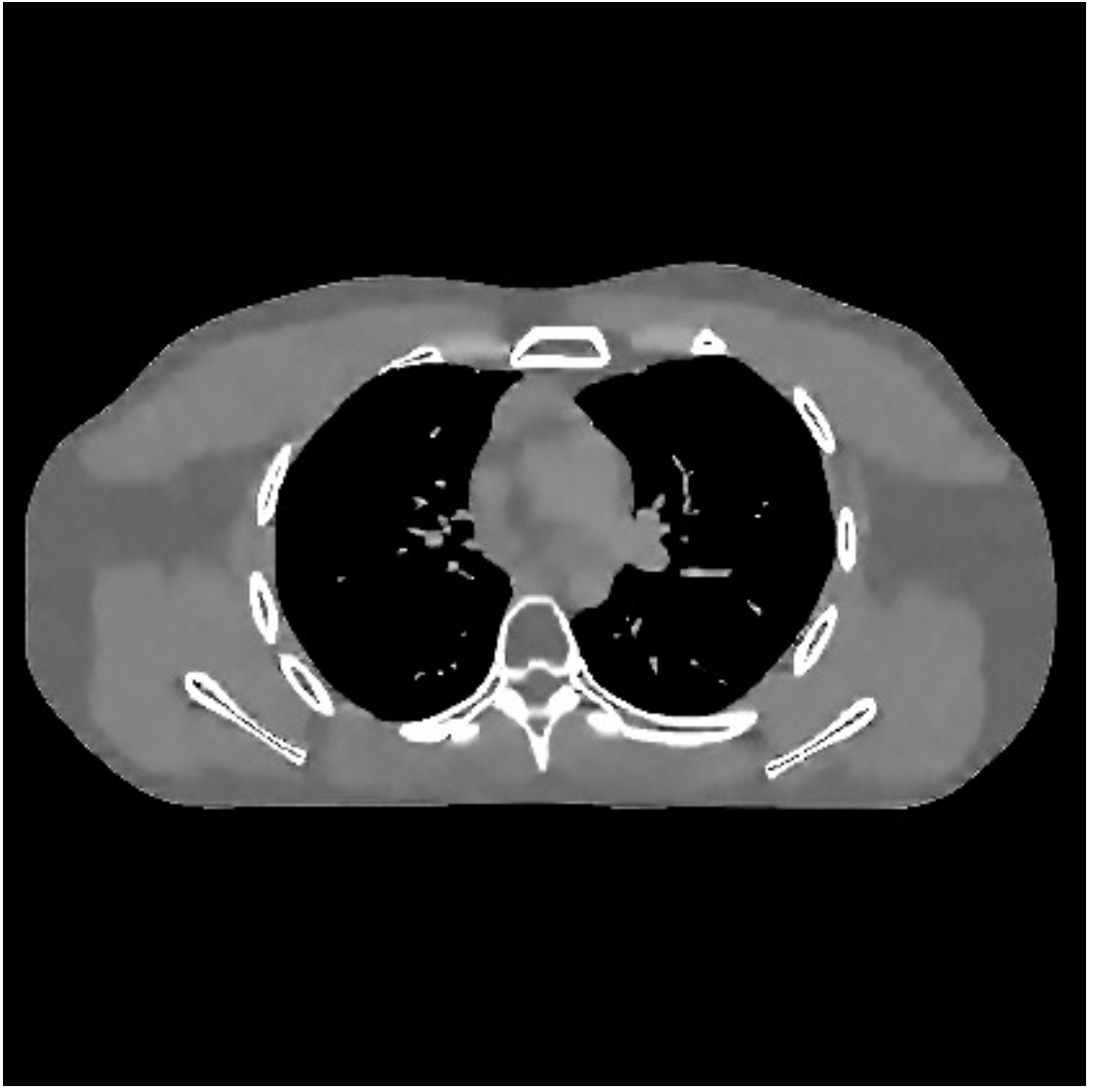}	
	}  
	\subfigure[Initialized with a constant image]{	
		\includegraphics[width=0.315\textwidth]{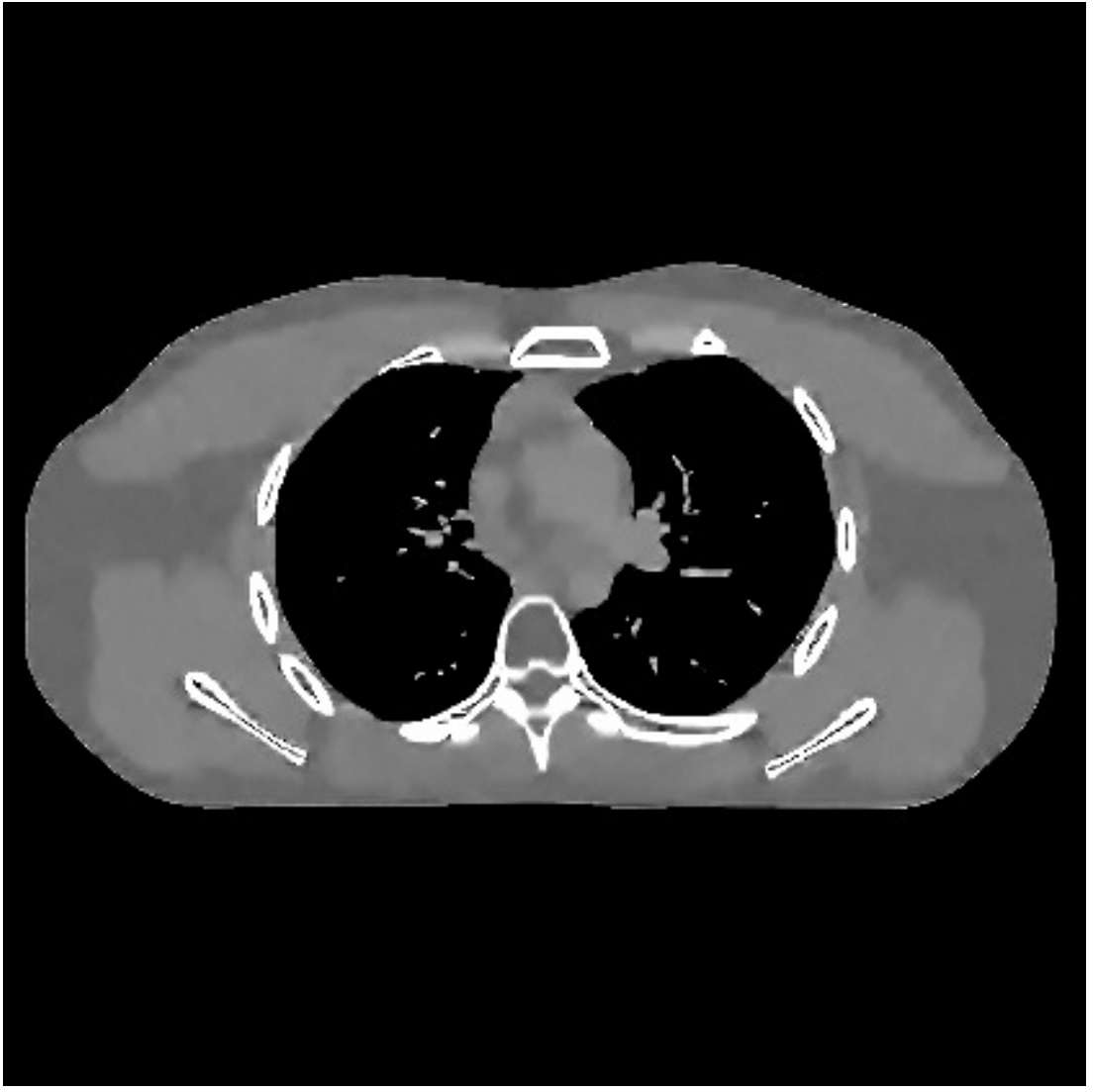}		
	}	\\
	\subfigure[PWLS-EP initialization]{
		\includegraphics[width=0.315\textwidth]{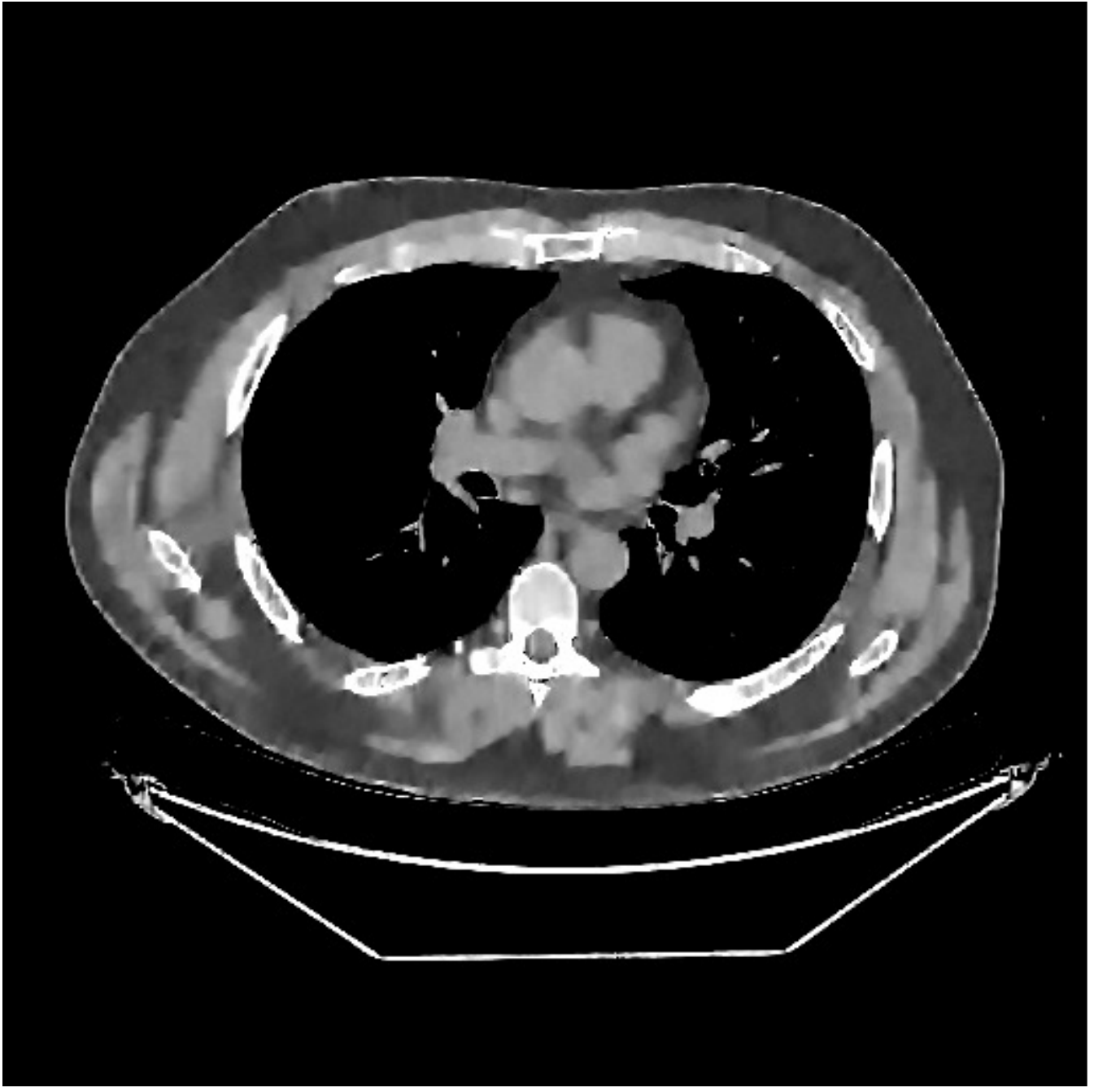}	
	} 
	\subfigure[FBP initialization]{
		\includegraphics[width=0.315\textwidth]{./2D_GE_Data/GEdata_FBPInit_123L1_lam8e-03_zeta80_30kap1_30kap2_iter2_1000outer_learn110.pdf}	
	}  
	\subfigure[Initialized with a constant image]{	
		\includegraphics[width=0.315\textwidth]{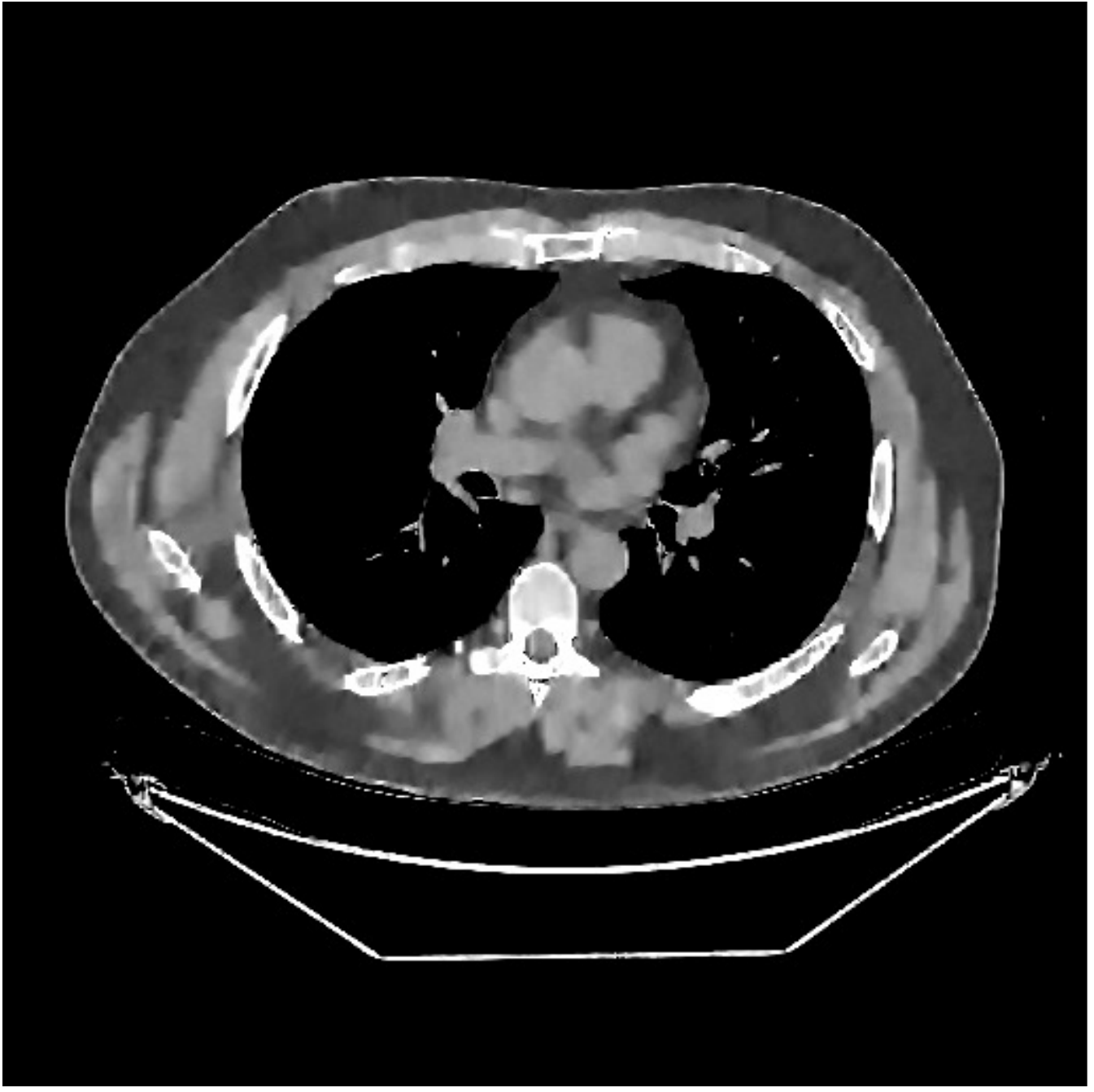}		
	}
	
	\caption{Comparison of 2D reconstructed images from phantom data (first row) and clinical data (second row) for the proposed PWLS-ST-$\ell_1$ method with different initializations. (2D fan-beam CT geometry with $123$ views; display window is $[800, 1200]$ HU). (a) RMSE = $25.8$ HU;  (b) RMSE = $25.9$ HU; (c) RMSE = $26.3$ HU. }
	\label{fig:recon_diff_init}
\end{figure*}

Fig.~\ref{fig:recon_diff_init} shows that the proposed PWLS-ST-$\ell_1$ method provides very similar reconstructed images with three different initialization images for both phantom data and clinical data, indicating that PWLS-ST-$\ell_1$ is robust to different initializations. (We used the same parameters for the three cases and ran a sufficient large number of iterations (i.e., $5000$ iterations) for the case initialized with an image of all ones.)	
Using a better initialization (e.g., the reconstructed image with PWLS-EP), the proposed PWLS-ST-$\ell_1$ method converges faster.

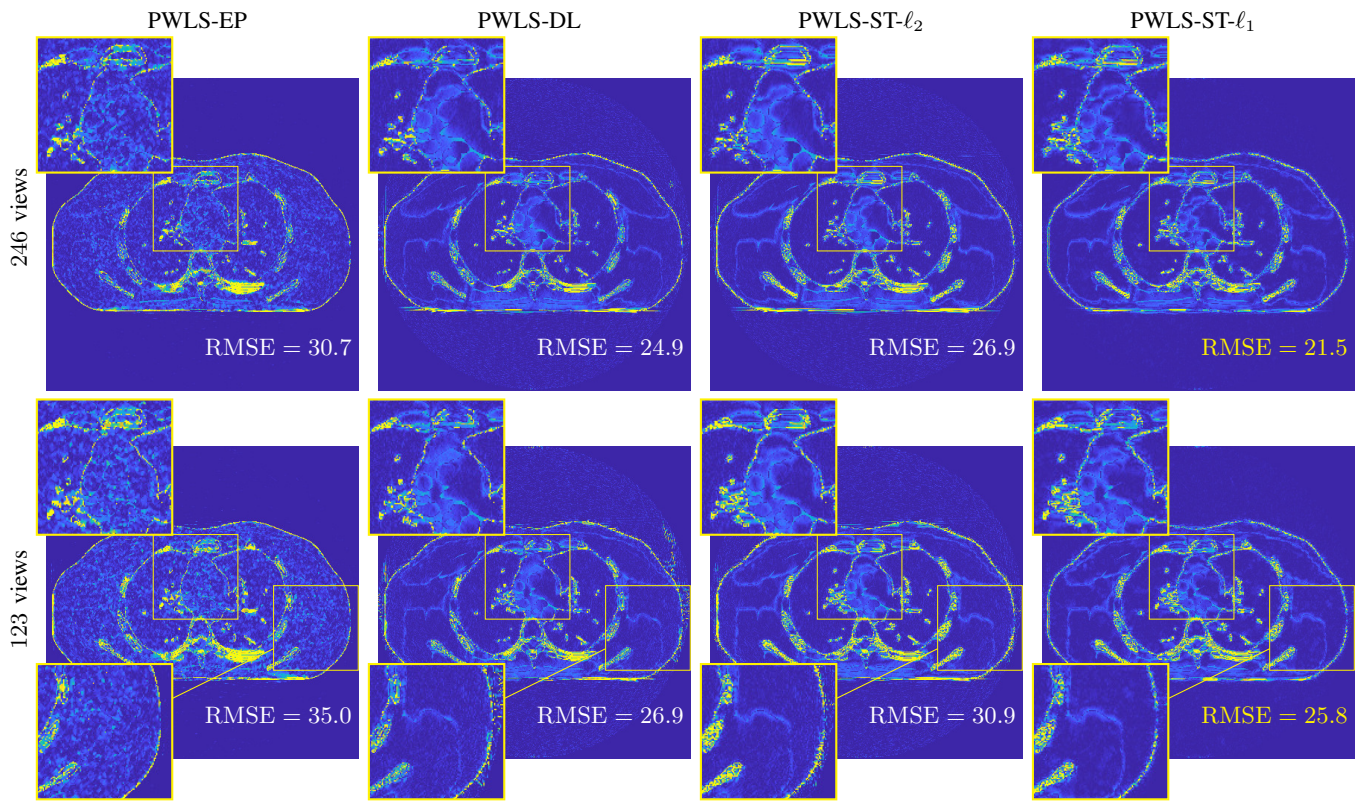
\begin{figure*}[!t]
	\centering
	\small\addtolength{\tabcolsep}{-7.5pt}

	\begin{tabular}{ccccc}
		
		{} & \small{PWLS-EP} & \small{PWLS-DL} & \small{PWLS-ST-$\ell_2$} & \small{PWLS-ST-$\ell_1$} \\
		
		\raisebox{-.5\height}{\begin{turn}{+90} \small{$246$ views} \end{turn}}~ &
		\raisebox{-.5\height}{
			\begin{tikzpicture}
			\begin{scope}[spy using outlines={rectangle,yellow,magnification=1.6,size=18mm}]
			\node {\includegraphics[scale=0.28]{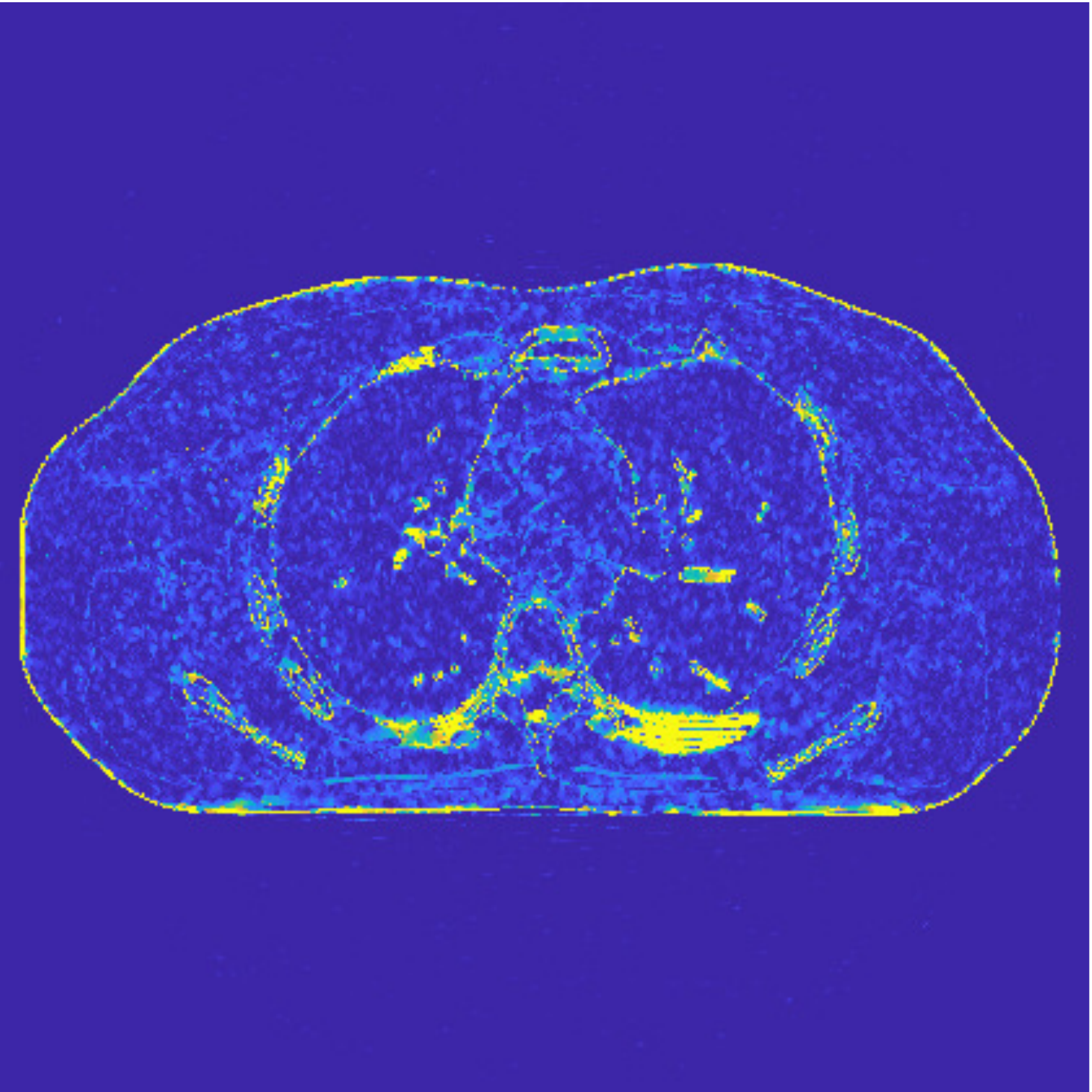}};
			\spy on (-0.1,0.34) in node [left] at (-0.41,1.72);
			\node [white] at (1,-1.5) {\small $\mathrm{RMSE} = 30.7$};
			\end{scope}
			\end{tikzpicture}} &
		\raisebox{-.5\height}{
			\begin{tikzpicture}
			\begin{scope}[spy using outlines={rectangle,yellow,magnification=1.6,size=18mm}]
			\node {\includegraphics[scale=0.28]{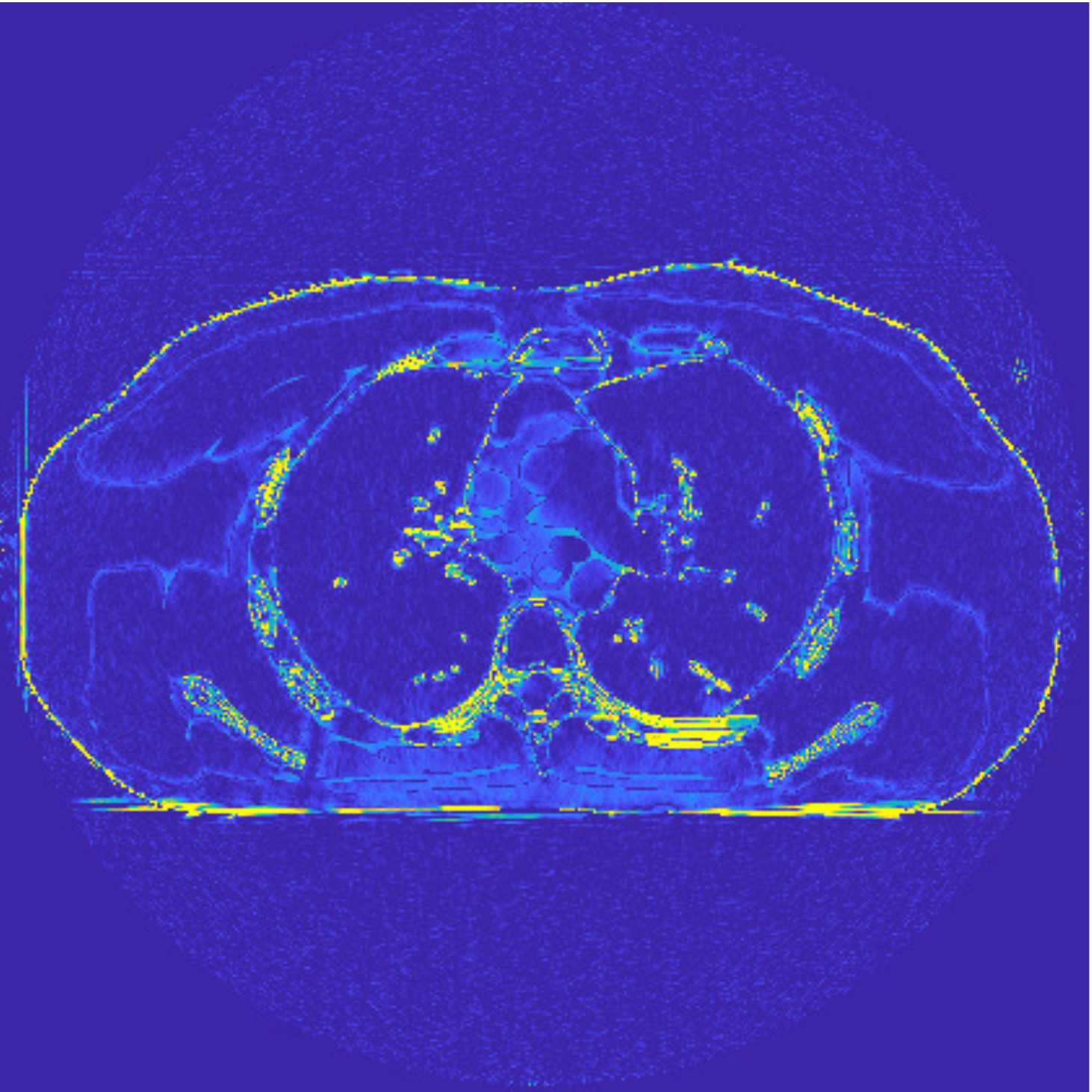}};
			\spy on (-0.1,0.34) in node [left] at (-0.41,1.72);
			\node [white] at (1,-1.5) {\small $\mathrm{RMSE} = 24.9$};
			\end{scope}
			\end{tikzpicture}} &
		\raisebox{-.5\height}{
			\begin{tikzpicture}
			\begin{scope}[spy using outlines={rectangle,yellow,magnification=1.6,size=18mm}]
			\node {\includegraphics[scale=0.28]{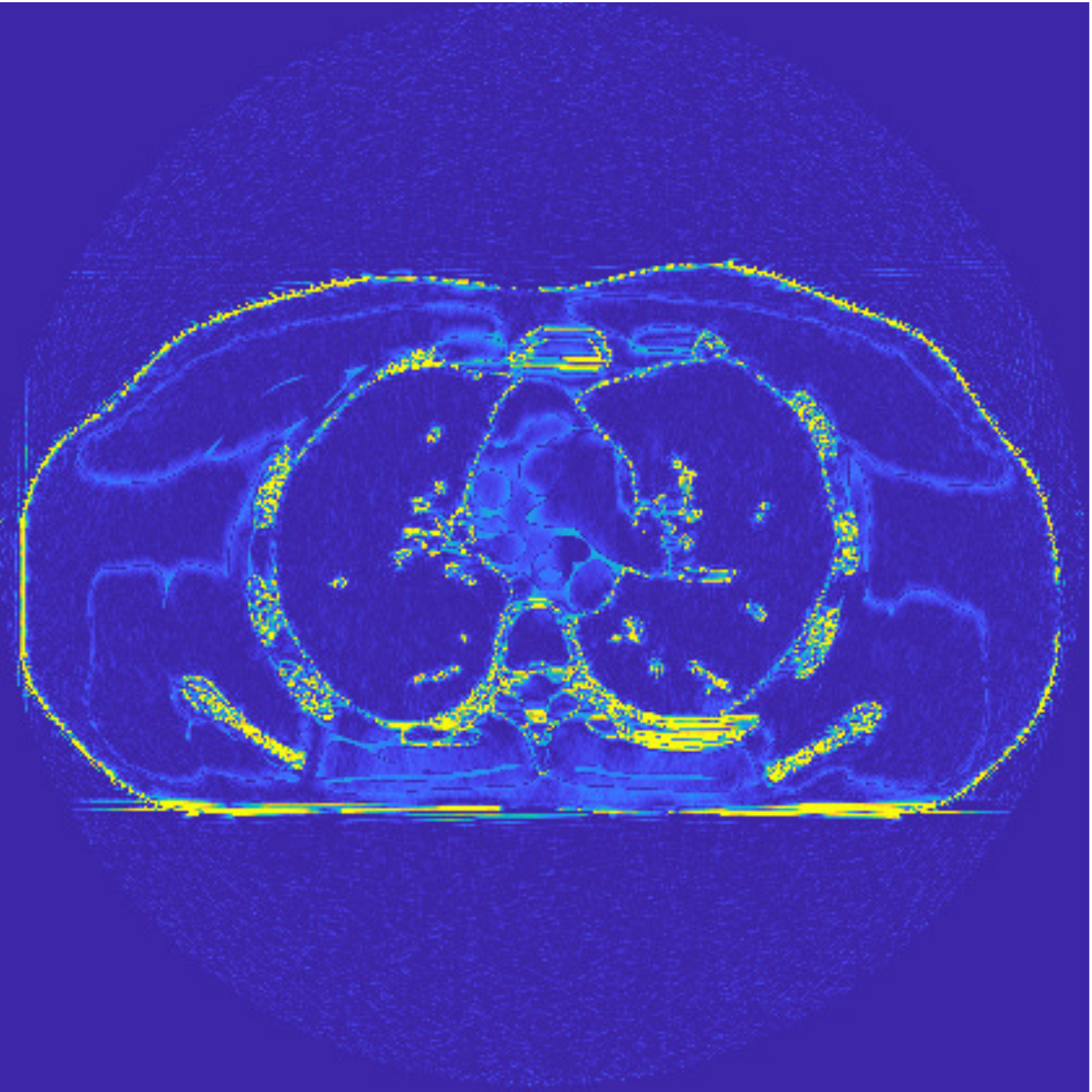}};
			\spy on (-0.1,0.34) in node [left] at (-0.41,1.72);
			\node [white] at (1,-1.5) {\small $\mathrm{RMSE} = 26.9$};
			\end{scope}
			\end{tikzpicture}} &
		\raisebox{-.5\height}{
			\begin{tikzpicture}
			\begin{scope}[spy using outlines={rectangle,yellow,magnification=1.6,size=18mm}]
			\node {\includegraphics[scale=0.28]{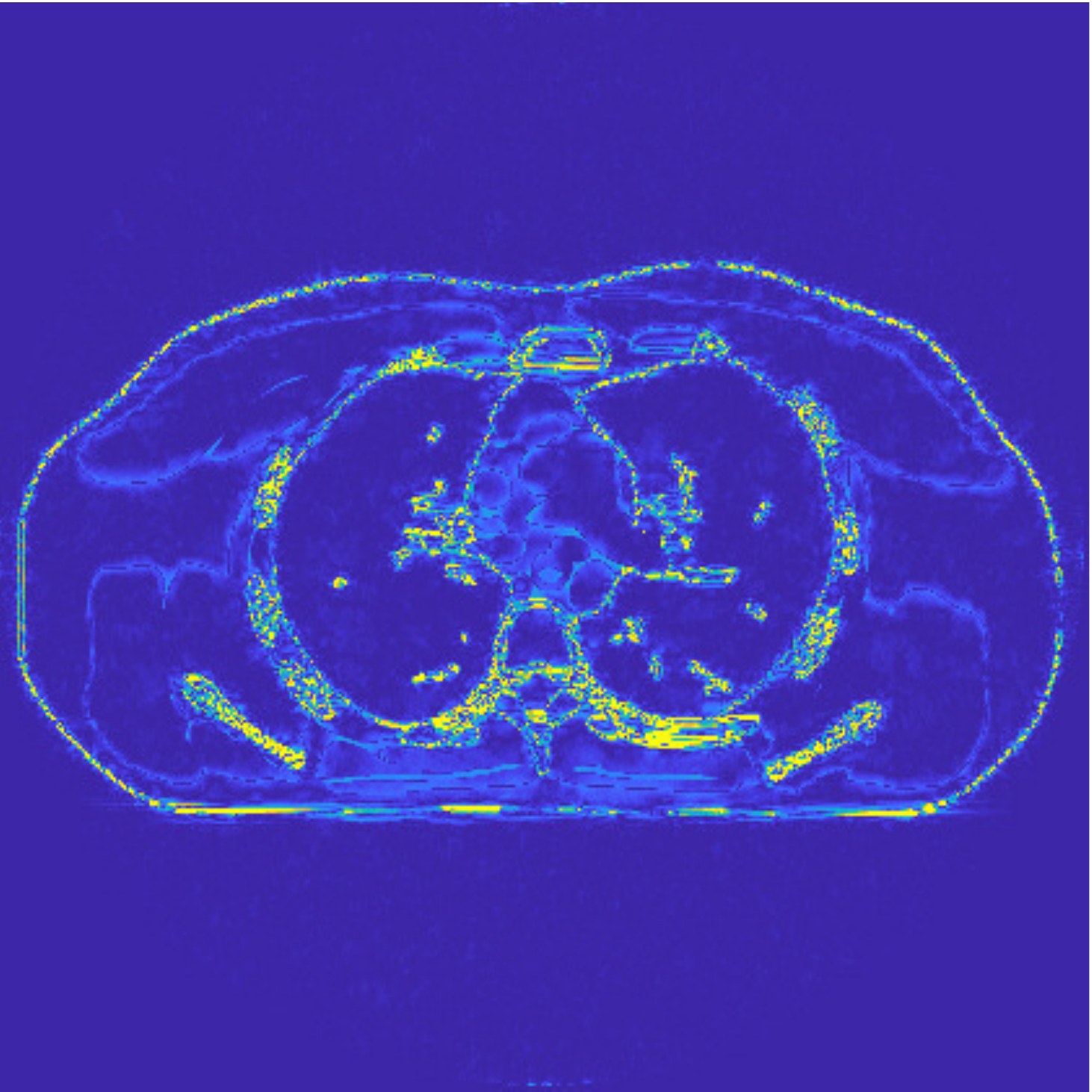}};
			\spy on (-0.1,0.34) in node [left] at (-0.41,1.72);
			\node [white] at (1,-1.5) {\small \color{yellow}{$\mathrm{RMSE} = 21.5$}};
			\end{scope}
			\end{tikzpicture}} \\

		\raisebox{-.5\height}{\begin{turn}{+90} \small{$123$ views} \end{turn}}~ &
		\raisebox{-.5\height}{
			\begin{tikzpicture}
			\begin{scope}[spy using outlines={rectangle,yellow,magnification=1.6,size=18mm,connect spies}]
			\node {\includegraphics[scale=0.28]{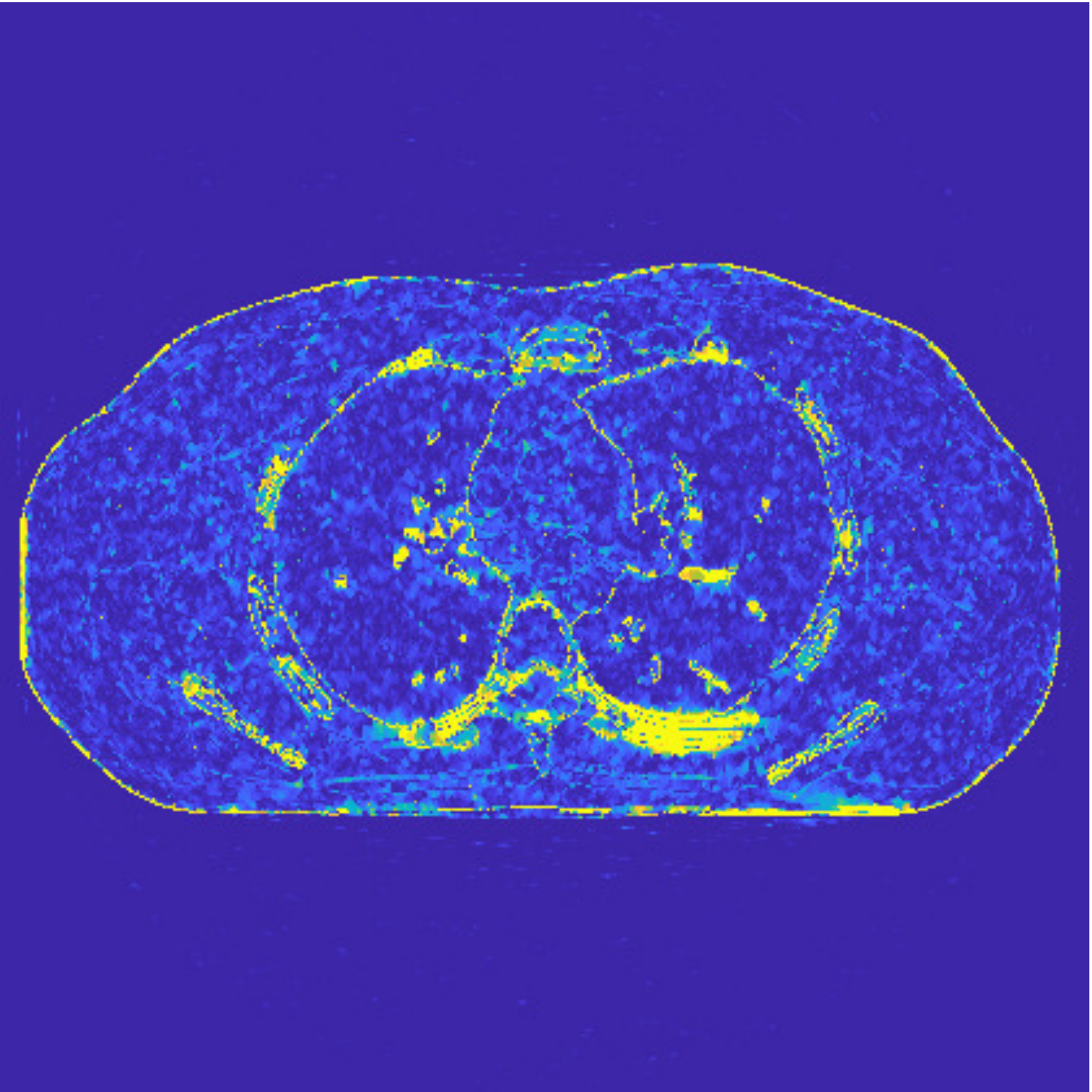}};
			\spy on (-0.1,0.34) in node [left] at (-0.41,1.8);
			\spy on (1.5,-0.34) in node [left] at (-0.41,-1.72);
			\node [white] at (1,-1.5) {\small $\mathrm{RMSE} = 35.0$};
			\end{scope}
			\end{tikzpicture}} &
		\raisebox{-.5\height}{
			\begin{tikzpicture}
			\begin{scope}[spy using outlines={rectangle,yellow,magnification=1.6,size=18mm,connect spies}]
			\node {\includegraphics[scale=0.28]{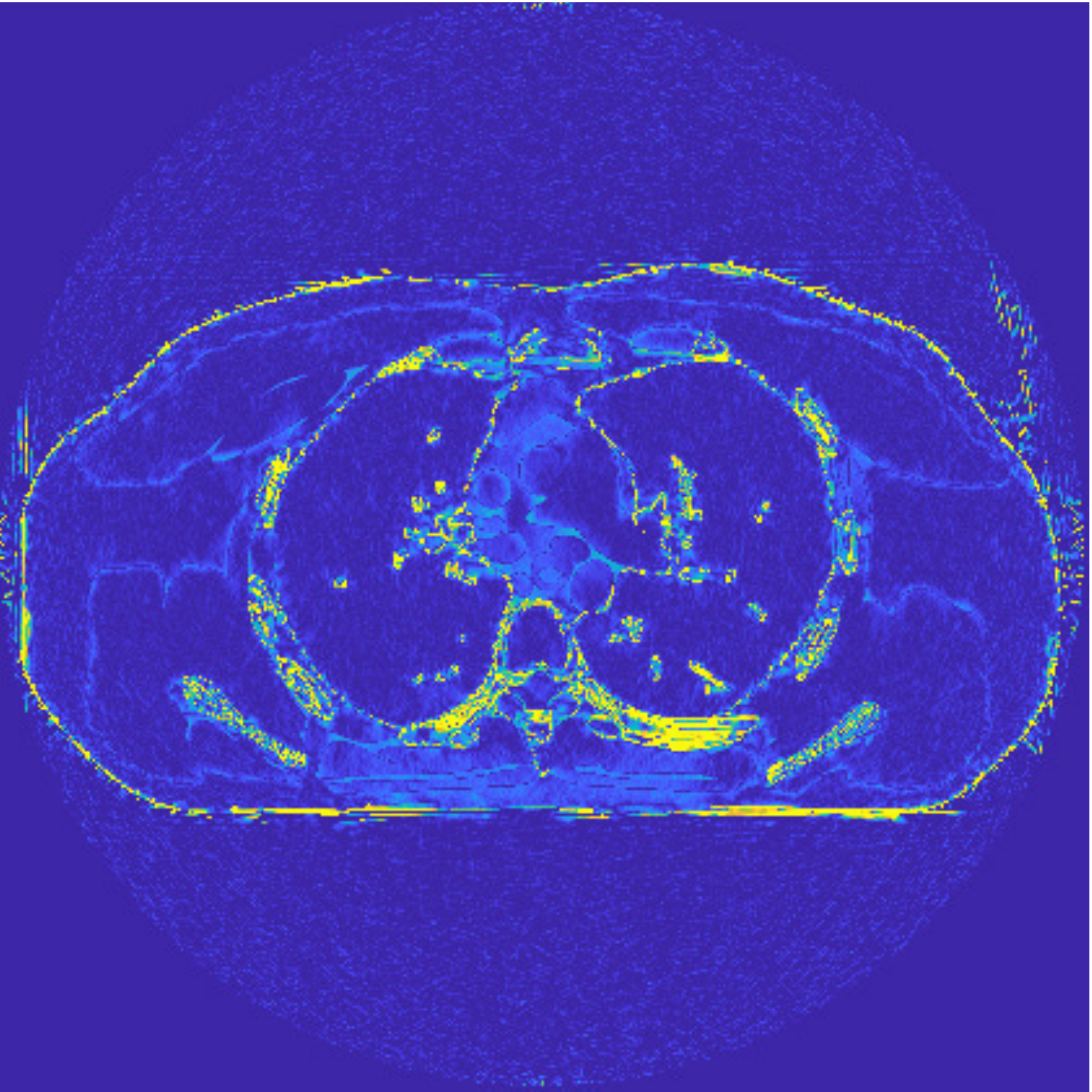}};
			\spy on (-0.1,0.34) in node [left] at (-0.41,1.8);
						\spy on (1.5,-0.34) in node [left] at (-0.41,-1.72);
			\node [white] at (1,-1.5) {\small $\mathrm{RMSE} = 26.9$};
			\end{scope}
			\end{tikzpicture}} &
		\raisebox{-.5\height}{
			\begin{tikzpicture}
			\begin{scope}[spy using outlines={rectangle,yellow,magnification=1.6,size=18mm,connect spies}]
			\node {\includegraphics[scale=0.28]{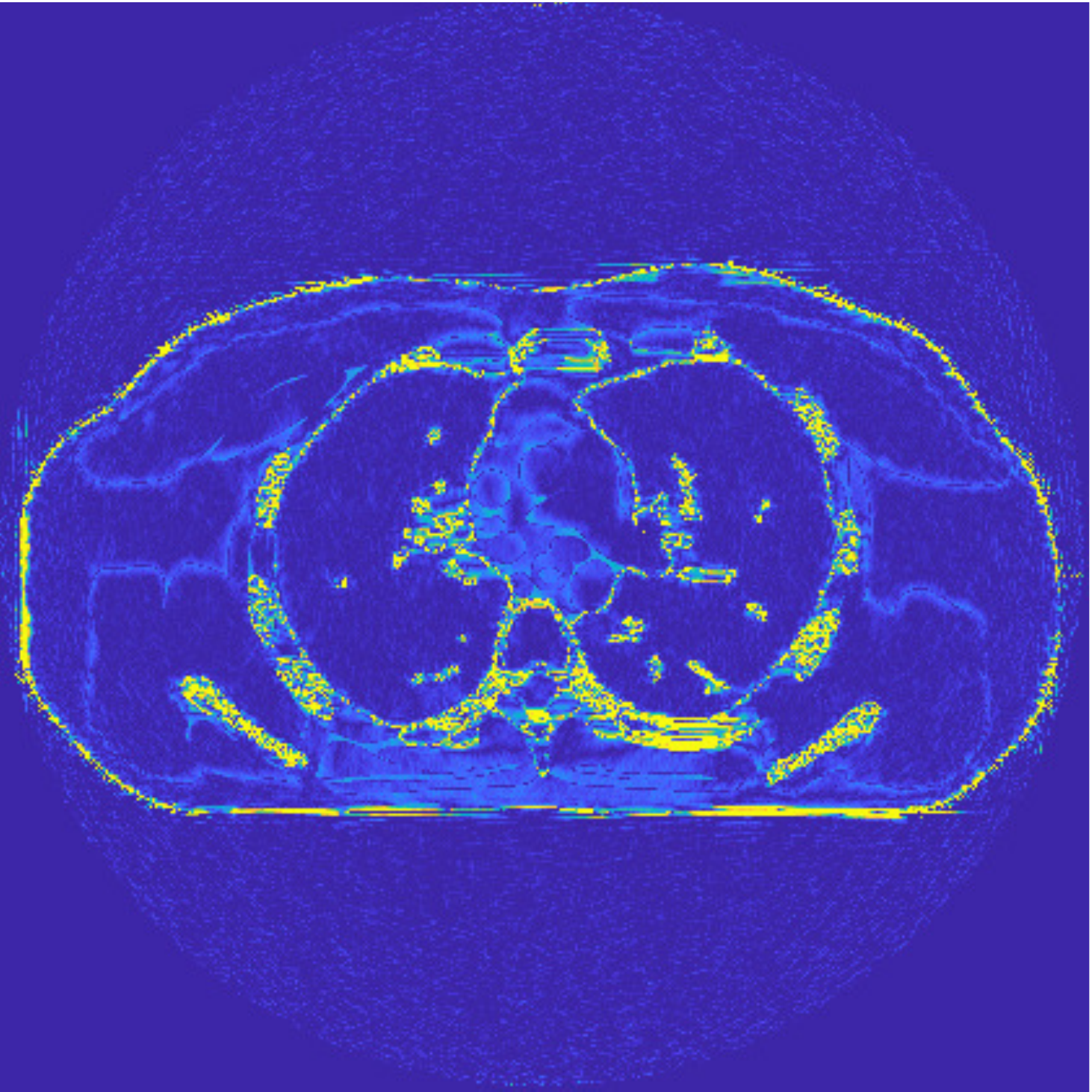}};
			\spy on (-0.1,0.34) in node [left] at (-0.41,1.8);
						\spy on (1.5,-0.34) in node [left] at (-0.41,-1.72);
			\node [white] at (1,-1.5) {\small $\mathrm{RMSE} = 30.9$};
			\end{scope}
			\end{tikzpicture}} &
		\raisebox{-.5\height}{
			\begin{tikzpicture}
			\begin{scope}[spy using outlines={rectangle,yellow,magnification=1.6,size=18mm,connect spies}]
			\node {\includegraphics[scale=0.28]{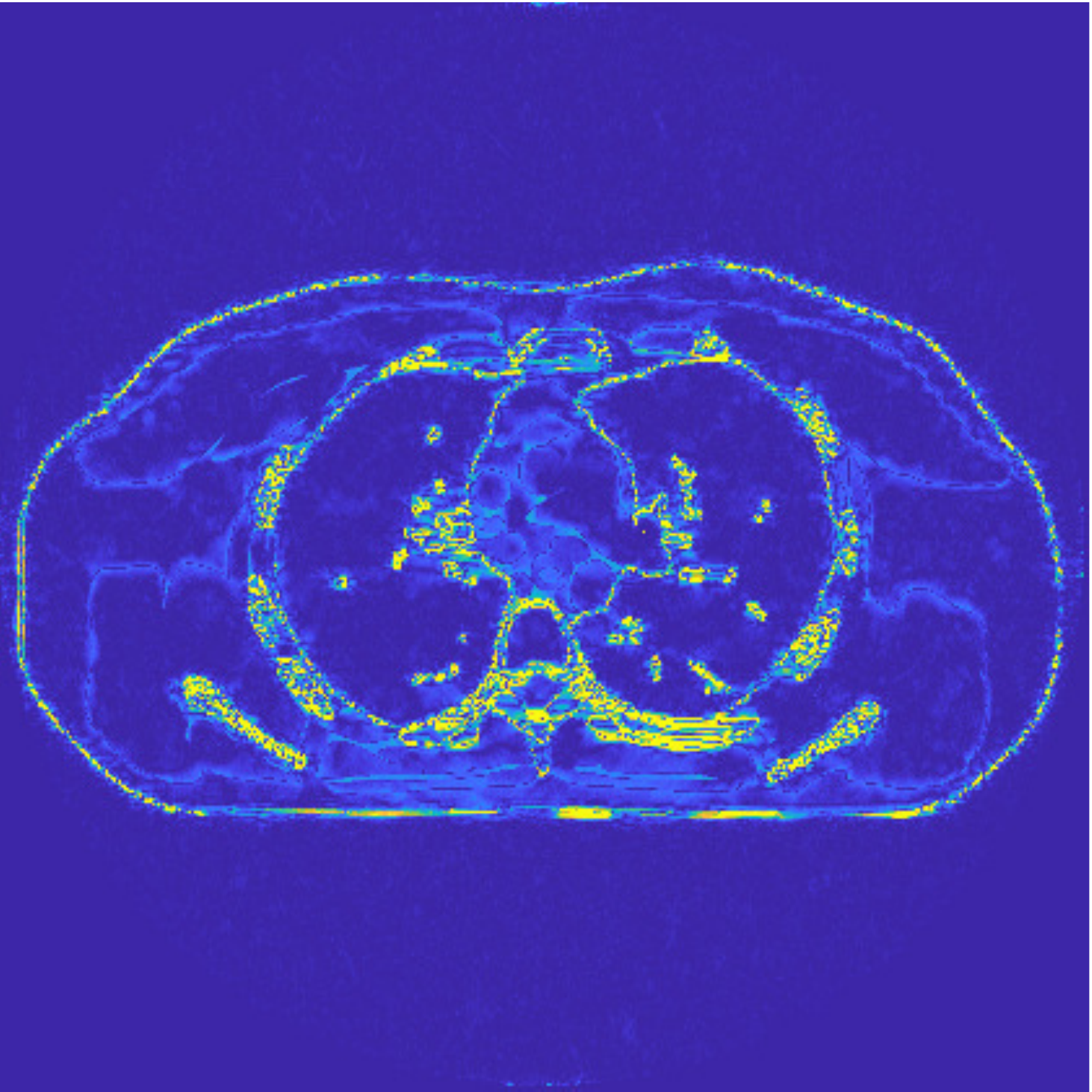}};
			\spy on (-0.1,0.34) in node [left] at (-0.41,1.8);
						\spy on (1.5,-0.34) in node [left] at (-0.41,-1.72);
			\node [white] at (1,-1.5) {\small \color{yellow}{$\mathrm{RMSE} = 25.8$}};
			\end{scope}
			\end{tikzpicture}} 
		
	\end{tabular}
	
	\caption{Corresponding error images of 2D reconstructed images from different X-ray CT reconstruction models with different number of views (2D fan-beam CT geometry and $\rho_0 = 10^5$). Display window is $[0, 100]$ HU.}
	\label{fig:2DCTrecon_err}
\end{figure*}

Fig.~\ref{fig:2DCTrecon_err} shows the error images (corresponding to Fig.~\ref{fig:2DCTrecon}(a)) of 2D reconstructions with the PWLS-EP, PWLS-DL, \mbox{PWLS-ST-$\ell_2$}, and \mbox{PWLS-ST-$\ell_1$} methods. The proposed PWLS-ST-$\ell_1$ approach consistently provides more accurate reconstructions compared to the other methods. Specifically, PWLS-ST-$\ell_1$ has smaller errors in the heart region (see zoom-ins) of 2D reconstructions than PWLS-ST-$\ell_2$ and PWLS-DL. In addition, PWLS-ST-$\ell_1$ does not have ringing artifacts around the edges with high transition. Compared to PWLS-ST-$\ell_1$, PWLS-ST-$\ell_2$ and PWLS-DL give more and stronger ringing artifacts in reconstruction for $123$ views (see zoom-ins).

\begin{figure*}[!b]
	\centering
	\small\addtolength{\tabcolsep}{-7.5pt}

	\begin{tabular}{ccccc}
		
		{} & \small{FBP} & \small{PWLS-EP} & \small{PWLS-ST-$\ell_2$} & \small{PWLS-ST-$\ell_1$} \\
		
		\raisebox{-.5\height}{\begin{turn}{+90} \small{$246$ views} \end{turn}}~ &
		\raisebox{-.5\height}{
			\begin{tikzpicture}
			\begin{scope}[spy using outlines={rectangle,yellow,magnification=1.55,size=18mm, connect spies}]
			\node {\includegraphics[scale=0.35]{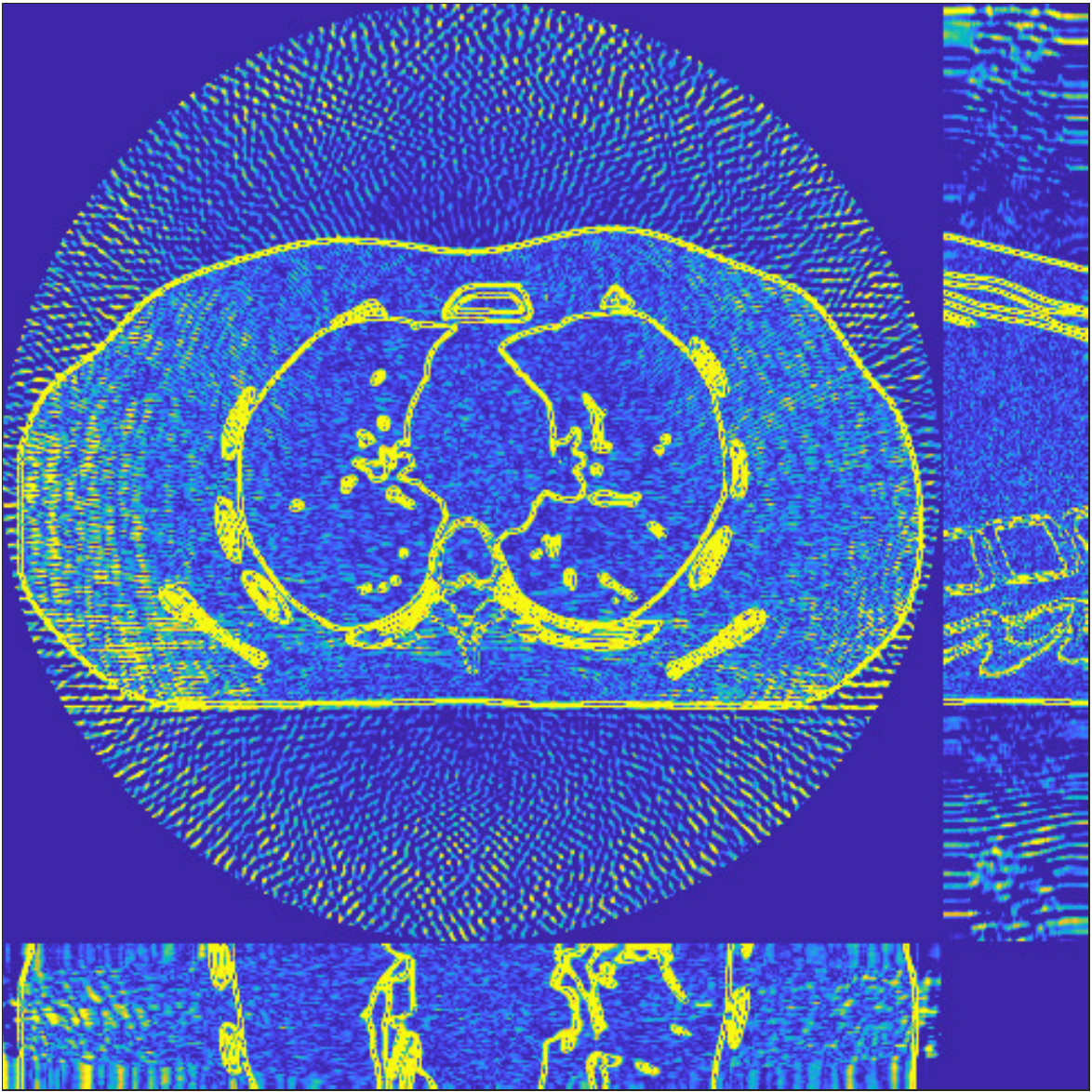}};
\spy on (0.9, 0.1) in node [left] at (-0.41,1.92);
			\node [white] at (1,1.85) {\small $\mathrm{RMSE} = 58.0$};
			\end{scope}
			\end{tikzpicture}} &
		\raisebox{-.5\height}{
			\begin{tikzpicture}
			\begin{scope}[spy using outlines={rectangle,yellow,magnification=1.55,size=18mm, connect spies}]
			\node {\includegraphics[scale=0.35]{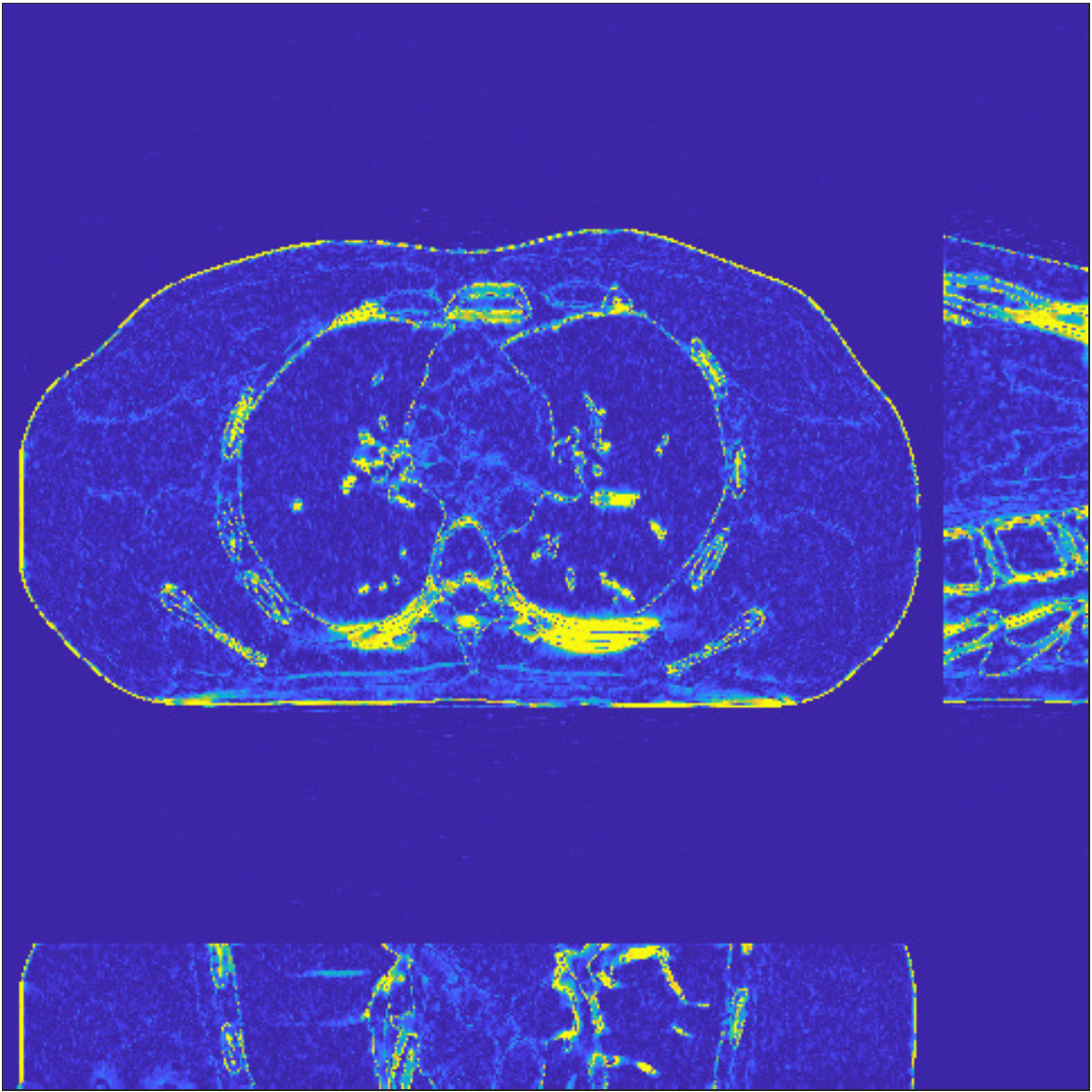}};
\spy on (0.9, 0.1) in node [left] at (-0.41,1.92);
			\node [white] at (1,1.85) {\small $\mathrm{RMSE} = 29.3$};
			\end{scope}
			\end{tikzpicture}} &
		\raisebox{-.5\height}{
			\begin{tikzpicture}
			\begin{scope}[spy using outlines={rectangle,yellow,magnification=1.55,size=18mm, connect spies}]
			\node {\includegraphics[scale=0.35]{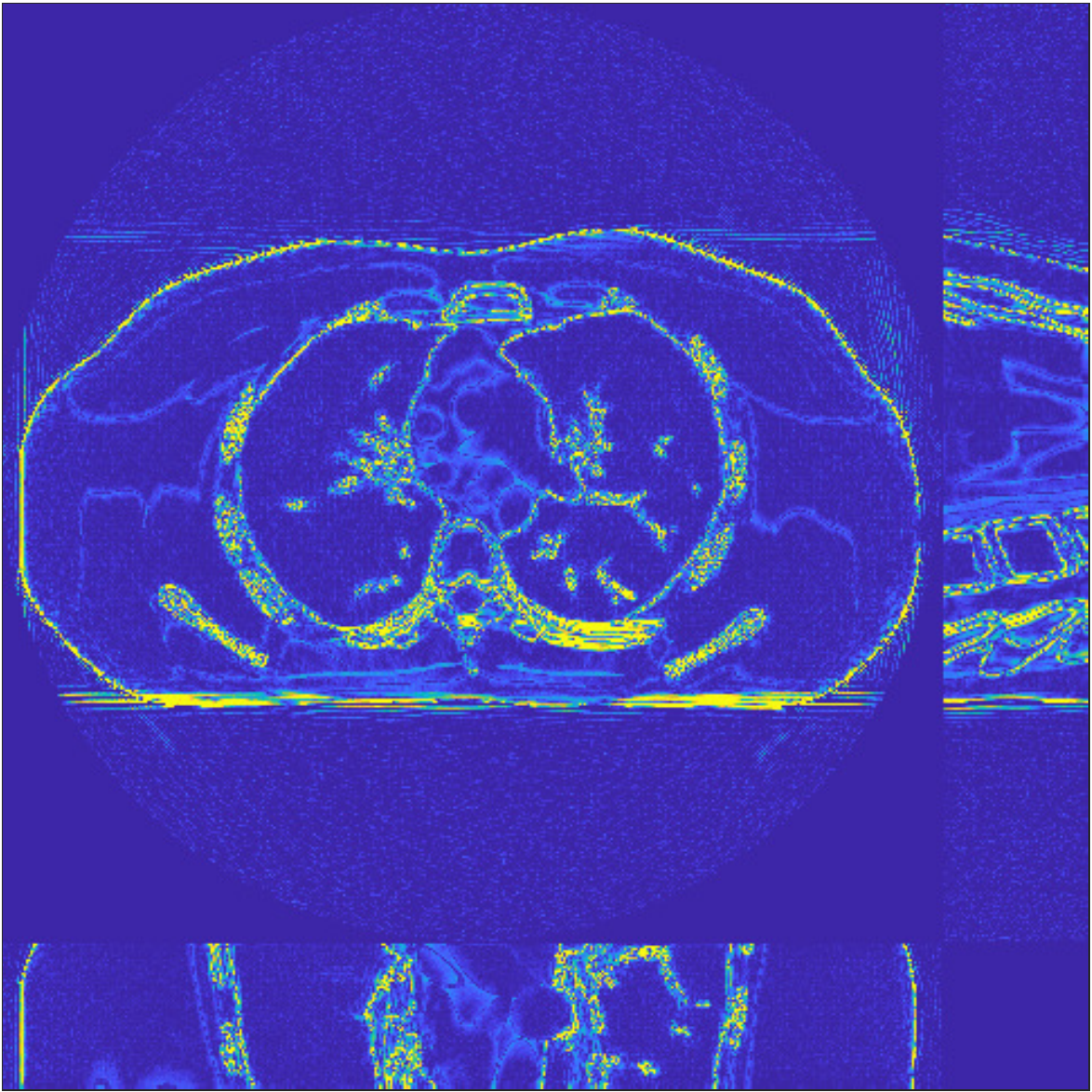}};
\spy on (0.9, 0.1) in node [left] at (-0.41,1.92);
			\node [white] at (1,1.85) {\small $\mathrm{RMSE} = 27.2$};
			\end{scope}
			\end{tikzpicture}} &
		\raisebox{-.5\height}{
			\begin{tikzpicture}
			\begin{scope}[spy using outlines={rectangle,yellow,magnification=1.55,size=18mm, connect spies}]
			\node {\includegraphics[scale=0.35]{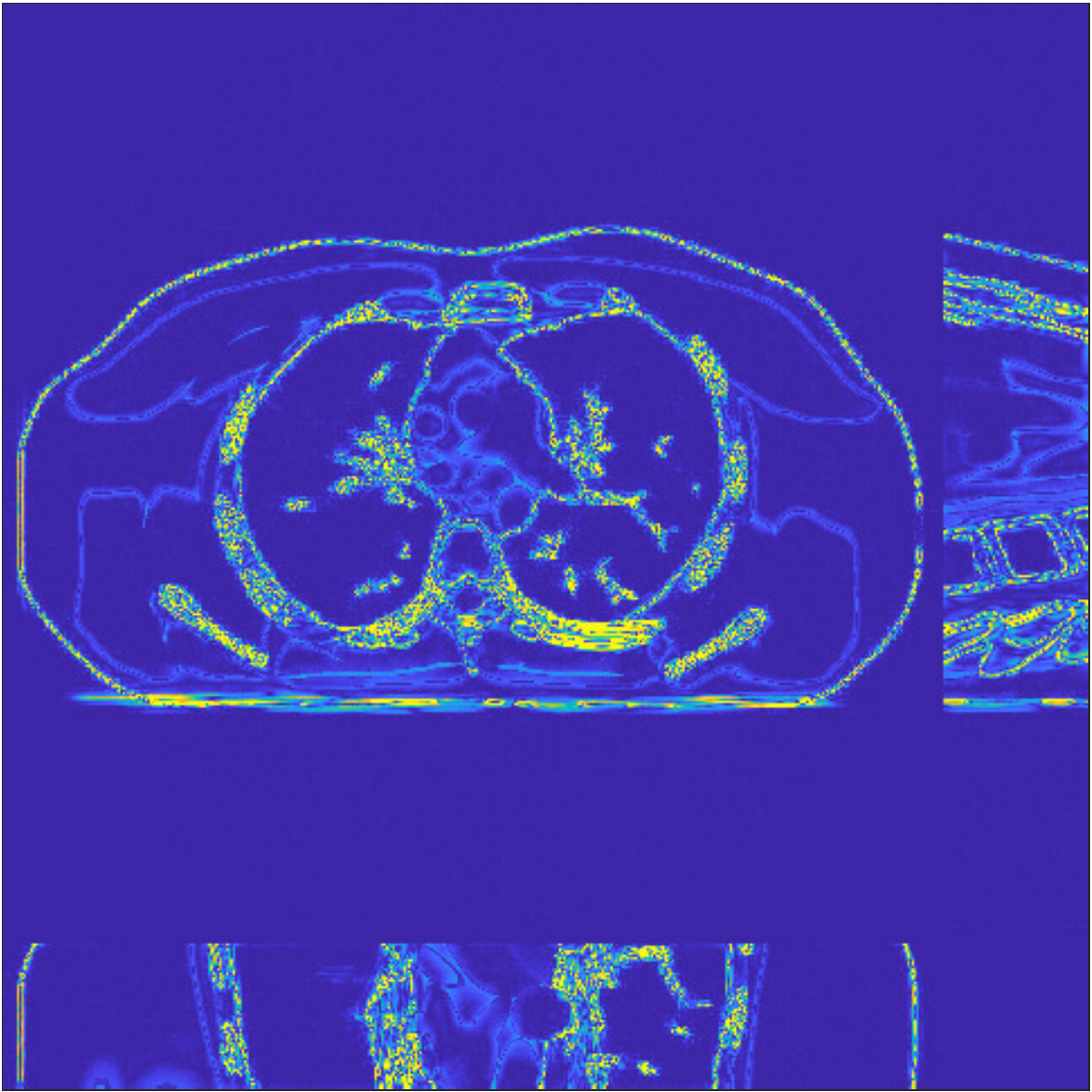}};
\spy on (0.9, 0.1) in node [left] at (-0.41,1.92);
			\node [white] at (1,1.85) {\small \color{yellow}{$\mathrm{RMSE} = 22.2$}};
			\end{scope}
			\end{tikzpicture}} \\

		\raisebox{-.5\height}{\begin{turn}{+90} \small{$123$ views} \end{turn}}~ &
		\raisebox{-.5\height}{
			\begin{tikzpicture}
			\begin{scope}[spy using outlines={rectangle,yellow,magnification=1.55,size=18mm, connect spies}]
			\node {\includegraphics[scale=0.35]{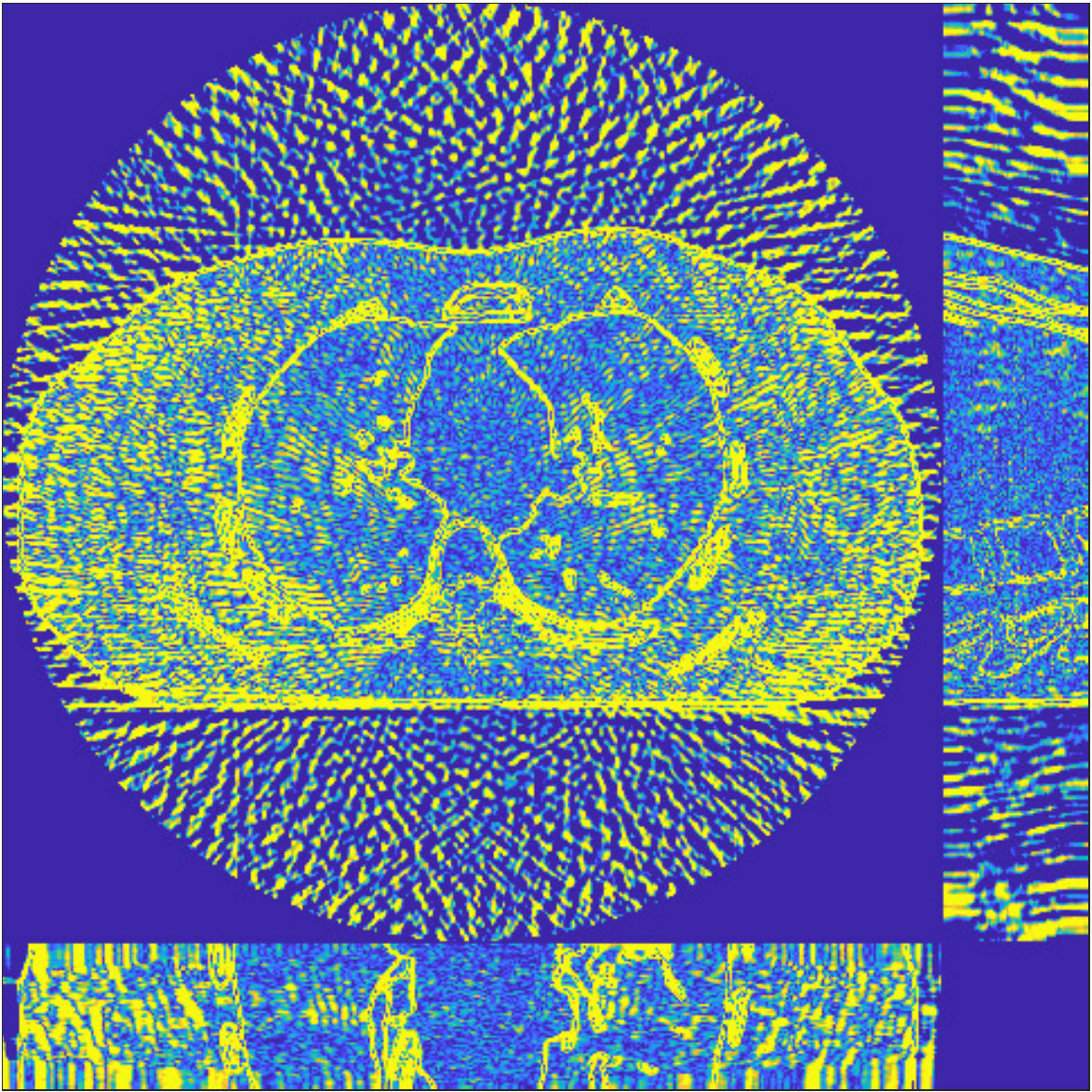}};
\spy on (0.9, 0.1) in node [left] at (-0.41,1.92);
			\node [white] at (1,1.85) {\small $\mathrm{RMSE} = 80.2$};
			\end{scope}
			\end{tikzpicture}} &
		\raisebox{-.5\height}{
			\begin{tikzpicture}
			\begin{scope}[spy using outlines={rectangle,yellow,magnification=1.55,size=18mm, connect spies}]
			\node {\includegraphics[scale=0.35]{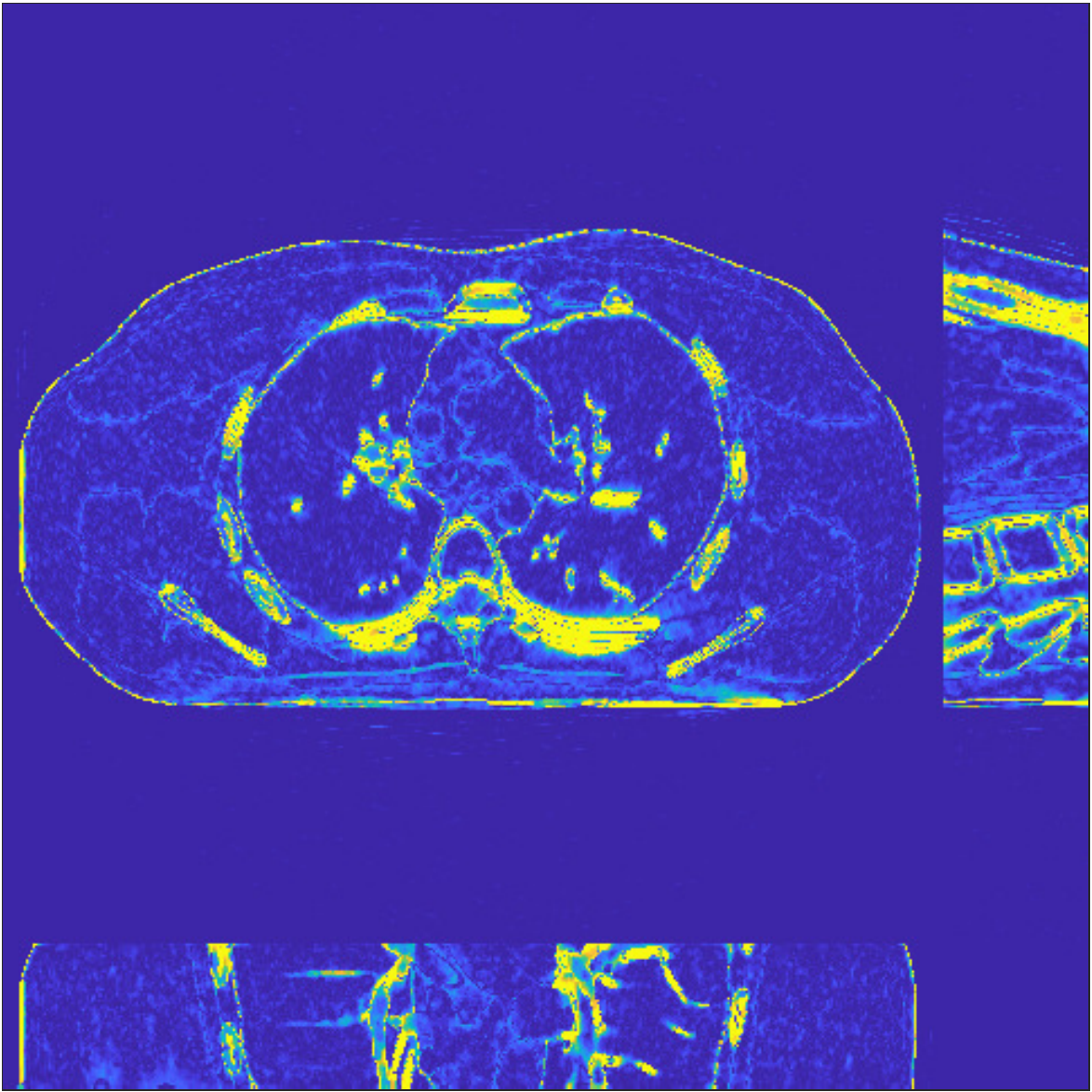}};
\spy on (0.9, 0.1) in node [left] at (-0.41,1.92);
			\node [white] at (1,1.85) {\small $\mathrm{RMSE} = 36.9$};
			\end{scope}
			\end{tikzpicture}} &
		\raisebox{-.5\height}{
			\begin{tikzpicture}
			\begin{scope}[spy using outlines={rectangle,yellow,magnification=1.55,size=18mm, connect spies}]
			\node {\includegraphics[scale=0.35]{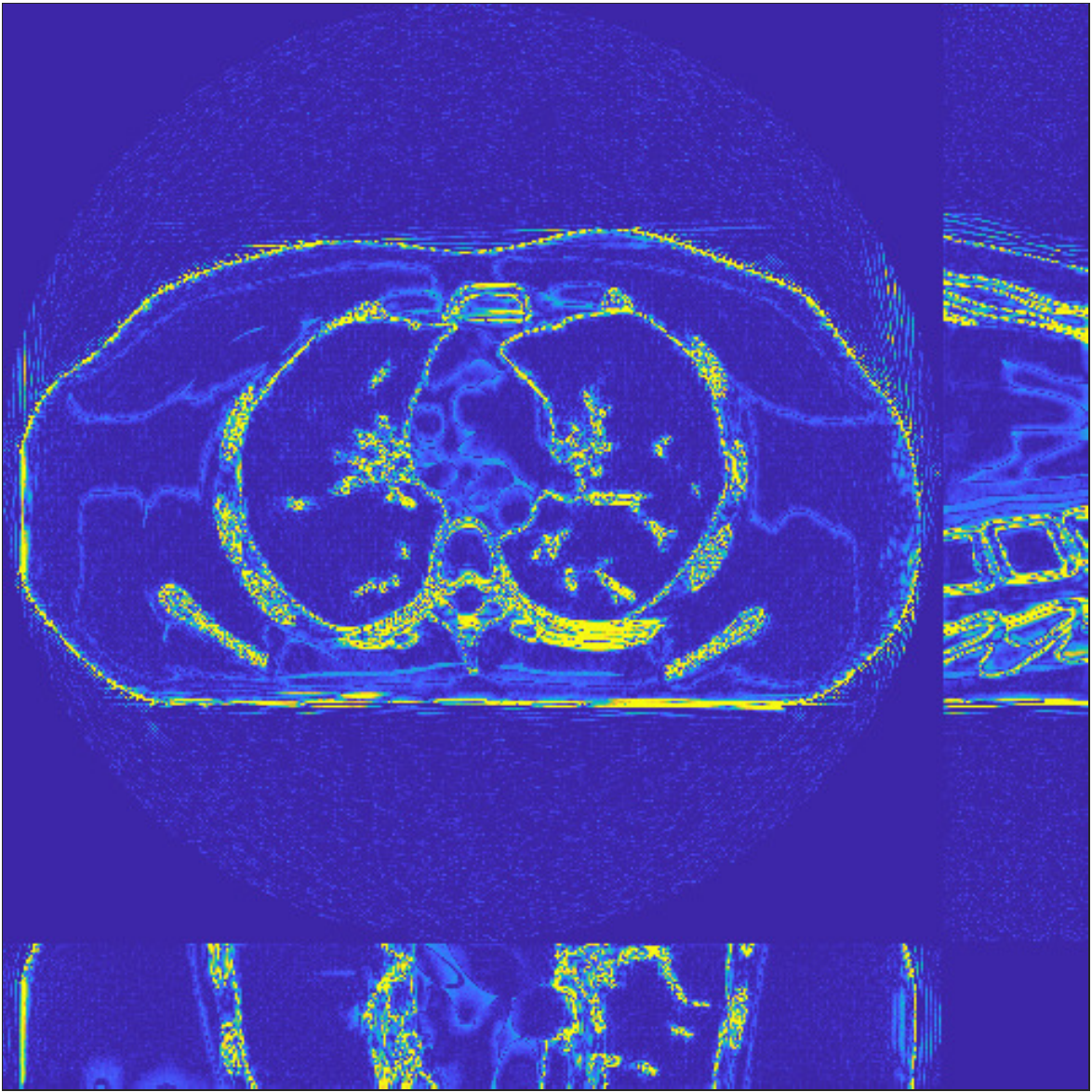}};
\spy on (0.9, 0.1) in node [left] at (-0.41,1.92);
			\node [white] at (1,1.85) {\small $\mathrm{RMSE} = 30.2$};
			\end{scope}
			\end{tikzpicture}} &
		\raisebox{-.5\height}{
			\begin{tikzpicture}
			\begin{scope}[spy using outlines={rectangle,yellow,magnification=1.55,size=18mm, connect spies}]
			\node {\includegraphics[scale=0.35]{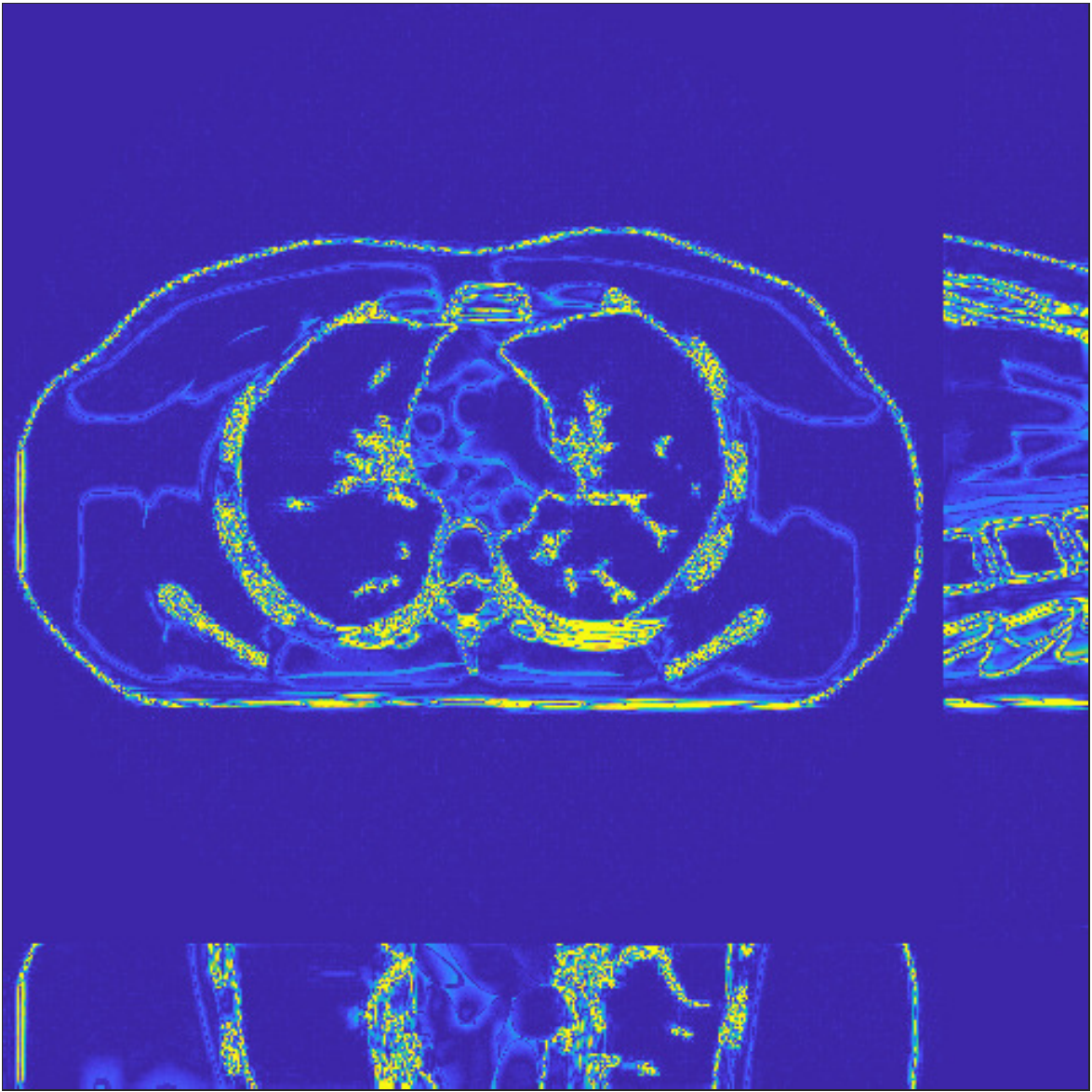}};
\spy on (0.9, 0.1) in node [left] at (-0.41,1.92);
			\node [white] at (1,1.85) {\small \color{yellow}{$\mathrm{RMSE} = 25.6$}};
			\end{scope}
			\end{tikzpicture}} 
		
	\end{tabular}
	\caption{Corresponding error images of 3D reconstructed images from different X-ray CT reconstruction models with different number of views (axial 3D cone-beam geometry and $\rho_0 = 10^5$; display window is $[0, 100]$ HU; displayed for the central axial, sagittal, and coronal planes).}
	\label{fig:3DCTrecon_err}
\end{figure*}
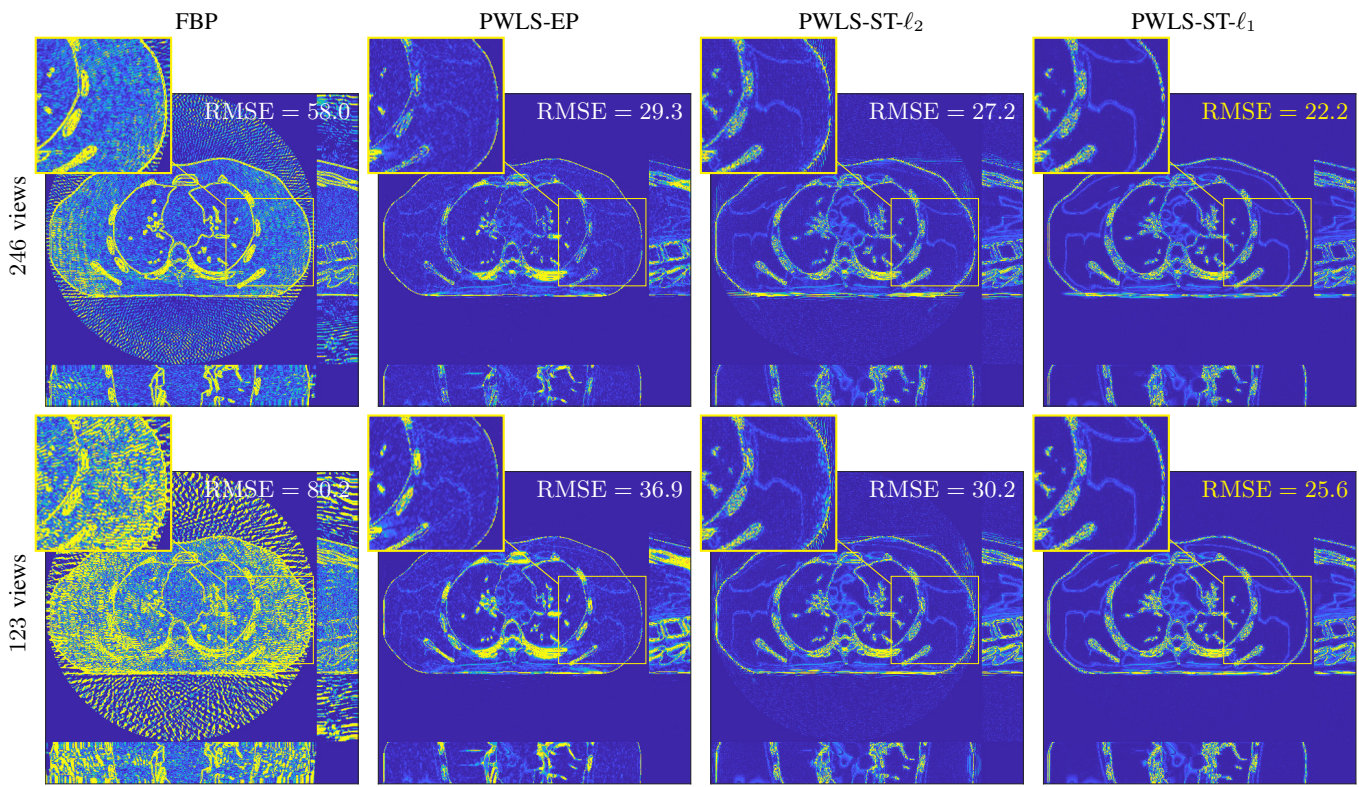

Fig.~\ref{fig:3DCTrecon_err} shows the error images  (corresponding to Fig.~\ref{fig:3DCTrecon}) of 3D reconstructed images with the FBP, PWLS-EP,  \mbox{PWLS-ST-$\ell_2$}, and \mbox{PWLS-ST-$\ell_1$} methods. The proposed PWLS-ST-$\ell_1$ method achieves the lowest RMSE by reducing more noise and reconstructing structural details better, compared to the other methods. In particular, PWLS-ST-$\ell_2$ has some ringing artifacts around the edges with high transition for both $123$ and $246$ views (see zoom-ins).

\begin{figure*}[!t]
	\centering  	
	\begin{tikzpicture}
	\begin{scope}[spy using outlines={rectangle,yellow,magnification=1.9,size=8mm, connect spies}]
	\node {\subfigure[$\lambda =  10^{-2},\gamma/\lambda = 50$]{\includegraphics[width=0.48\textwidth]{./2D_GE_Data/GEdata_FBPInit_246L1_lam10e-03_zeta50_30kap1_30kap2_iter2_1000outer_learn110.pdf}}};
	\end{scope}
	\end{tikzpicture}
	\begin{tikzpicture}
	\begin{scope}[spy using outlines={rectangle,yellow,magnification=1.9,size=8mm, connect spies}]
	\node {\subfigure[$\lambda =  10^{-2},\gamma/\lambda = 200$]{\includegraphics[width=0.48\textwidth]{./2D_GE_Data/GEdata_FBPInit_246L1_lam10e-03_zeta200_30kap1_30kap2_iter2_1000outer_learn110.pdf}}};
	\end{scope}
	\end{tikzpicture}\\
			\vspace{-0.1in}
	\begin{tikzpicture}
	\begin{scope}[spy using outlines={rectangle,yellow,magnification=1.9,size=8mm, connect spies}]
	\node {\subfigure[$\lambda =  10^{-2},\gamma/\lambda = 500$]{\includegraphics[width=0.48\textwidth]{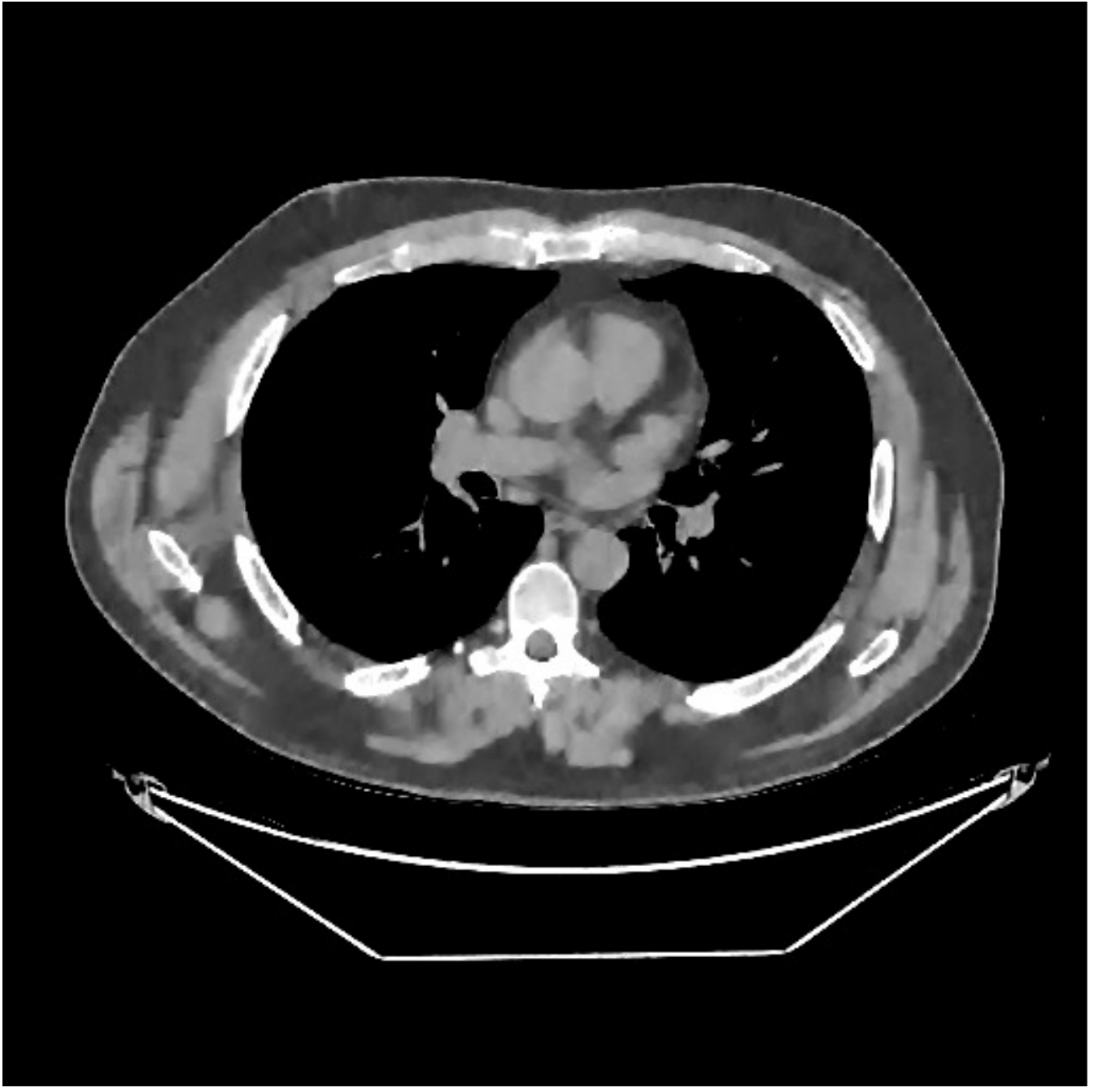}}};
	\end{scope}
	\end{tikzpicture}
	\begin{tikzpicture}
	\begin{scope}[spy using outlines={rectangle,yellow,magnification=1.9,size=8mm, connect spies}]
	\node {\subfigure[$\lambda =  10^{-2},\gamma/\lambda = 1000$]{\includegraphics[width=0.48\textwidth]{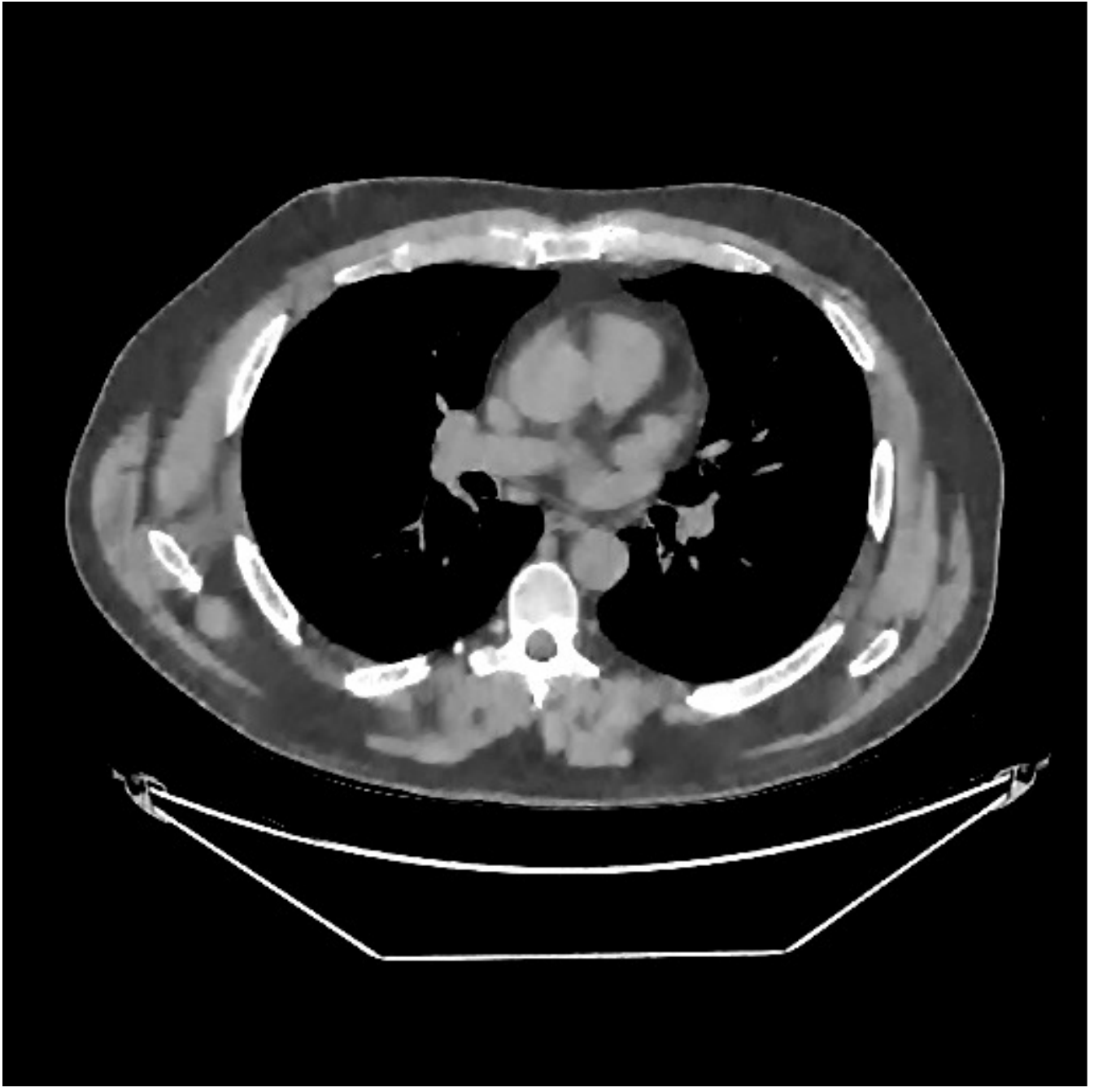}}};
	\end{scope}
	\end{tikzpicture}
		\vspace{-0.15in}
	\caption{Comparison of 2D reconstructed images from clinical data for the proposed PWLS-ST-$\ell_1$ method with $25$\% ($246$) views and different $\gamma$ values. Display window is $[800, 1200]$ HU.}
	\label{fig:2D_GE_thres}
\end{figure*}

Fig.~\ref{fig:2D_GE_thres} shows an additional comparison of 2D reconstructed images from clinical data for the proposed PWLS-ST-$\ell_1$ method with $25$\% ($246$) views and different $\gamma$ values. 
The reconstruction with $\gamma/\lambda = 1000$ is very smilar to the one with $\gamma/\lambda = 500$, and only slightly smoother than the one with $\gamma/\lambda = 200$. These results show that in reconstructing the clinical data, once the $\lambda$ value is properly chosen, PWLS-ST-$\ell_1$ is robust to a wide range of $\gamma$ values.

\newpage

\newpage

\newpage



\end{document}